\documentclass[]{imag-ms-template}

\usepackage{makecell}
\usepackage{amsmath,bm}
\usepackage{graphicx}
\usepackage{subcaption}
\usepackage{listings}
\usepackage{booktabs}
\usepackage{textcomp}
\usepackage{url}
\usepackage{array}
\usepackage{multirow}
\usepackage{wasysym}
\usepackage{wrapfig}
\usepackage[table, x11names]{xcolor}
\usepackage{boldline}
\usepackage{float}
\usepackage[T1]{fontenc}

\title{Decoding Naturalistic Emotion Dynamics from the Brain: An LLM-Enhanced Regression Framework} 

\author{Lemei Zhang$^{1}$, Peng Liu$^{1\ast}$, Hans Dahle Kvadsheim$^{1}$, August Sætre Aasvær$^{1}$, \\
Shuer Ye$^{2}$, Reza Bonyadi$^{3}$, Maryam Ziaei$^{2}$, Jon Atle Gulla$^{1}$\\
{\small $^{1}$Department of Computer Science, NTNU,Trondheim 7491, Norway}\\
{\small $^{2}$Kavli Institute for Systems Neuroscience, NTNU, Trondheim 7491, Norway}\\
{\small $^{3}$Microsoft, Trondheim 7012, Norway}\\
{\small $^\ast$Correspondence:  peng.liu@ntnu.no}
}

\begin{document} 

\maketitle


\begin{abstract}
  Decoding emotional states from neural signals has been typically framed as a discrete, single-label classification task based on emotionally stable stimuli, a formulation that oversimplifies the continuous, fluid, and co-occurring nature of human affect. This study reconceptualizes emotion decoding by adopting a multi-target regression framework to track multiple overlapping emotional dimensions as continuous trajectories over time. Leveraging the robust generalization capabilities of Large Language Models (LLMs), we extracted fine-grained, continuous sentiment profiles from a naturalistic auditory narrative, Alice in Wonderland, to serve as scalable proxies for subjective affect from human fMRI dataset. Departing from standard classification paradigms or mass-univariate subtractive contrasts that filter out network dynamics, we leverage regularized and kernel-based machine learning algorithms as continuous estimators to track the magnitude of macroscale neural state variations. We demonstrate that models trained on temporal snapshots of Dynamic Functional Connectivity (DFC) significantly outperform static region-of-interest (ROI) amplitude representations, effectively capturing continuous emotional trajectories under rapidly fluctuating narrative input. Furthermore, by implementing graph-theoretical Explainable AI (XAI) techniques, we deconstruct the underlying predictive features to reveal highly interpretable, emotion-specific topological configurations. Collectively, these results highlight the utility of LLM-automated annotation in affective neuroscience and provide compelling empirical evidence for psychological constructionist frameworks, demonstrating that dynamic, distributed network interactions offer superior explanatory power over strictly locationist accounts of emotion.
\end{abstract}

\section{Introduction}
Decoding emotions from neural activations has traditionally been framed as a discrete, single-target classification task \citep{Saarimaki2016, saarimaki2022, Kassam2013, Wager2015, Xu2023}. However, both categorical and multidimensional theories of emotion emphasize that emotional states are neither static nor mutually exclusive; rather, they vary continuously in intensity, frequently co-occur, and shift dynamically over time \citep{Plutchik1980, russell_circumplex_2003}. Consequently, formulating emotion decoding as a single-label task therefore risks oversimplifying the complexity of human emotional experience. To address this limitation, the present study adopts a multi-target regression framework, enabling the simultaneous tracking of multiple emotional dimensions as continuous signals, thereby more faithfully capturing the fluid and overlapping nature of real-world affect.

Most prior research in neural decoding has relied on emotionally static stimuli and tightly controlled experimental designs optimized to evoke isolated, dominant emotions over fixed intervals \citep{Kassam2013, saarimaki2022}. To identify sustained, stable brain states, researchers have typically aggregated neural data over discrete blocks—such as 45-second musical excerpts or short text passages—to classify canonical categories like happiness or sadness. Recent paradigms have begun to bridge this gap by transitioning toward prolonged, naturalistic stimuli. For instance, \cite{Xu2023} utilized 10-minute movie clips to elicit reliable emotional responses, while \cite{roshanaei2025eeg} leveraged graph-theoretical analyses of EEG connectivity to distinguish subtle, transient "emotional episodes" during voice-guided imagination. However, even within these naturalistic settings, emotion is frequently treated as a stable categorical epoch rather than a continuous, high-frequency trajectory. Such frameworks fail to fully reflect the context-rich, evolving affective landscapes inherent to real-world experiences, particularly those driven by narrative complexity.

Methodologically, earlier work on traditional classification paradigms requires dense, expert-labeled neural data, which is typically acquired via functional magnetic resonance imaging (fMRI) or electroencephalography (EEG) \citep{Saarimaki2016, huang2023graph, Xu2023}. However, generating continuous emotional annotations for naturalistic, text-based stimuli is exceptionally resource-intensive, highly subjective, and difficult to scale, leaving vast repositories of legacy neuroimaging datasets underutilized. Large Language Models (LLMs), characterized by robust zero- and few-shot generalization capabilities \citep{few_shot, scaling_laws}, offer a scalable, automated alternative for narrative annotation. Despite noted limitations regarding prompt sensitivity in low-shot contexts \citep{reynolds2021prompt}, prior work confirms that high-parameter LLMs can reliably generate fine-grained emotional annotations for complex textual narratives \citep{gpt_annotation1, gpt_annotation2}, enabling standardized, scalable labeling within validated taxonomic frameworks such as Plutchik’s eight-emotion model \citep{Plutchik1980}.

The integration of LLMs into affective neuroscience has already yielded promising insights. \cite{vos2025decoding} demonstrated that LLM embeddings can successfully map textual emotional expressions to localized neuroanatomical structures through computational inference, offering a blueprint for understanding how language models represent affective space without direct neural measurements. Conversely, \cite{santavirta2025gpt} utilized fMRI measures to show that GPT-4 can accurately predict human emotional ratings and model their corresponding hemodynamic responses across multimodal visual and narrative stimuli. However, such analysis relied on a traditional mass-univariate approach, convolving emotional ratings with a canonical hemodynamic response function and isolating affective regions via  subtracting control conditions from the main effect of emotion. While powerful for identifying where the brain processes specific emotions that is consistent with a locationist perspective that treats brain regions as isolated specialists, this subtractive approach inherently filters out the coordinated, time-varying interactions between distributed brain networks.

Furthermore, a significant gap remains regarding the interpretability of network-level predictive models. While connectivity-based machine learning approaches can achieve high predictive accuracy, they frequently operate as "black boxes", failing to provide mechanistic explanations for how specific topological configurations map onto distinct affective states. As machine learning becomes increasingly central to neuroimaging, Explainable AI (XAI) frameworks have become essential to verify model transparency, validate that data-driven patterns align with established neuroscientific theory, and uncover novel neurobiological signatures \citep{farahani2022explainable}. For example, recent investigations have used XAI to bridge artificial and biological vision, identifying shared neural correlates of emotional perception during naturalistic viewing \citep{borriero2024explainable}. However, extending this interpretive capacity to the internal, time-varying functional connectivity patterns that underpin continuous narrative comprehension remains an open challenge.

The objectives of the current study are two-fold. First, \textbf{we evaluate the efficacy of a multi-target regression framework in decoding overlapping, continuous emotional trajectories from complex, naturalistic auditory narratives using a semantically dense fMRI dataset based on \emph{Alice in Wonderland}.} Second, \textbf{we implement graph-theoretical XAI techniques to deconstruct model behavior and extract meaningful neural signatures underlying these decoded trajectories.} We hypothesize that Dynamic Functional Connectivity (DFC)—captured via temporal network snapshots—will significantly outperform static region-of-interest (ROI)-based BOLD amplitudes in tracking these high-frequency emotional trajectories. This prediction aligns with psychological constructionist frameworks, which argue that affective states are encoded within distributed, time-varying network interactions rather than localized regional activations \citep{emotion_brain_basis}. Furthermore, we predict that these analytical models will reveal stable, interpretable network signatures characterized by emotion-specific topological configurations. Ultimately, this approach aims to demonstrate how leveraging machine learning algorithms as continuous estimators of network state changes can provide superior explanatory power over traditional univariate contrasts, clarifying the distributed neural architecture of dynamic affect.

\section{Method}

\begin{figure}
    \begin{center}
    \includegraphics[width=401pt, height=179pt]{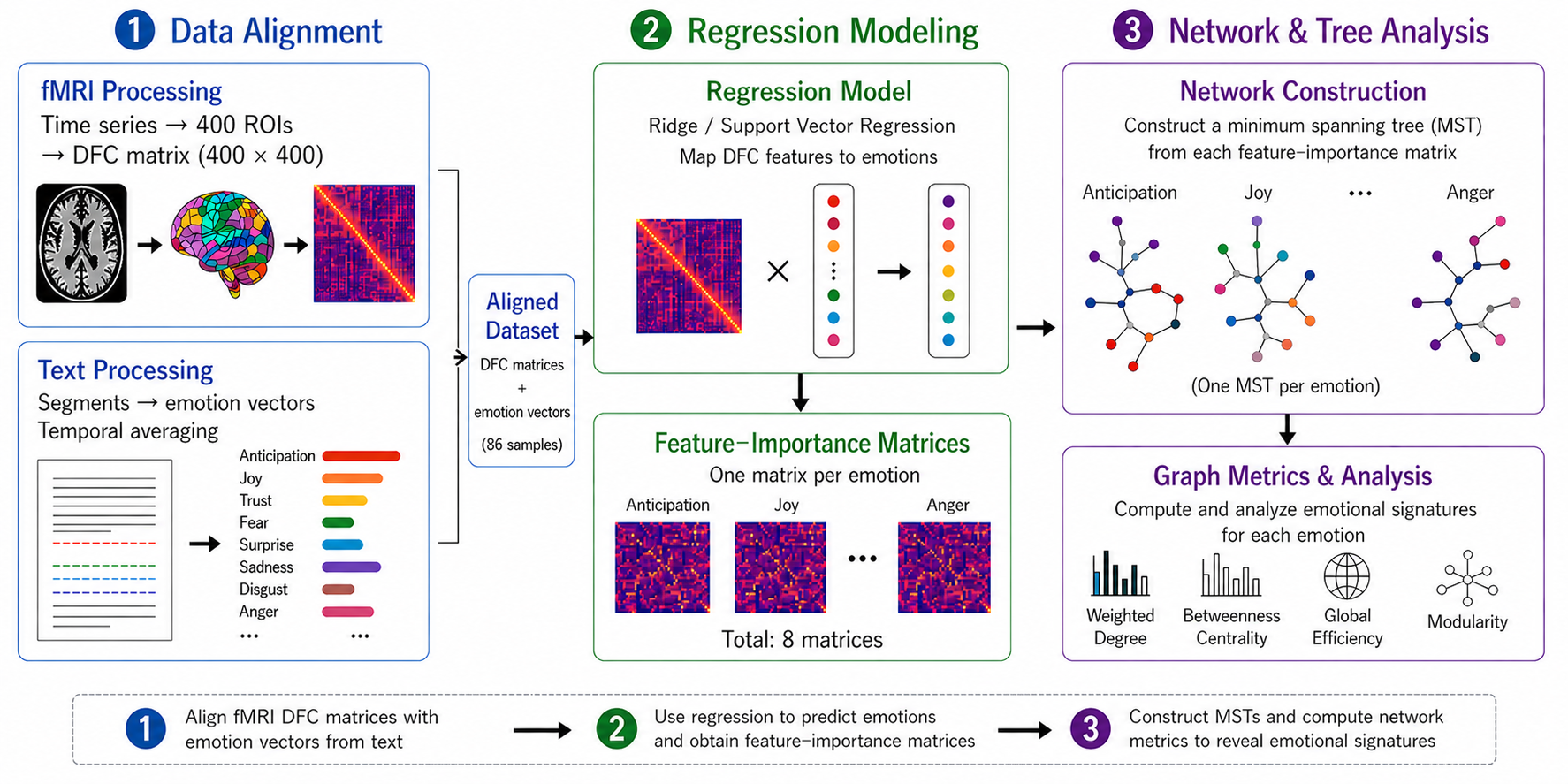}
    \caption{Overview of the proposed pipeline linking dynamic functional connectivity (DFC) with emotion representations. (1) Data alignment: fMRI time series are parcellated into 400 ROIs to form DFC matrices, while text is segmented and converted into temporally averaged emotion vectors; both modalities are aligned into a joint dataset. (2) Regression modeling: designed machine learning models map DFC features to emotion vectors, producing feature-importance matrices per emotion. (3) Network analysis: Emotion-specific MSTs are constructed and analyzed with graph metrics to characterize network signatures.}
    \label{fig:DFC_explanation_pipeline}
    \end{center}
    \vspace{-12pt}
\end{figure}

Fig.\ref{fig:DFC_explanation_pipeline} illustrates the architecture of the proposed three-stage analytical pipeline to map neural features to continuous emotional representations. In the initial stage, fMRI-derived ROI amplitude time-series and DFC matrices are temporally synchronized and integrated with the continuous text-based emotion vectors. In the second stage, multivariate regression frameworks map these localized ROI and DFC configurations to the corresponding affective targets to generate empirical feature-importance matrices. Finally, these connectivity-derived feature weight matrices are subjected to graph-theoretical decomposition to construct emotion-specific minimum spanning trees (MSTs), enabling the computation of topological metrics that characterize distinct, distributed neural signatures for discrete emotional dimensions.

\subsection{Dataset Description} 
\subsubsection{Alice Dataset}
The openly available  Alice dataset was used in this study. This  multimodal neuroimaging dataset contains structural and functional MRI as well as EEG data from twenty-six participants, publicly available through OpenNeuro\footnote{https://openneuro.org/datasets/ds002322/versions/1.0.4}. In this experiment, participants listened to an auditory narration of the first chapter of \textit{Alice in Wonderland}, read by a female speaker. The chapter contains 2,129 words across 84 sentences, and the audio was slowed by 20\% to improve comprehension during the scanning. Although the original study included twenty-nine participants, two were excluded due to excessive head motion that exceeded an absolute threshold of 1 mm, and one for failing a comprehension test, resulting in a final sample of fifteen females and eleven males aged 18–24. All participants were right-handed native English speakers with no reported psychological or neurological conditions.

During data acquisition, whole-brain fMRI data were collected on a 3T scanner with a 32-channel head coil while participants listened to the 13-minute auditory stimulus via MRI-compatible headphones. For functional images, thirteen participants were scanned using a T2-weighted echo planar imaging (EPI) sequence (repetition time = 2000 ms, echo time = 27 ms, voxel size = 3 $\times$ 3 $\times$ 3 mm, flip angle = 77$^\circ$, acceleration factor = 2, field of view = 216 $\times$ 216 mm, matrix size = 72 $\times$ 72, 44 slices). Sixteen participants were scanned with a three-echo EPI sequence (repetition time = 2000 ms, echo time = 27.5 ms, voxel size = 3.75 $\times$ 3.75 $\times$ 3.8 mm, flip angle = 77$^\circ$, acceleration factor = 2, field of view = 240 $\times$ 240 mm, matrix size = 72 $\times$ 72, 33 slices). Anatomical images were obtained using a T1-weighted gradient-echo sequence (voxel size = 1 $\times$ 1 $\times$ 1 mm$^3$). More details about the scanning parameters can be found in the dataset source on OpenNeuro.

\subsubsection{Preprocessing} \label{m: preprocessing}
Functional and anatomical neuroimaging data were preprocessed using Statistical Parametric Mapping (SPM) \citep{SPM8}, adhering to the analytical pipeline established in prior investigations utilizing this specific dataset \citep{Kassam2013, Xu2023, lettieri2019emotionotopy, huang_dataset2}. To correct head-motion artifacts, spatial realignment was performed utilizing a 2-degree B-spline interpolation embedded within a six-parameter rigid-body transformation. Functional and anatomical images were subsequently co-registered via a mutual information criterion. To optimize the signal-to-noise ratio (SNR) and alleviate residual spatial disparities, the functional data were smoothed using a 3-mm isotropic Gaussian kernel and subsequently normalized to the Montreal Neurological Institute (MNI) space. Low-frequency scanner drifts and physiological noise were attenuated by applying a high-pass filter with a cutoff frequency of $1/128\text{ Hz}$.Following voxel-wise preprocessing, dimensionality reduction was performed through atlas-based region-of-interest (ROI) time-series extraction. Specifically, cortical blood-oxygen-level-dependent (BOLD) signals were parcellated into 400 discrete functional nodes using the Schaefer 400-parcel cortical atlas \citep{Schaefer2018}. These 400 constituent ROIs are hierarchically nested within seven canonical large-scale resting-state networks including Visual, Dorsal Attention, Salience/Ventral Attention, Somatomotor, Limbic, Control, and Default Mode networks.

\subsection{Automatic Emotional Labeling} \label{m: labeling}
Prior to automated emotion labeling of the Alice dataset, the audio stimulus was transcribed and the text segmented. Multiple segmentation approaches were evaluated, including sentence-based rules and fixed temporal windows. Sentence-based segmentation produced substantial variability in length, ranging from 2 to 102 words. In contrast, fixed-duration windows frequently yielded segments that either terminated mid-word or failed to preserve semantic coherence, limiting the effectiveness of large language model–based sentiment annotation. Accordingly, an eight-second window served as the baseline, with each segment manually adjusted to ensure semantic completeness while minimizing intervention. To better capture the temporal continuity of emotional expression, each segment included a four-second overlap with the preceding segment.

Following segmentation, several pretrained large language models (LLMs) of varying scales and architectures were evaluated for sentiment extraction. Given evidence that emotional responses and neural activations accumulate over continuous stimuli \citep{Xu2023}, the models including LLaMA-3.1-70B\footnote{https://huggingface.co/meta-llama/Llama-3.1-70B}, Gemma-9B-IT\footnote{https://huggingface.co/google/gemma-2-9b-it}, Mixtral-8×7B-instruct\footnote{https://huggingface.co/mistralai/Mixtral-8x7B-Instruct-v0.1}, and GPT-4 \citep{openai2024gpt4technicalreport}, were instructed to incorporate the current segment together with the preceding $n$ segments as contextual input, as defined in Eq. \ref{eq:context}. Each model generated intensity scores ranging from 0 to 1 for Plutchik’s eight emotion categories \citep{Plutchik1980}, with the temperature fixed at 0 to ensure deterministic and reproducible outputs. Prompts extracting emotions from listed LLMs are listed in \ref{appendix prompt}.
\begin{equation} \label{eq:context}
Context_n = \cup_{i=0}^{n-1} segment_i
\end{equation}
To evaluate labeling accuracy, a random sample of 20\% of the segments was selected for human validation. Ten evaluators who were university students (five male, five female; aged 22 to 27 years) and fluent in English, were instructed to review the contextual input prior to assessing the model-generated emotional labels. The evaluators rated the model annotations with respect to the sentiments expressed in the input context, using an 11-point scale ranging from 0 (inaccurate) to 10 (perfect correspondence with the text). Among the models tested, GPT-4 was the only system that consistently produced well-structured and prompt-stable outputs, with the mean agreement score of 7.83 (SD = 1.38), and was therefore selected for final annotation. Instructions for human annotators and the emotional distribution plots are illustrated in \ref{emotion survey} and \ref{appendix emotional analysis}, respectively.

\subsection{ROI-based Emotional Dataset} \label{m: ROI datasets}
To align the LLM annotated segments with 400 ROIs, for each segment, the fMRI volume corresponding to the end of the segment was paired with its emotional label. Alignment was achieved by converting the end-of-segment timestamp $t_{eos_n}$ to the nearest subsequent multiple of 2s, yielding the ROI-time-series timestamp $t_n$:
\begin{equation}  \label{eq: t_n}
        t_n = t_{eos_{n}} + (2 - (t_{eos_{n}}\mod{2})) 
\end{equation}
The corresponding ROI index $i_n$ in the time series was then computed using the repetition time denoted as $TR=2s$:
\begin{equation} \label{eq: index}
i_n = \frac{t_n}{TR} = \frac{t_n}{2s}
\end{equation}
To increase the effective number of samples, the previous and the subsequent $\mathbf{ROI}$ vectors were also included under the assumption that small temporal offsets would not substantially alter the associated emotional state. For each subject $s$, the dataset $D_s$ therefore consisted of the triplets:
\begin{equation} \label{eq: per-subject roi dataset}
D_{s} = \mathbf{U}_{n=1}^{|\mathbf{E}|} \{(\mathbf{ROI}^s_{i_{n - 1}}, \mathbf{e}_{n}), (\mathbf{ROI}_{i_n}^s, \mathbf{e}_{n}), (\mathbf{ROI}^s_{i_{n + 1}}, \mathbf{e}_{n}) \}
\end{equation}
where $\mathbf{ROI}_{i_n}$ denotes the ROI matrix of subject $s$ at index $i_n$, $e_n$ denotes the intensity score of the $i$-th emotion in Plutchik emotion category, and $|\mathbf{E}|$ is the number of emotional labels. The full ROI dataset was obtained by concatenating all subject-level datasets $D = \mathbf{U}_{s=1}^{N} {D_s}$ with $N$ denoting the total number of subjects. This procedure yields a total of $|D|=6234$ samples with corresponding emotional labels.

\begin{figure}
    \begin{center}
    \includegraphics[width=0.95\textwidth, height=162pt]{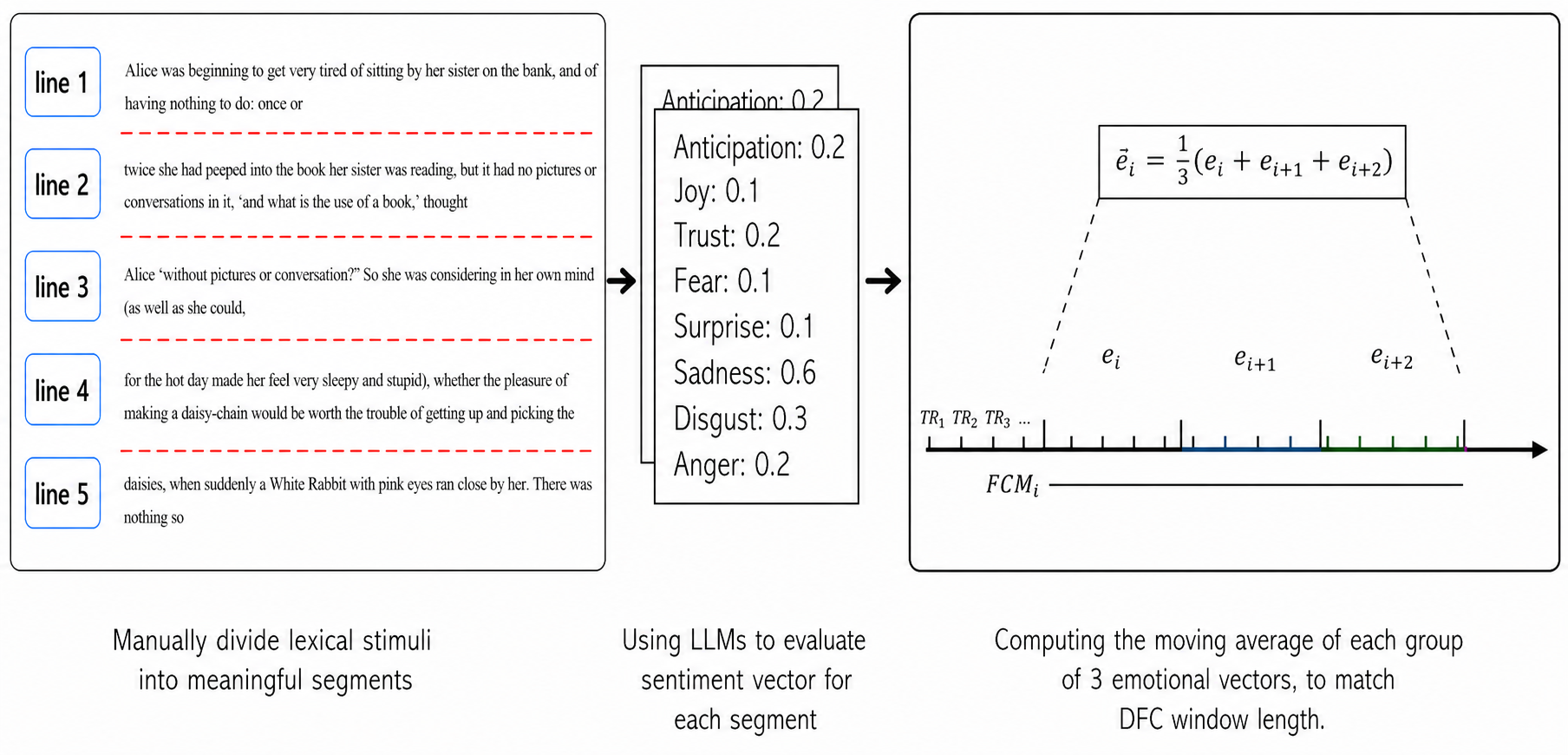}
    \caption{Pipeline for generating DFC datasets. After segmentation of the fMRI-data, ROIs are aligned with continuous LLM-derived emotional labels, which are then processed via a sliding-window approach to construct time-varying connectivity profiles.} 
    \label{fig:dataset_gen}
    \end{center}
\end{figure}

\subsection{DFC Dataset} \label{m: dfc datasets}
As shown in Fig. \ref{fig:dataset_gen}, Dynamic functional connectivity (DFC) matrices were computed from the ROI time series using a timestamp-aligned fixed-window approach. Within each window, functional connectivity was estimated via pairwise Pearson Correlation among all ROIs. The window length $W$, rather than adopting conventional durations of 30 or 60 seconds \citep{30_60seconds1, 30_60seconds2}, was set to $12$ $TRs$ to balance temporal resolution with the requirement to capture rapidly changing emotional states. This choice is consistent with prior work suggesting that such a window length is sufficient for tracking dynamic affective processes \citep{22seconds}. To account for hemodynamic delay from the dataset, the $i_n$-th DFC window was assigned an averaged label $\hat{\mathbf{e}}_i$ derived from $\mathbf{e}_{i_n}$ and its consecutive emotional labels defined as:
\begin{equation} \label{eq:dfc time window label}
    \hat{\mathbf{e}}_i = \frac{1}{3} \left(\mathbf{e}_{i_n} + \mathbf{e}_{i_n+1} + \mathbf{e}_{i_n+2}\right).
\end{equation}
After applying this alignment procedure, 86 DFC matrices per subject were retained for the final dataset.

\subsection{Model Training and Evaluation Strategy}
Considering the trade-off between model interpretability and predictive performance, we evaluated and compared six machine learning models — Linear Regression (LR),  LR with Lasso and Ridge regularization (termed as Lasso Regression and Ridge Regression), Support Vector Regression (SVR) with linear kernel as well as Radial Basis Function (RBF)
kernel, and Random Forest Regression (RFR) — on both DFC and ROI datasets. All machine learning models were executed using Python scikit-learn\footnote{https://scikit-learn.org}. Both datasets were partitioned into training and testing sets using two chronological splits: 90–10 and 80–20, respectively. This unlike conventional approaches commonly employed in related work, which typically divide data into $k$ folds and evaluate model performance through cross-validation. This design was motivated by the consideration that emotional changes in this project evolved in accordance with plot development, rather than being contingent solely upon the immediate input data. All datasets were normalized to a standard normal distribution prior to training.

\begin{table}
\centering
\caption{Optimized Hyperparameters for ROI-based and DFC-based Regression Models.}
\label{tab:hyperparameters_comparison}
\footnotesize 
\begin{tabular}{@{}>{\raggedright\arraybackslash}p{2.75cm} p{5.4cm} p{5cm}@{}}
\toprule
\textbf{Model} & \textbf{ROI Parameters} & \textbf{DFC Parameters} \\
\midrule
Linear Regression & — & — \\
\addlinespace
Lasso Regression & \texttt{alpha=0.0001} & \texttt{alpha=0.0007} \\
\addlinespace
Ridge Regression & \texttt{alpha=1.7575} & \texttt{alpha=3237.4575} \\
\addlinespace
SVR (RBF) & \makecell[l]{\texttt{C=1, epsilon=0.01,} \\ \texttt{kernel=``rbf''}} & \makecell[l]{\texttt{C=10, epsilon=0.01,} \\ \texttt{kernel=``rbf'', gamma=``scale''}} \\
\addlinespace
Linear SVR & \makecell[l]{\texttt{C=0.01, epsilon=0.01,} \\ \texttt{kernel=``linear'', gamma=``scale''}} & \makecell[l]{\texttt{C=0.01, epsilon=0.01,} \\ \texttt{kernel=``linear''}} \\
\addlinespace
Random Forest Regressor & \makecell[l]{\texttt{max\_depth=None,} \\ \texttt{min\_samples\_split=2,} \\ \texttt{min\_samples\_leaf=1,} \\ \texttt{max\_features=``sqrt''}} & \makecell[l]{\texttt{max\_depth=30,} \\ \texttt{min\_samples\_split=5,} \\ \texttt{min\_samples\_leaf=2,} \\ \texttt{max\_features=``sqrt''}} \\
\bottomrule
\end{tabular}
\end{table}

Optimal parameter pairs for each machine learning model were determined using a grid search method, with model performance evaluated based on the Coefficient of Determination ($R^2$) and the Mean Squared Error (MSE) to measure the differences between the predicted value and the ground-truth value. The parameter configurations that yielded the best performance on each dataset are reported in Table \ref{tab:hyperparameters_comparison}. $R^2$ and MSE are defined as follows:
\begin{equation}
\text{MSE} = \frac{1}{S} \sum_{i=1}^{S} (\hat{y}_i - \mathbf{y}_i)^{2}
\label{eq:mse}
\end{equation}

\begin{equation}
R^2 = 1 - \frac{\sum_{i=1}^{S} (\hat{h}_i - \mathbf{y}_i)^{2}}{\sum_{i=1}^{S} (\bar{\mathbf{y}} - \mathbf{y}_i)^{2}}
\label{eq:r_squared}
\end{equation}

Where $\hat{y}_i$ and $y_i$ denote the predicted and ground-truth sentiment values at time stamp $i$, respectively. $\bar{\mathbf{y}}$ represents the mean predicted value across all evaluation samples. The MSE and $R^2$ metrics range from 0 to 1. Lower MSE values indicate smaller deviations between the predictions and ground truth, whereas higher $R^2$ values indicate that the regression model explains a greater proportion of the variance in the dataset.

\subsection{Explanation Generation} \label{sec:explanation}
After model training, post-hoc interpretability analyses were conducted for both ROI and DFC models to identify the features and connectivity patterns most strongly associated with each emotional state. For the linear, Lasso, Ridge and linear SVR models, feature importance could be derived directly from model coefficients, whereas for RFR, separate models were trained for each emotion label, enabling extraction of per-emotion feature importance values. 

For the ROI analyses, each feature was mapped back to its corresponding spatial location of the Schaefer atlas \cite{Schaefer2018}. Combining voxel-to-ROI assignments with model-derived feature importance allowed construction of per-emotion glass-brain visualizations, highlighting the regions most influential for model predictions. These ROI-based maps served as the concept-level interpretability output for the ROI models.

For the DFC models, interpretability focused on connectivity patterns. Feature-importance matrices were interpreted as weighted undirected graphs, from which maximum spanning trees (MSTs) were extracted per emotion using Kruskal’s algorithm \cite{kruskal1956shortest}. This procedure isolated the strongest non-cyclic connectivity structure while reducing graph complexity. Node-wise and global graph measures—including weighted degree, betweenness centrality, modularity, global efficiency, and tree diameter—were computed for each MST to characterize the topological organization of functional connectivity across emotional states. To facilitate higher-level interpretation, these metrics were subsequently aggregated within canonical large-scale networks including Schaefer 7- and 17-network parcellations. Specifically, metrics for ROIs within each network were averaged to derive network-level measures, yielding a system-level perspective on emotion-specific connectivity patterns. An overview of the DFC explanation pipeline is shown in Fig.~\ref{fig:DFC_explanation_pipeline} (2) and (3).

\section{Results}
\subsection{Experimental Results}
\subsubsection{Prediction Effectiveness of Machine Learning Models}
The results obtained from the 90-10 data split are presented in Table \ref{tab:transposed_performance}. Consistent trends and outcomes were observed across all experiments using the 80-20 split, and therefore, due to page limit, the 80-20 split results are provided in \ref{appendix_model_performance}. Statistical significance was assessed at $p\le 0.01$.

\begin{table}
\centering
\caption{Model Performance Comparison ($R^2$ and MSE) across ROI and DFC Datasets, using a 90-10 Train-Test split.}
\vspace{2mm}
\label{tab:transposed_performance}
\footnotesize
\begin{tabular}{@{}>{\raggedright\arraybackslash}m{2.4cm}@{}>{\raggedright\arraybackslash}m{0.8cm}>{\centering\arraybackslash}m{1.1cm}>{\centering\arraybackslash}m{1.1cm}>{\centering\arraybackslash}m{1.1cm}>{\centering\arraybackslash}m{1.1cm}>{\centering\arraybackslash}m{1.1cm}>{\centering\arraybackslash}m{1.2cm}>{\centering\arraybackslash}m{1.2cm}@{}}
\toprule
\textbf{Metric} & \textbf{Data} & \textbf{Linear Reg.} & \textbf{Lasso Reg.} & \textbf{Ridge Reg.} & \textbf{SVR (RBF)} & \textbf{Linear SVR} & \textbf{Random Forest} & \textbf{MM-RFR} \\
\midrule
\textbf{Test $R^2$}  & ROI & 0.0677 & 0.0293 & 0.0800 & $\bm{0.2621}$ & -0.0549 & 0.1978 & 0.1953 \\
                     & DFC & 0.3955 & 0.4742 & 0.5589 & $\bm{0.5931}$ & 0.4291 & 0.1413 & 0.1849 \\
\addlinespace
\textbf{Test MSE}    & ROI & 0.0202 & 0.0212 & 0.0202 & $\bm{0.0161}$ & 0.0233 & 0.0174 & 0.0175 \\
                     & DFC & 0.0066 & 0.0058 & 0.0050 & $\bm{0.0047}$ & 0.0064 & 0.0103 & 0.0097 \\
\bottomrule
\end{tabular}%
\vspace{2mm}
\end{table}

Among the evaluated models, the RBF-based SVR exhibited superior performance on both the ROI and DFC datasets, achieving the lowest MSE values of 0.0161 and 0.0047, as well as the highest $R^2$ values of 0.2621 and 0.5931, respectively. These findings suggest that both the ROI and DFC data contain nonlinear patterns relevant to dynamic sentiment evolution. On the ROI dataset, models from the RFR series attained higher $R^2$ values and lower MSE values compared to linear models on the test set, indicating that RFR models are more effective at capturing the nonlinear structures inherent in the ROI data. Notably, on the DFC dataset, linear models demonstrated comparable performance to their nonlinear counterparts in predicting dynamic connectivity within the brain’s sentiment network. In some cases, linear models even achieved lower MSE values and higher $R^2$ values than the RFR series models.

To isolate predictive signals from confounding neural variance extraneous to linguistic or affective processing, and to systematically quantify the feature utility of brain networks during sentiment decoding, we performed a leave-one-network-out exclusion analysis with a linear regression model. The results are 
\begin{wraptable}{r}{0.4\textwidth}
\centering
\label{tab:leave_one_network_mse}
\footnotesize
\begin{tabular}{l c}
\toprule
\textbf{Left-Out Network} & \textbf{Test MSE} \\
\midrule
Visual (Vis)              & 0.0202 \\
Somatomotor (SomMot)      & 0.0206 \\
Dorsal Attention (DorsAttn) & 0.0203 \\
Ventral Attention (Sal/VentAttn) & 0.0201 \\
Limbic                    & 0.0203 \\
Frontoparietal (Control) & 0.0204 \\
Default Mode (DMN)        & 0.0204 \\
\bottomrule
\end{tabular}
\caption{MSE for Linear Models Trained Using Leave-One-Network-Out Cross-Validation}
\label{tab:leave_out_network}
\vspace{-5pt}
\end{wraptable}
presented in Table \ref{tab:leave_out_network}. Removing any single network produced only negligible changes in performance, with a maximum difference of $0.5\permil$ between the highest and lowest MSE values. However, excluding the Ventral Attention network slightly improved performance, whereas excluding the Visual network had no measurable effect.

To determine whether the targeted ablation of the Visual and Ventral Attention networks alters predictive performance across other adopted models, two additional datasets were constructed: $D_{-SalVentAttn}$, in which the Salience/Ventral Attention network was excluded, and $D_{-SalVentAttn \cup Visual}$, in which both the Visual and Salience/Ventral Attention networks were excluded. The corresponding experimental results are reported in Table \ref{tab:transposed_ablation}. We observed that, aside from a modest improvement in $R^2$ for linear regression, removing the Ventral Attention network alone or in combination with the Visual network did not improve prediction performance. On the contrary, most models exhibited reduced performance. These findings suggest that both the Ventral Attention and Visual networks contribute non-trivial, predictive variance that is essential for continuous affect prediction in this naturalistic paradigm.

\subsection{Explanation Analyses}
\subsubsection{ROI dataset}

\begin{table}
\centering
\caption{Impact of Ventral Attention and Visual Network Ablation on Predictive Performance.}
\label{tab:transposed_ablation}
\footnotesize
\begin{tabular}{>{\raggedright\arraybackslash}m{2.7cm}>{\centering\arraybackslash}m{1.3cm}>{\centering\arraybackslash}m{1.3cm}>{\centering\arraybackslash}m{1.6cm}>{\centering\arraybackslash}m{1.3cm}>{\centering\arraybackslash}m{1.3cm}>{\centering\arraybackslash}m{1.6cm}}
\toprule
\multirow{3.5}{*}{\textbf{Model}} & \multicolumn{3}{c}{\textbf{Test $R^2$}} & \multicolumn{3}{c}{\textbf{Test MSE}} \\
\cmidrule(lr){2-4} \cmidrule(lr){5-7}
 & \textbf{Full} & \textbf{w/o Ven} & \textbf{w/o Ven+Vis} & \textbf{Full} & \textbf{w/o Ven} & \textbf{w/o Ven+Vis} \\
\midrule
\textbf{Linear Reg.}     & 0.0677 & 0.0760 & 0.0729 & 0.0202 & 0.0202 & 0.0203 \\
\textbf{Lasso Reg.}      & 0.0293 & 0.0283 & 0.0267 & 0.0212 & 0.0212 & 0.0212 \\
\textbf{Ridge Reg.}      & 0.0800 & 0.0779 & 0.0717 & 0.0202 & 0.0203 & 0.0205 \\
\textbf{SVR (RBF)}       & $\bm{0.2621}$ & $\bm{0.2562}$ & $\bm{0.2485}$ & $\bm{0.0161}$ & $\bm{0.0163}$ & $\bm{0.0165}$ \\
\textbf{Linear SVR}      & -0.0549 & -0.0577 & -0.0612 & 0.0233 & 0.0234 & 0.0234 \\
\textbf{Random Forest}   & 0.1978 & 0.1979 & 0.2004 & 0.0174 & 0.0174 & 0.0173 \\
\textbf{MM-RF}           & 0.1953 & 0.1947 & 0.1979 & 0.0175 & 0.0175 & 0.0174 \\
\bottomrule
\addlinespace
\multicolumn{7}{l}{\footnotesize \textit{Note:} Ven: Ventral Attention Network; Vis: Visual Network; Full: All ROI features.}
\end{tabular}
\vspace{-5pt}
\end{table}

\begin{figure}
    \centering
    \begin{subfigure}[b]{0.45\textwidth}
        \includegraphics[width=\textwidth]{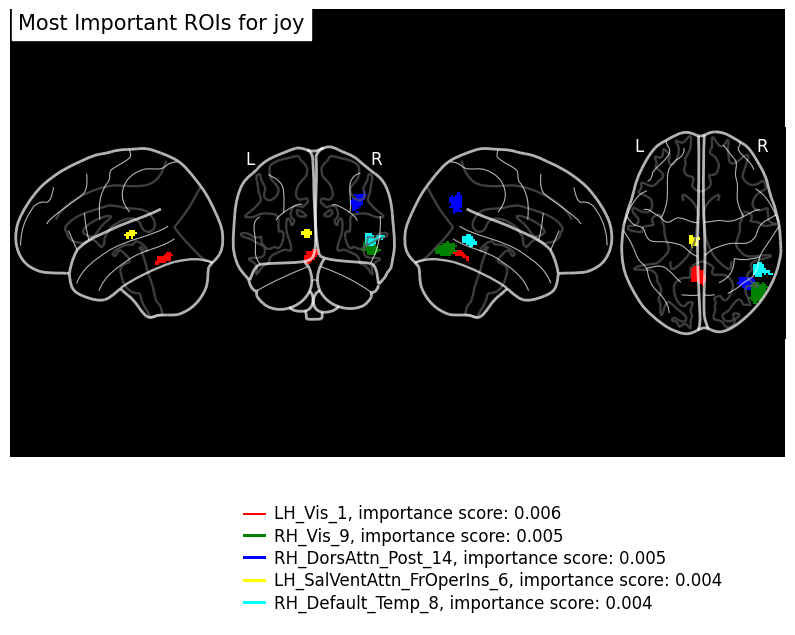}
        \caption{}
        \label{fig:RFR explanation joy}
    \end{subfigure}
    \hspace{0.05\textwidth}
    \begin{subfigure}[b]{0.45\textwidth}
        \includegraphics[width=\textwidth]{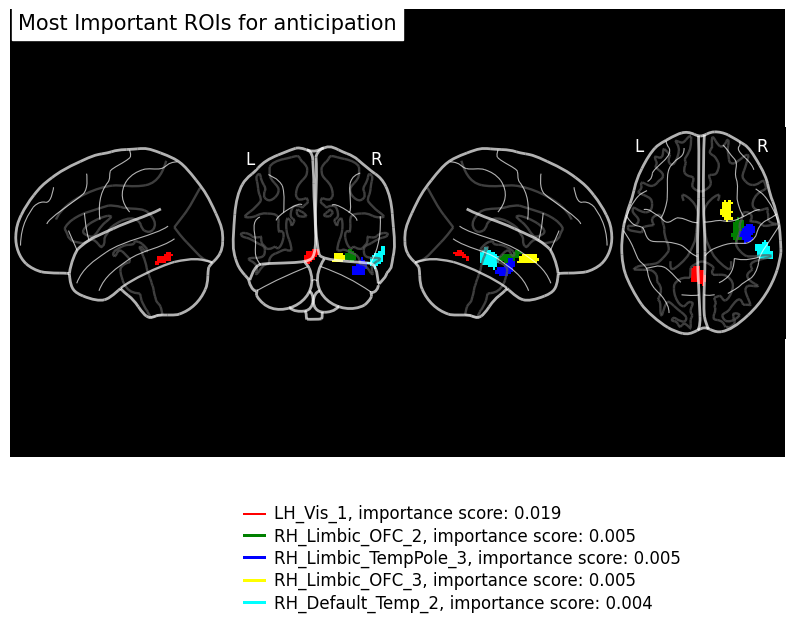}
        \caption{}
        \label{fig:RFR explanation anticipation}
    \end{subfigure}
    \caption{Explanation for Random Forrest Regressor for Joy and Anticipation}
    \label{fig:rfr joy and anticipation}
\end{figure}

Based on the model interpretability framework described in Section \ref{sec:explanation}, we extracted the top-5 ROIs contributing most substantially to emotion prediction for each emotion across five models: linear, Lasso and Ridge models, linear SVR, and multi-model RFR (MM-RFR). Firstly, we find that for a given model, the brain regions with the highest predictive importance varied across emotions. For instance, Fig. \ref{fig:rfr joy and anticipation} presents the top-5 ROIs with the strongest weights for predicting \textit{Joy} and \textit{Anticipation} in the MM-RFR model. Although both emotions shared the most influential ROI, their broader network-level importance patterns differed: Anticipation was characterized by strong weightings in three Limbic ROIs and one ROI from the Default Mode Network, whereas Joy relied on a more distributed networks. Moreover, no consistent top-contributing ROIs emerged across different models for the same emotion. For example, the MM-RFR and Ridge regression models identified completely distinct sets of ROIs as the primary feature drivers for predicting \textit{Fear}. Full explanations on top-5 most substrantial ROIs in predicting each emotion across different models can be found in \ref{appendix_roi_explanations}.

\subsubsection{DFC dataset}
As shown in Fig. \ref{fig:DFC_explanation_pipeline} (2) and (3), the explanations of DFC-based regression models are realized based on the MST representations constructed from the feature-importance matrices. Each MST provides a compact network structure that highlights the most informative functional interactions underlying each decoded emotion. 

\begin{figure}[htbp]
    \centering
    \begin{subfigure}[b]{0.45\textwidth}
        \includegraphics[width=\textwidth]{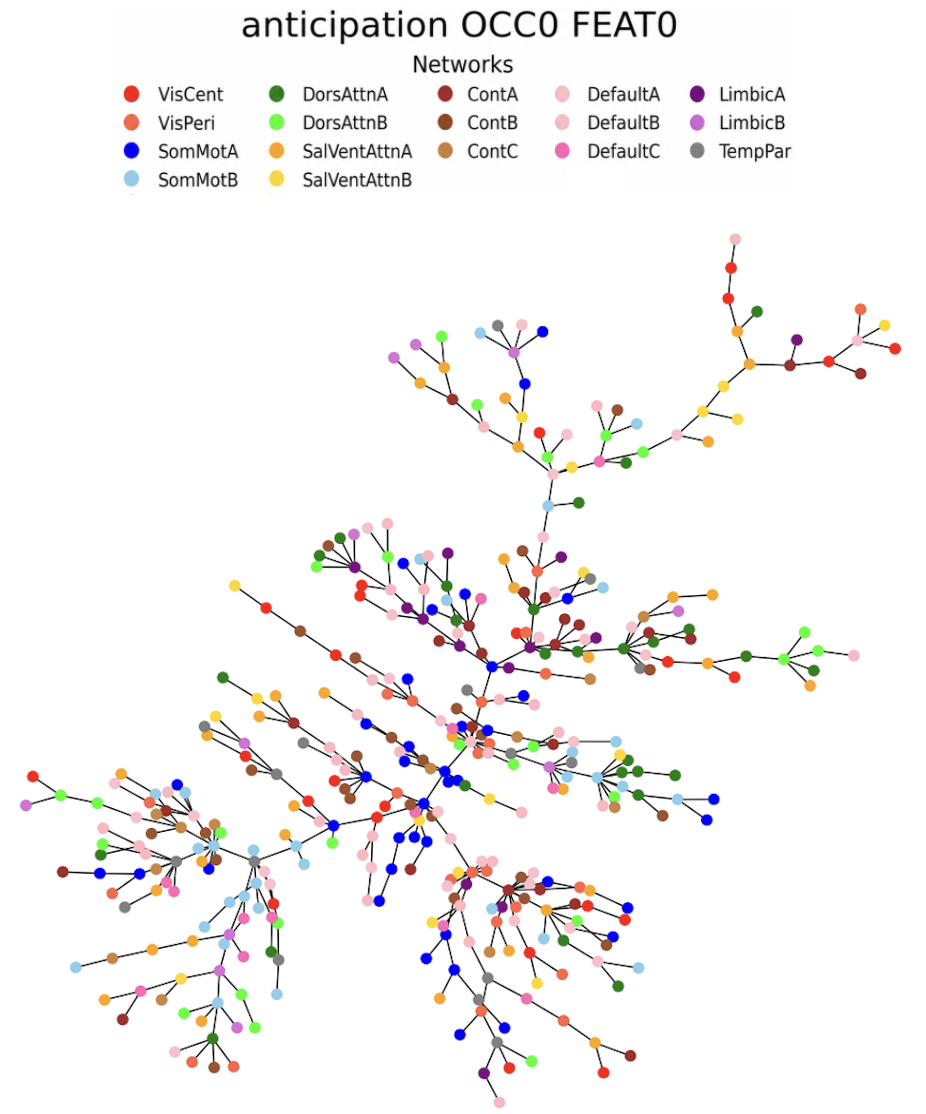}
        \caption{a) maximum spanning tree}
        \label{fig:ridge_17N_anticipation_graph}
    \end{subfigure}
    \hspace{0.05\textwidth}
    \begin{subfigure}[b]{0.45\textwidth}
        \includegraphics[width=\textwidth]{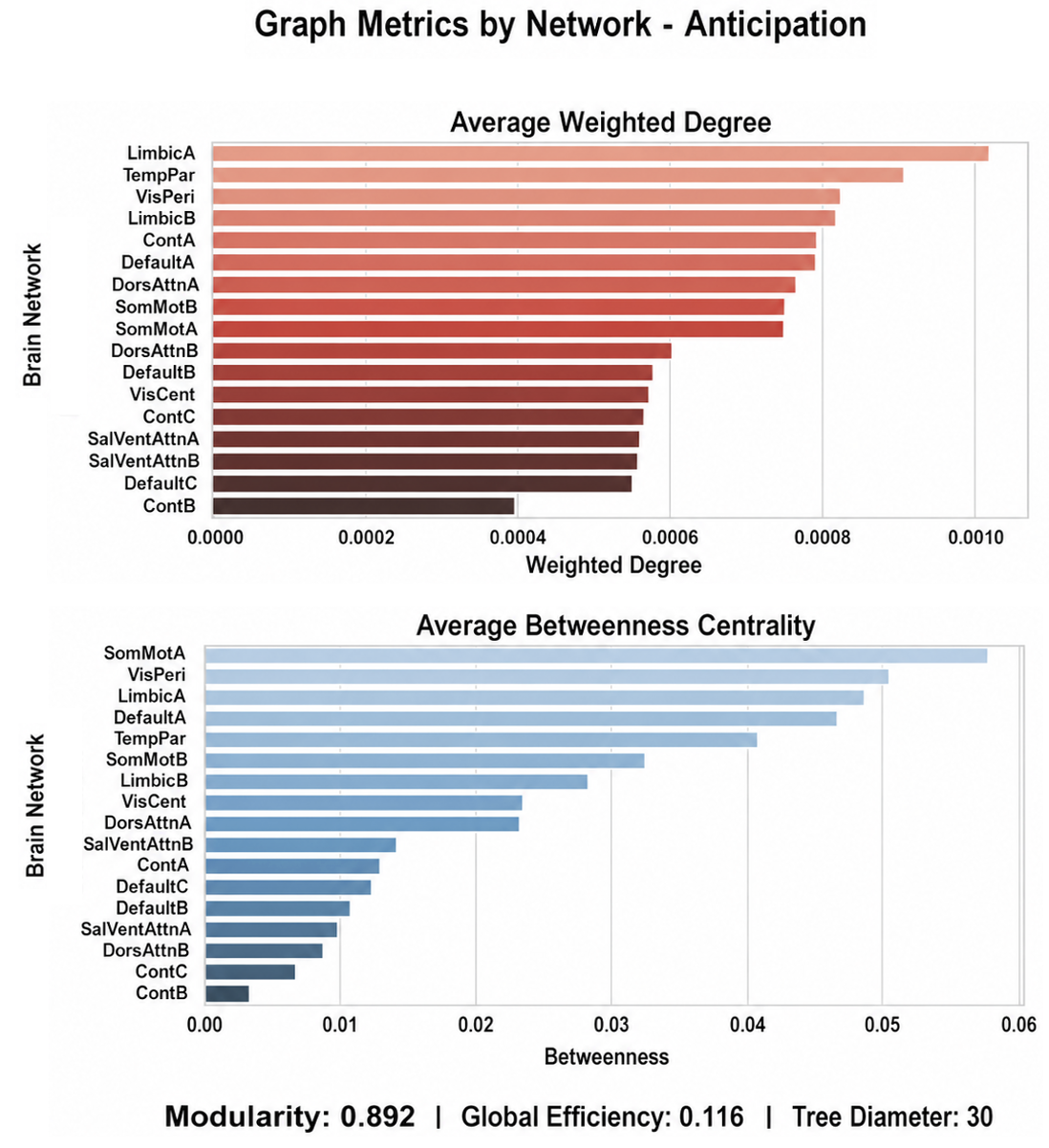}
        \caption{b) computed graph metrics}
        \label{fig:ridge_17N_anticipation_metrics}
    \end{subfigure}
    \\[1.0cm]
    \begin{subfigure}[b]{0.45\textwidth}
        \includegraphics[width=\textwidth]{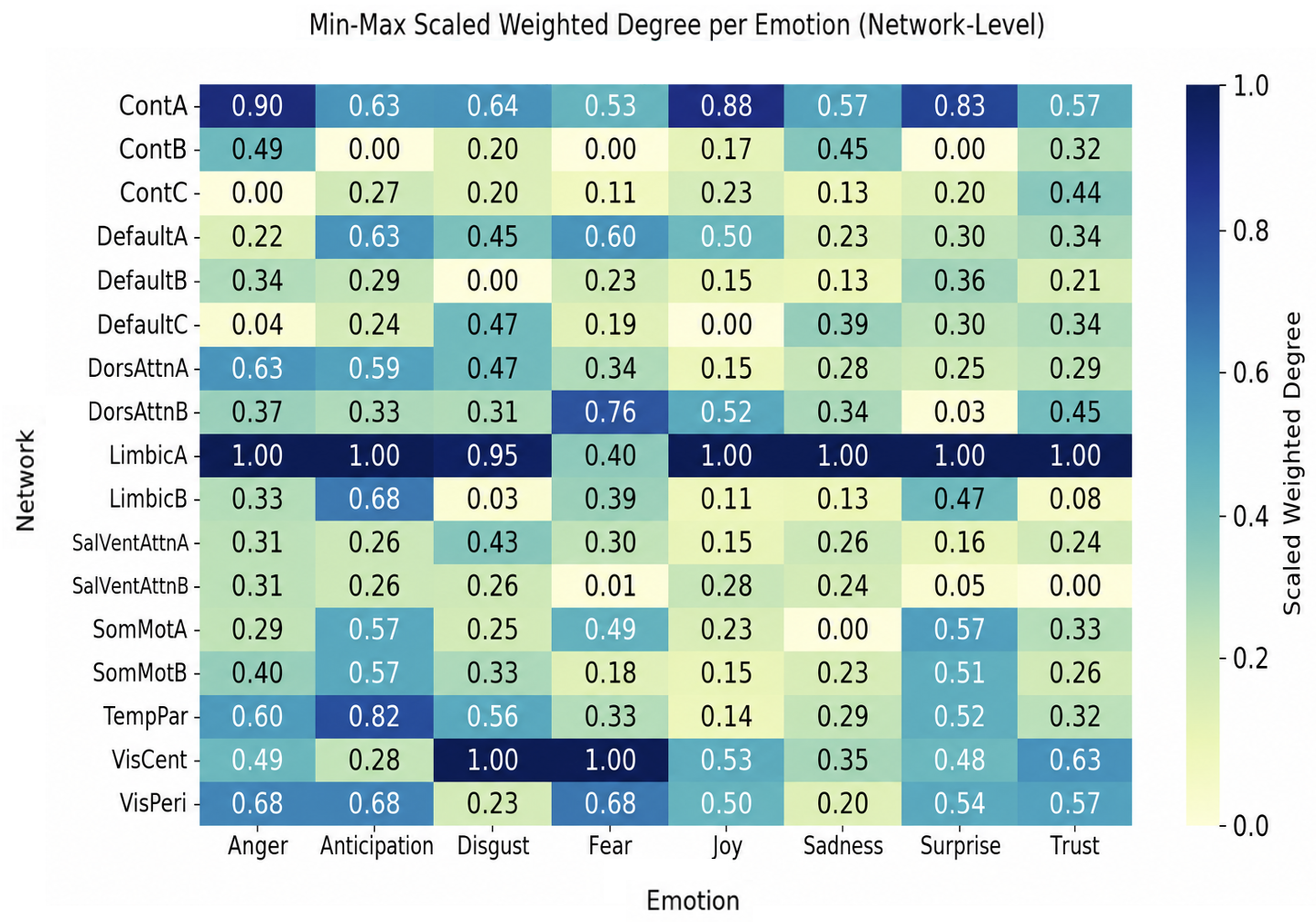}
        \caption{c) weighted degree}
        \label{fig:ridge_17N_anticipation_weighted}
    \end{subfigure}
    \hspace{0.05\textwidth}
    \begin{subfigure}[b]{0.45\textwidth}
        \includegraphics[width=\textwidth]{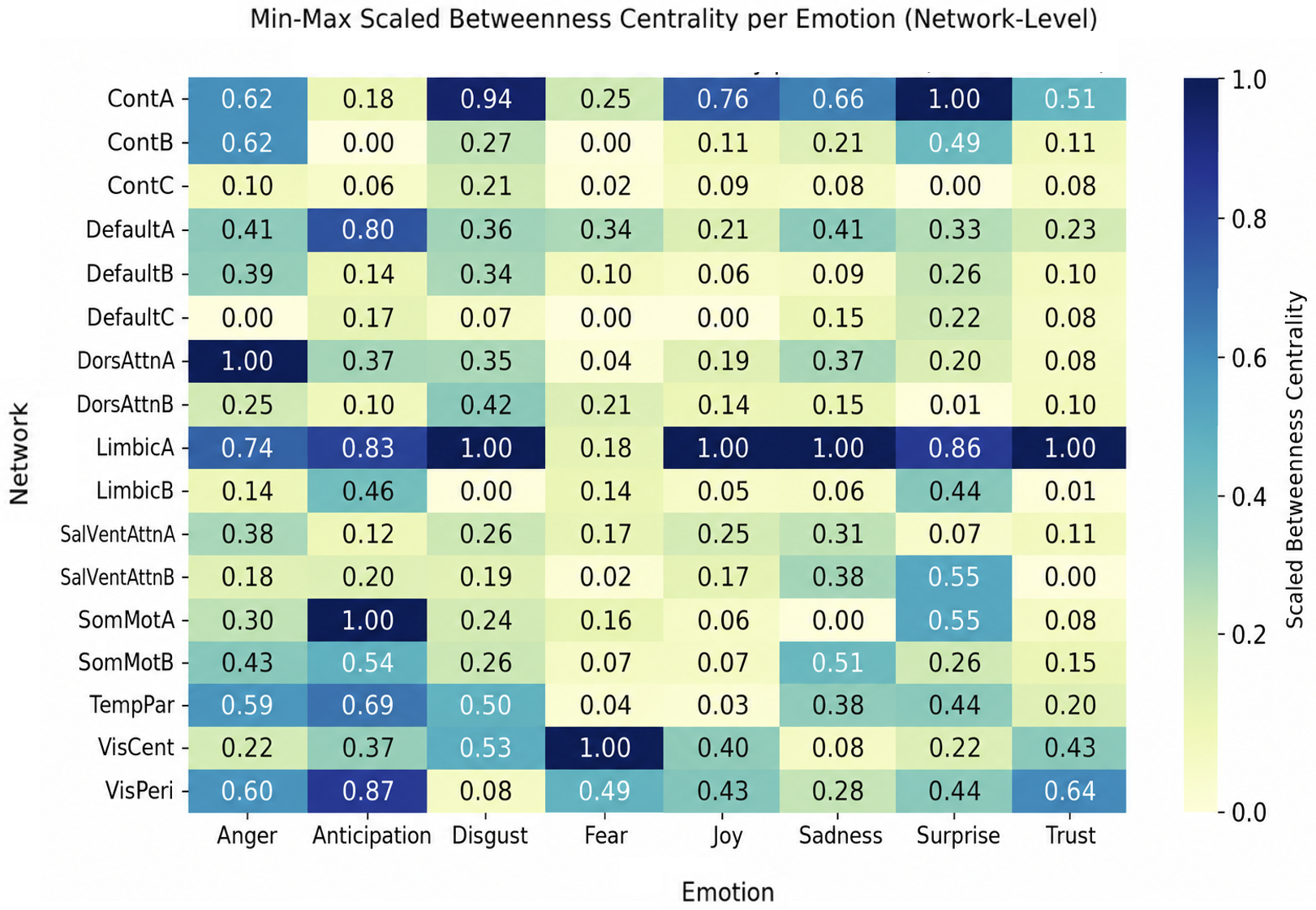}
        \caption{d) betweenness centrality}
        \label{fig:ridge_17N_anticipation_betweenness}
    \end{subfigure}
    \caption{Topological MST projection and graph metrics of the Ridge prediction for decoded \textit{Anticipation} using 17 networks.}
    \label{fig:DFC_SVM_anticipation}
\end{figure}

We trained five models on the DFC dataset, consistent with the ROI analyses, to enable comparative evaluation across feature types. Owing to page limit, we present the Ridge regression results based on the 17-network parcellation in the main text for two reasons. First, as shown in Table \ref{tab:transposed_performance}, Ridge achieved the second-best predictive performance in terms of MSE and $R^2$, following RBF-based SVR, with an MSE only 0.03\% higher than that of the best-performing model. Second, the brain network connectivity features identified by this model were broadly representative of those observed across the other models. Interpretability analyses of the remaining models trained on the DFC dataset, as well as analyses based on the 7-network parcellation, are provided in \ref{Appendix DFC explanations}. Specifically, our analysis is shown in terms of the following perspectives:

\textbf{Graph Visualization.} As shown in Fig. \ref{fig:ridge_17N_anticipation_graph}, MST visualizations were generated by coloring nodes according to their corresponding large-scale brain networks based on the 17-network parcellation. Graph layouts were optimized to reflect topological, rather than anatomical proximity in order to emphasize network structure. From the figure, we can see that the MST for \textit{Anticipation} exhibits a distributed yet modular organization, with several densely connected local hubs linked by longer inter-network branches. Default Mode, Salience/Ventral Attention, and Dorsal Attention networks occupy central bridging positions, suggesting an integrative role in coordinating information flow during anticipation. In contrast, Visual and Somatomotor networks are more frequently located at peripheral branches, indicating comparatively specialized or locally constrained contributions. Limbic nodes appear embedded within intermediate positions of the tree, consistent with their involvement in affective valuation during anticipatory states. Overall, the topology suggests that \textit{Anticipation} is supported by coordinated interactions between attention-control, default-mode, and affective systems rather than by a single dominant network.

\textbf{Network-Level Aggregation.} To identify which large-scale systems are most involved in each emotion, node-level metrics were aggregated at the network level. For every MST, the average weighted degree and betweenness centrality were computed within each 17-network parcelations, indicating which systems function as hubs (degree) and bridges (betweenness). We sorted the weighted degree and betweenness centrality values of the 17 brain networks for \textit{Anticipation} from highest to lowest. The results are presented in Fig. \ref{fig:ridge_17N_anticipation_metrics} to facilitate reference analysis with the MST. LimbicA shows the highest average weighted degree, followed by TempPar, VisPeri and LimbicB, indicating that nodes within these networks participate in more and/or stronger MST connections. This is consistent with the MST, where Limbic (purple/violet) nodes are embedded in central regions and connect widely to multiple systems, while Temporo-Parietal (gray) nodes extend along major branches that link distant modules. For average betweenness centrality, SomMotA ranks highest, followed by VisPeri, LimbicA and DefaultA, suggesting that these networks provide crucial bridging roles for information transfer across the tree. In the MST, SomMotA (dark blue) nodes frequently occupy branch points that connect different modules, and VisPeri (orange) and DefaultA (light pink) nodes also lie on key inter-network paths. Together, the MST and centrality analyses indicate that Anticipation relies on coordinated interactions among affective (Limbic), attention (DorsAttn, SalVentAttn), default-mode (DefaultA/B/C), and sensorimotor/association systems (SomMot, VisPeri, TempPar), Affective and temporo-parietal networks show high local involvement, whereas sensorimotor and visual/DMN networks play prominent bridging ROIs that integrate information across the brain.

Furthermore, Fig. \ref{fig:ridge_17N_anticipation_weighted} and \ref{fig:ridge_17N_anticipation_betweenness} present heatmaps of the aggregated the average weighted degree and betweenness centrality across all emotions under the Ridge model. All values are min–max scaled per emotion to highlight relative contributions, as absolute magnitudes vary substantially across emotions and models. As shown in Fig. \ref{fig:ridge_17N_anticipation_weighted}, LimbicA consistently exhibits the highest values across nearly all emotions, indicating a stable hub-like role with dense local connectivity during affective processing. Additional high-degree contributions are observed in ControlA, Visual Central, Visual Peripheral, and Temporal-Parietal networks, although their relative prominence varies by emotion; for example, Visual Central reaches maximal values for Disgust and Fear, whereas Temporal-Parietal is particularly elevated for Anticipation. As observed in Fig. \ref{fig:ridge_17N_anticipation_betweenness}, a more dynamic pattern emerges. LimbicA again shows consistently high values, suggesting a recurrent bridging role across emotional states, while several non-limbic networks become dominant for specific emotions: Dorsal Attention A peaks for Anger, Somatomotor A and Visual Peripheral for Anticipation, Visual Central for Fear, and ControlA for Disgust and Surprise. Overall, these findings indicate that although limbic circuitry maintains a stable central role, the broader functional organization of emotional processing is emotion-dependent, with distinct large-scale networks assuming specialized hub and bridge functions across affective states.

\begin{table}
\centering
\caption{Global graph metrics for each emotion using Ridge models.}
\footnotesize
\begin{tabular}{lccc}
\toprule
\textbf{Emotion} & \textbf{Modularity} & \textbf{Global Efficiency} & \textbf{Tree Diameter} \\
\midrule
Surprise & 0.8886 & 0.1161 & 27 \\
Sadness & 0.8907 & 0.1033 & 31 \\
Disgust & 0.8938 & 0.1002 & 37 \\
Anger & 0.8854 & 0.1104 & 32 \\
Fear & 0.8865 & 0.1194 & 25 \\
Anticipation & 0.8921 & 0.1164 & 30 \\
Joy & 0.8885 & 0.1255 & 24 \\
Trust & 0.8925 & 0.1091 & 26 \\
\bottomrule
\end{tabular}
\label{tab:global_metrics_svr+ridge}
\end{table}

\textbf{Global Graph Properties.} Each MST was further characterized using three global graph measures: modularity, capturing the extent to which the tree separates into internally cohesive but externally sparse communities; global efficiency, reflecting how effectively information can be integrated across the network; and tree diameter, defined as the longest shortest path between any two nodes. The Ridge model results are reported in Table \ref{tab:global_metrics_svr+ridge}. We can see from the table that the global graph metrics reveal distinct topological signatures across emotional valences. Negative emotions—specifically Disgust, Sadness, and Anger—are characterized by higher modularity, lower global efficiency, and larger tree diameters, indicating a more segregated, less integrated, and spatially extended network architecture. In contrast, positive emotions such as Joy and Trust exhibit lower modularity, higher global efficiency, and smaller tree diameters, reflecting a more compact and efficiently integrated network topology. Disgust serves as the extreme case within the negative valence, demonstrating the highest modularity (0.8938), the lowest global efficiency (0.1002), and the largest tree diameter (37), while Joy represents the opposite pole with the highest global efficiency (0.1255) and smallest diameter (24). Anticipation presents a hybrid profile, sharing the high modularity and extended diameter typical of negative emotions while maintaining a mid-range global efficiency that sits between the positive and negative extremes. Collectively, these findings suggest that under the Ridge model, the topological organization of functional brain networks during emotional processing scales systematically with valence, with negative emotions driving network segregation and spatial elongation while positive emotions promote integration and compactness.

\begin{figure}
    \centering
    \begin{subfigure}[b]{0.45\textwidth}
        \includegraphics[width=\textwidth,height=172pt]{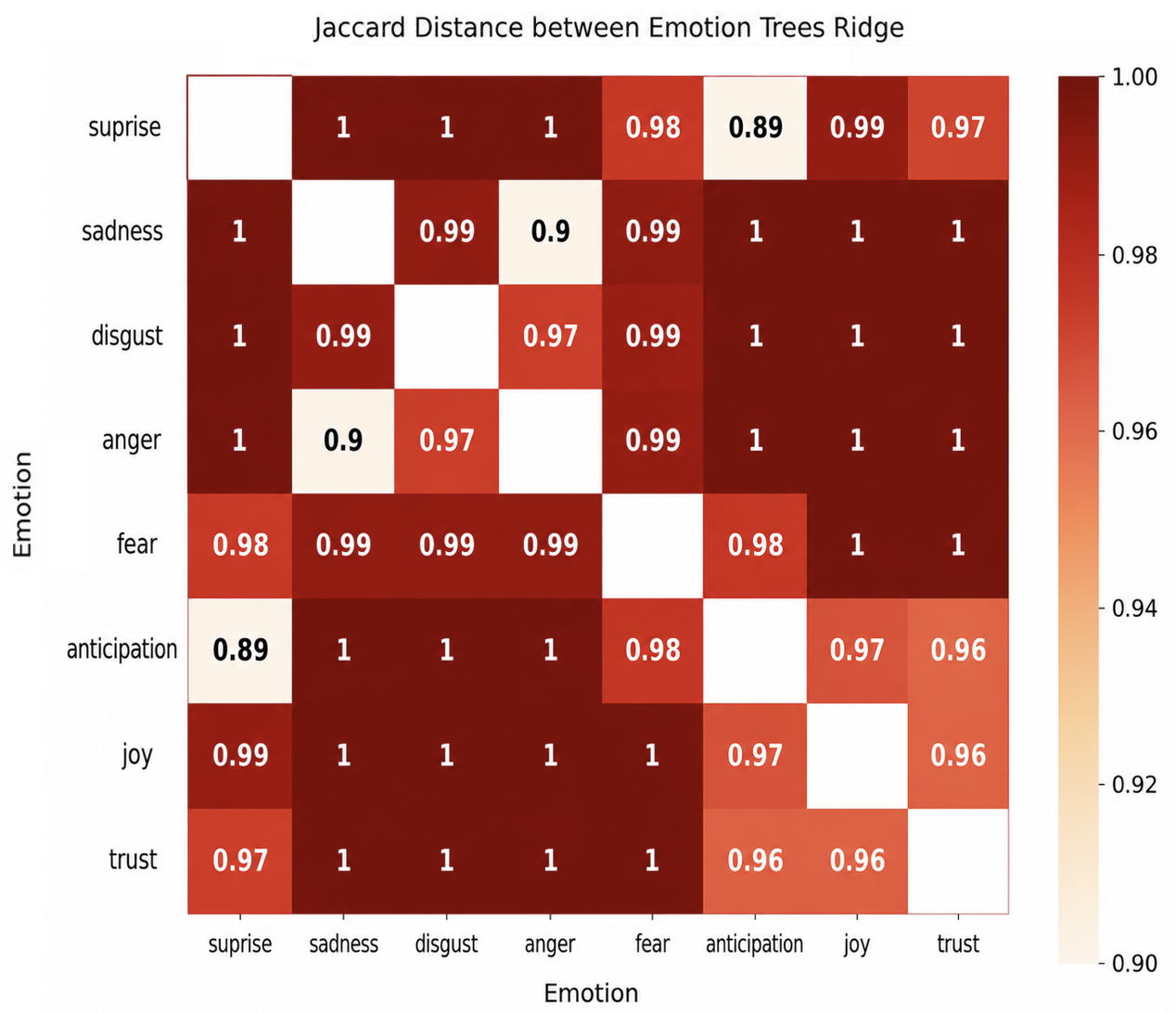}
        \caption{(a) Heatmap for Jaccard Distances, Ridge}
        \label{fig:dfc_jaccard_heatmap}
    \end{subfigure}
    \hspace{0.05\textwidth}
    \begin{subfigure}[b]{0.45\textwidth}
        \includegraphics[width=\textwidth]{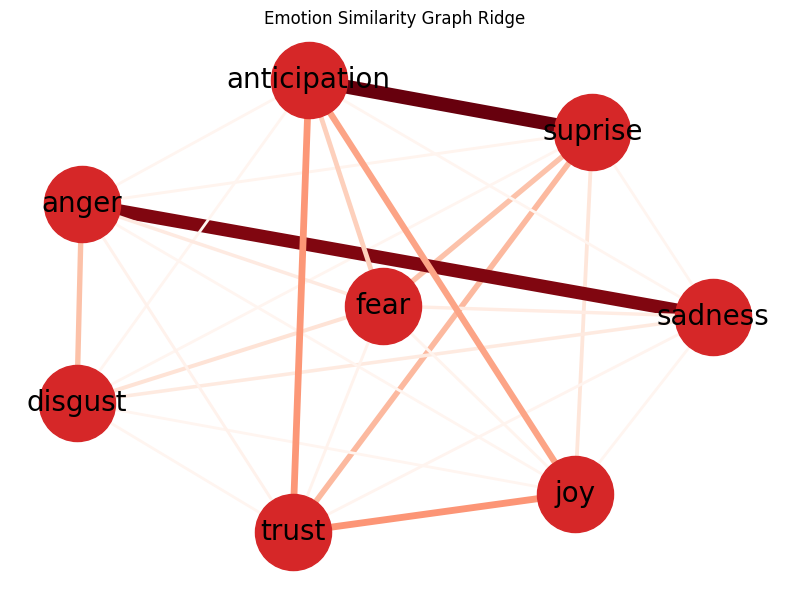}
        \caption{(b) Jaccard Distances, Ridge}
        \label{fig:dfc_jaccard_plot}
    \end{subfigure}
    \caption{Heatmaps and graph visualizations of Jaccard distances between emotional trees for the Ridge model. (a) The heatmap illustrates pairwise distances across emotions; (b) the graph depict the corresponding emotional similarity structures with thicker edges indicating stronger topological similarity between emotions.}
    \label{fig:dfc_jaccard}
\end{figure}

\textbf{Inter-Emotion Similarity.} To assess how structurally distinct each emotional MST is from the others, we performed pairwise comparisons using the \textit{Weighted Jaccard distance}, which quantifies the overlap of edge weights between trees. This metric reflects the degree of topological similarity between emotions. The resulting distance matrix is presented in Fig. \ref{fig:dfc_jaccard_heatmap}, which shows uniformly high similarity values across nearly all emotion pairs (range between 0.89–1.00), indicating substantial overlap in the edge composition of the MSTs across emotional states. Most pairwise comparisons approach maximal similarity, suggesting that the core functional connectivity scaffold identified by the Ridge model is largely preserved across emotions. Relatively lower similarity values are observed for pairs involving surprise and anticipation (e.g., 0.89), implying modestly more distinct topological configurations for these states. The network visualization in Fig. \ref{fig:dfc_jaccard_plot} further summarizes these relationships, where thicker and darker edges denote greater similarity between emotion pairs. Strong links are evident among clusters such as anger–sadness, anticipation–surprise, and joy–trust, whereas weaker links reflect comparatively differentiated connectivity structures. Overall, the figure suggests that emotion-specific brain connectivity patterns share a highly conserved global architecture, with only subtle reconfigurations distinguishing individual affective states.

\section{Discussion}
In this study, we investigated how classical regression models, applied to ROI-based and DFC-based two distinct neural representations, perform in decoding dynamic emotional states from fMRI data. We systematically evaluated their predictive accuracy, the neuroscientific interpretability of their feature importance, and the theoretical implications, namely Locationist and Constructionist theories of emotions in relation to these results.

\subsection{ROI VS. DFC Performance}
A comparative analysis of the linear models and the nonlinear models used in this paper across both datasets reveals distinct performance trade-offs. Although the nonlinear model—namely, RBF-based SVR—achieved the highest predictive accuracy overall, the linear models demonstrated highly competitive performance on the DFC dataset. Specifically, the Ridge model effectively captured emotional variance trends while yielding an MSE marginally higher than the best nonlinear model by a negligible 0.0003. Conversely, for the ROI dataset, while the models produced moderate MSE metrics, the corresponding $R^2$ scores indicate an inability to accurately track underlying data trends. While a constrained sample size likely contributed to this suboptimal performance, the linear models ultimately exhibited superior robustness when predicting functional brain region connectivity.

Furthermore, the superior performance of DFC-based decoding carries a vital theoretical implication: it suggests that fine-grained emotional states are more effectively captured by the temporally evolving patterns of connectivity between brain networks than by the static amplitude of activity within predefined anatomical regions. The finding that the systematic exclusion of specific network-level ROIs failed to significantly impact performance further underscores that emotion-related information is distributed across dynamic interactions rather than localized within isolated areas. This pattern of results provides compelling empirical support for psychological constructionist frameworks of emotion, as opposed to strictly locationist accounts.

\subsection{Explanation Comparison on ROI vs DFC Representations}
The models exhibited markedly different interpretability characteristics across datasets. First, the ROI-based explanations did not reveal a clear or consistent neural signature for specific emotions across models. This may reflect limitations in the models’ ability to capture the necessary feature transformations, insufficient discriminative information within ROI-based representations themselves, or a combination of both factors. Nonetheless, several noteworthy patterns emerged. For example, emotion-specific analyses of feature importance revealed network-level clusters for certain emotions and models. The RFR assigned high importance to ROIs within the Limbic Network for \textit{Anticipation} and \textit{Trust}—a network that, according to \cite{Schaefer2018}, includes the orbitofrontal cortex, a region closely linked to emotion processing \citep{rolls2019limbic}. This suggests that the RFR may be sensitive to emotion-related activity in this region. Additionally, the RFR consistently identified a specific ROI within the Left Hand Visual Network as predictive across all emotions. Although the underlying reason remains unclear, prior work \citep{Xu2023} has shown that interactions between sensory and higher-order networks can be informative for decoding emotions from visual stimuli, which may partly explain this recurring pattern.

In the interpretive analysis of DFC data, MST-based graph analysis revealed distinct, emotion-specific configurations in both local and global topological properties. Although the feature weight distributions across emotions exhibited varying degrees of dispersion, this variance is largely an artifact of the specific modeling methodologies. For instance, Ridge regression captured broader, more diffuse network involvement, whereas Linear SVR identified highly localized and focused emotional patterns. Crucially, despite these algorithmic variations, the ranking of brain network involvement maintained notable consistency across specific emotions, yielding several key observations.

\textbf{The temporal pole, limbic, visual, and frontoparietal control networks consistently emerged as central hubs of emotion-related brain activity across both models and emotional states}. Across all models and parcellations, the limbic network emerged most reliably as a core structure for nearly all emotional states. At nodal level, the left temporal pole consistently ranked among the top nodes in weighted degree and betweenness centrality. These findings align with established literature characterizing the temporal pole as a critical locus for integrating social, semantic, and affective signals \citep{olson2007temporalpole}. Its prominence across diverse affective states and models reinforces its role as a convergence zone for emotionally laden narrative comprehension. Notably, visual processing regions—specifically encompassing both the Central Visual Network (VisCent) and the Peripheral Visual Network (VisPeri)—exhibited marked prominence across most emotions, particularly during surprise, sadness, and joy. Given the absence of direct visual input in this paradigm, this network involvement likely reflects narrative-induced mental imagery and internal scene simulation. This interpretation is consistent with previous evidence demonstrating that visual cortices are robustly recruited during emotionally salient imagined experiences \citep{klasen2011emotion}. The high metric values of Control A (dorsolateral prefrontal cortex and parts of the superior parietal lobule) across all emotions support its potential role as a common cognitive hub. In addition to the limbic and visual networks, Control A may contribute to stabilizing executive function. While limbic areas are likely involved in affective valuation and visual networks in mental imagery and scene construction, the consistent centrality of Control A suggests that it may help align internal models with dynamic narrative input, thereby providing regulatory oversight and scaffolding for emotional experience. This interpretation aligns with previous findings on the role of the Control network in emotion processing \citep{morawetz2016emotion}.

\textbf{Certain emotions also demonstrated distinct configurations of network-level involvement}. \textit{Anticipation}, for instance, engaged a robust co-activation of the somatomotor and limbic networks. This specific coupling suggests a preparatory neural state that pairs emotional arousal with somatic readiness, potentially reflecting motor simulation or internal action representation. Similarly, surprise recruited the somatomotor network, which may reflect rapid sensory-motor gating in response to unexpected narrative shifts. This motor readiness aligns with the immediate physical adjustments inherent to the cognitive processing of unexpected stimuli \citep{Grabenhorst2025}.

\textbf{Global topological metrics, including modularity and efficiency, also exhibited emotion-specific variations.} \textit{Trust}, \textit{Anticipation}, and \textit{Joy} demonstrated high modularity across different modeling approaches. This structural organization suggests that positive or future-oriented emotional states are characterized by enhanced network segregation, potentially reflecting heightened top-down cognitive control and modular specialization during the processing of social or goal-directed affect. Conversely, patterns of global efficiency displayed greater model-dependent variability. For instance, within the SVR framework, \textit{Anticipation} and \textit{Disgust} yielded the highest global efficiency, whereas \textit{Joy} emerged as the most efficient state in the Ridge regression model. Interestingly, \textit{Trust} consistently exhibited the lowest global efficiency across both models. This diminished efficiency may reflect the highly distributed, non-localized neural architecture required to process complex, socially embedded emotional states. Methodological divergences were also apparent in the MST diameter, where the Ridge model captured more pronounced variance than the SVR model. Specifically, negative emotions such as \textit{Disgust}, \textit{Anger}, and \textit{Sadness} were characterized by larger diameters, whereas \textit{Joy} and \textit{Trust} exhibited more compact topologies. However, the absence of a convergent trend among different mathematical frameworks suggests that tree diameter may be highly sensitive to the modeling constraints of the classifier, rendering it a less robust metric for differentiating discrete emotional states in this specific context. Inter-emotion similarity analysis revealed a distinct topological convergence between \textit{Anticipation} and \textit{Surprise}, as well as between \textit{Anger} and \textit{Sadness}. \textit{Anticipation} and \textit{Surprise} exhibited marked structural similarities across both models, likely driven by their shared recruitment of the Somatomotor and Visual networks involved in somatic readiness and heightened sensory processing. This commonality suggests that these two affective states share overlapping neural mechanisms dedicated to preparing for and adapting to dynamic environmental stimuli. Similarly, \textit{Anger} and \textit{Sadness} demonstrated substantial structural overlap, potentially reflecting a shared reliance on Limbic and Control networks mediated by negative affect and cognitive emotion regulation. Intriguingly, despite also sharing a negative valence, \textit{Fear} and \textit{Disgust} did not exhibit comparable topological overlap. This divergence may be attributed to their dependence on highly specialized, stimulus-specific subcortical and insular systems. For instance, \textit{Fear} processing heavily engages the amygdala for rapid threat detection and adaptive survival responses \citep{ledoux_amygdala}, whereas \textit{Disgust} is associated with the anterior insula, a region critical for the evaluation of bodily sensations \citep{phan_insula_disgust}. These specialized circuits likely engender distinct functional connectivity patterns that diverge from the more regulatory and evaluative networks characteristically observed in \textit{Anger} and \textit{Sadness}. Crucially, it must be noted that subcortical structures, including the amygdala, were not captured in this analysis, as the Schaefer brain atlas utilized herein is strictly restricted to cortical parcellations.

\subsection{Locationist vs. Constructionist Hypythesis}
Because the current analysis does not provide definitive evidence to support locationist theories of emotion, a conclusive theoretical determination cannot be made at this time. Instead, our findings point toward a highly distributed, dynamic neural architecture wherein emotional states are more accurately predicted by time-varying connectivity trajectories than by isolated, regional activations. The consistent predictive superiority of the DFC-based models strongly reinforces this perspective, supporting the view that emotions manifest as emergent properties of large-scale interacting brain systems—a core tenet of the psychological constructionist framework. Furthermore, our comparative model analysis offers programmatic insight into the geometric nature of this emotional neural code. The profound overfitting exhibited by unregularized ordinary least squares (OLS) linear models on the DFC data, contrasted with the robust generalization achieved by Ridge-based regularized linear and kernel-based SVR approaches, indicates that the underlying predictive signal is embedded within high-dimensional, multivariate covariance structures rather than a sparse set of independent regional features. This interpretation is further underscored by the divergent performance profiles of our ensemble methods. While Random Forest architectures achieved adequate generalization on the lower-dimensional ROI metrics, they failed to generalize effectively on the high-dimensional DFC configurations. This discrepancy highlights a fundamental structural limitation of standard tree-based algorithms, which lack intrinsic inductive biases to model the high-dimensional relational dependencies and continuous temporal dynamics inherent to functional connectivity manifolds.

\subsection{Limitations}
Several methodological and conceptual limitations of the current study should be acknowledged, as they provide critical context for interpreting our findings and define clear trajectories for future research.

\textbf{Dataset and Generalizability.} The empirical findings presented herein are derived from a single, naturalistic fMRI dataset characterized by a modest sample size. While this dataset provides high temporal resolution and ecological validity, the limited sample constrains the statistical power and generalizability of the observed decoding patterns and emotion-specific network configurations. Future studies employing larger, more demographically diverse cohorts and multi-modal data are needed to confirm the robustness and broader applicability of these results.

\textbf{Label Validity and Semantic Confounds.} A primary constraint involves the operationalization of our target variables. The emotional labels were computationally derived from narrative content via a Large Language Model (LLM) rather than captured directly through participant self-report. Although this algorithmic approach enables highly granular, continuous annotation, it introduces a degree of uncertainty regarding the ground-truth subjective experiences of the participants. Furthermore, this design introduces a potential semantic confound, as the machine learning models may partially decode the underlying narrative structure or thematic text features that co-vary with, but are distinct from, actual neurobiological affective states.

\textbf{Representational Constraints.} The neural feature space was subject to several technical constraints. First, utilizing a strictly cortical parcellation (the Schaefer atlas) effectively omitted subcortical structures—most notably the amygdala and hippocampus—that are fundamentally central to affective processing. Consequently, our models lacked access to these critical subcortical signaling pathways. Second, our DFC estimations relied on a fixed 24-second sliding window. This choice reflects a classic trade-off between temporal resolution and statistical reliability, meaning our framework may not optimally capture emotional phenomena operating on faster or more protracted temporal scales. Finally, our graph-theoretical interpretations were restricted to MSTs; while MSTs successfully isolate the core backbone of a network, they intentionally discard lower-weight edges that may still contain informative connectivity architecture.

\textbf{Modeling Assumptions.} Our analytical pipeline utilized classical regularized and kernel-based regression frameworks (Ridge and SVR), which enforce specific linear or geometric mappings between connectivity features and affective states. While these architectures demonstrated robust performance, they are inherently limited in their ability to capture highly non-linear, hierarchical representations of emotional processing. Additionally, using feature importance metrics for model interpretability implicitly assumes a degree of feature independence, a mathematical assumption that is frequently violated by the highly collinear, multivariate nature of macroscopic brain connectivity.

\textbf{Subject-Level Variability.} This study focused primarily on characterizing group-level network patterns, which potentially obscures meaningful idiosyncratic variations in functional anatomy, emotional reactivity, and cognitive strategy. Future investigations should leverage hierarchical linear modelling or personalized, subject-specific decoding frameworks to more effectively disentangle within-subject dynamics from between-subject variance.

\section{Conclusion}
his study demonstrates that classical regression frameworks—particularly regularized SVR—can effectively decode dynamic emotional states from fMRI data when leveraging dynamic functional connectivity (DFC) representations. The consistent predictive superiority of DFC-based models over static region-of-interest (ROI) summaries indicates that affective information is preferentially encoded within time-varying network interactions rather than isolated regional activity. Furthermore, feature importance analyses yielded highly interpretable, theoretically aligned neurobiological signatures, highlighting critical integrative hubs such as the temporal pole alongside distinct network configurations unique to specific emotional states. Collectively, these findings provide compelling empirical support for the psychological constructionist view of emotions as emergent states arising from the coordinated orchestration of distributed, dynamic brain networks, thereby directly challenging strictly locationist accounts of human affect.
 
\section{Supplementary Material}
The fully documented version control repository containing this codebase is publicly accessible on GitHub [\url{https://github.com/HansDahleKvadsheim/Emotional-Decoding}].

\printbibliography

@article{santavirta2025gpt,
  title={GPT-4 accurately predicts human emotions and their neural correlates},
  author={Santavirta, Severi and Suominen, Lauri and Wu, Yuhang and Sander, David and Nummenmaa, Lauri},
  journal={bioRxiv},
  pages={2025--09},
  year={2025},
  publisher={Cold Spring Harbor Laboratory}
}

@article{vos2025decoding,
  title={Decoding neural emotion patterns through large language model embeddings},
  author={Vos, Gideon and Ebrahimpour, Maryam and Van Eijk, Liza and Sarnyai, Zoltan and Azghadi, Mostafa Rahimi},
  journal={Neurocomputing},
  pages={132513},
  year={2025},
  publisher={Elsevier}
}

@article{farahani2022explainable,
  title={Explainable AI: A review of applications to neuroimaging data},
  author={Farahani, Farzad V and Fiok, Krzysztof and Lahijanian, Behshad and Karwowski, Waldemar and Douglas, Pamela K},
  journal={Frontiers in Neuroscience},
  volume={16},
  pages={906290},
  year={2022},
  publisher={Frontiers Media SA}
}

@inproceedings{borriero2024explainable,
  title={Explainable emotion decoding for human and computer vision},
  author={Borriero, Alessio and Milazzo, Martina and Diano, Matteo and Orsenigo, Davide and Villa, Maria Chiara and DiFazio, Chiara and Tamietto, Marco and Perotti, Alan},
  booktitle={World Conference on Explainable Artificial Intelligence},
  pages={178--201},
  year={2024},
  organization={Springer}
}

@article{scaling_laws,
  title={Scaling laws for neural language models},
  author={Kaplan, Jared and McCandlish, Sam and Henighan, Tom and Brown, Tom B and Chess, Benjamin and Child, Rewon and Gray, Scott and Radford, Alec and Wu, Jeffrey and Amodei, Dario},
  journal={arXiv preprint arXiv:2001.08361},
  year={2020},
  url = {https://arxiv.org/pdf/2001.08361}
}

@article{few_shot,
  title={Language models are few-shot learners},
  author={Brown, Tom B},
  journal={arXiv preprint arXiv:2005.14165},
  year={2020},
  url={https://arxiv.org/pdf/2005.14165}
}

@book{Plutchik1980,
  author    = {Robert Plutchik},
  title     = {Emotion: A Psychoevolutionary Synthesis},
  year      = {1980},
  publisher = {Harper \& Row},
  address   = {New York},
}

@article{fMRI,
  title={Functional Magnetic Resonance Imaging (fMRI): An Invaluable Tool in Translational Neuroscience},
  author={Lori A. Whitten},
  journal={RTI Press publication No. OP-0010-1212. Research Triangle Park, NC: RTI Press},
  year={2012},
  url={https://www.ncbi.nlm.nih.gov/books/NBK538909/pdf/Bookshelf_NBK538909.pdf}
}

@article{Schaefer2018,
  author    = {Alexander Schaefer and others},
  title     = {Local-Global Parcellation of the Human Cerebral Cortex from Intrinsic Functional Connectivity MRI},
  journal   = {Cerebral Cortex},
  year      = {2018},
  volume    = {28},
  number    = {9},
  pages     = {3095-3114},
  doi       = {10.1093/cercor/bhx179},
}

@article{saarimaki2022,
  author    = {Saarimäki, H. and Glerean, E. and Smirnov, D. and Mynttinen, H. and Jääskeläinen, I.P. and Sams, M. and Nummenmaa, L.},
  title     = {Classification of emotion categories based on functional connectivity patterns of the human brain},
  journal   = {NeuroImage},
  volume    = {247},
  article   = {118800},
  year      = {2022},
  doi       = {10.1016/j.neuroimage.2021.118800}
}

@article{Xu2023,
  author    = {Xu, S. and Zhang, Z. and Li, L. and Zhou, Y. and Lin, D. and Zhang, M. and Zhang, L. and Huang, G. and Liu, X. and Becker, B. and others},
  title     = {Functional connectivity profiles of the default mode and visual networks reflect temporal accumulative effects of sustained naturalistic emotional experience},
  journal   = {NeuroImage},
  volume    = {269},
  article   = {119941},
  year      = {2023},
  doi       = {10.1016/j.neuroimage.2023.119941}
}

@misc{openai2024gpt4technicalreport,
      title={GPT-4 Technical Report}, 
      author={OpenAI and Josh Achiam and Steven Adler and Sandhini Agarwal and Lama Ahmad and Ilge Akkaya and Florencia Leoni Aleman and Diogo Almeida and Janko Altenschmidt and Sam Altman and Shyamal Anadkat and Red Avila and Igor Babuschkin and Suchir Balaji and Valerie Balcom and Paul Baltescu and Haiming Bao and Mohammad Bavarian and Jeff Belgum and Irwan Bello and Jake Berdine and Gabriel Bernadett-Shapiro and Christopher Berner and Lenny Bogdonoff and Oleg Boiko and Madelaine Boyd and Anna-Luisa Brakman and Greg Brockman and Tim Brooks and Miles Brundage and Kevin Button and Trevor Cai and Rosie Campbell and Andrew Cann and Brittany Carey and Chelsea Carlson and Rory Carmichael and Brooke Chan and Che Chang and Fotis Chantzis and Derek Chen and Sully Chen and Ruby Chen and Jason Chen and Mark Chen and Ben Chess and Chester Cho and Casey Chu and Hyung Won Chung and Dave Cummings and Jeremiah Currier and Yunxing Dai and Cory Decareaux and Thomas Degry and Noah Deutsch and Damien Deville and Arka Dhar and David Dohan and Steve Dowling and Sheila Dunning and Adrien Ecoffet and Atty Eleti and Tyna Eloundou and David Farhi and Liam Fedus and Niko Felix and Simón Posada Fishman and Juston Forte and Isabella Fulford and Leo Gao and Elie Georges and Christian Gibson and Vik Goel and Tarun Gogineni and Gabriel Goh and Rapha Gontijo-Lopes and Jonathan Gordon and Morgan Grafstein and Scott Gray and Ryan Greene and Joshua Gross and Shixiang Shane Gu and Yufei Guo and Chris Hallacy and Jesse Han and Jeff Harris and Yuchen He and Mike Heaton and Johannes Heidecke and Chris Hesse and Alan Hickey and Wade Hickey and Peter Hoeschele and Brandon Houghton and Kenny Hsu and Shengli Hu and Xin Hu and Joost Huizinga and Shantanu Jain and Shawn Jain and Joanne Jang and Angela Jiang and Roger Jiang and Haozhun Jin and Denny Jin and Shino Jomoto and Billie Jonn and Heewoo Jun and Tomer Kaftan and Łukasz Kaiser and Ali Kamali and Ingmar Kanitscheider and Nitish Shirish Keskar and Tabarak Khan and Logan Kilpatrick and Jong Wook Kim and Christina Kim and Yongjik Kim and Jan Hendrik Kirchner and Jamie Kiros and Matt Knight and Daniel Kokotajlo and Łukasz Kondraciuk and Andrew Kondrich and Aris Konstantinidis and Kyle Kosic and Gretchen Krueger and Vishal Kuo and Michael Lampe and Ikai Lan and Teddy Lee and Jan Leike and Jade Leung and Daniel Levy and Chak Ming Li and Rachel Lim and Molly Lin and Stephanie Lin and Mateusz Litwin and Theresa Lopez and Ryan Lowe and Patricia Lue and Anna Makanju and Kim Malfacini and Sam Manning and Todor Markov and Yaniv Markovski and Bianca Martin and Katie Mayer and Andrew Mayne and Bob McGrew and Scott Mayer McKinney and Christine McLeavey and Paul McMillan and Jake McNeil and David Medina and Aalok Mehta and Jacob Menick and Luke Metz and Andrey Mishchenko and Pamela Mishkin and Vinnie Monaco and Evan Morikawa and Daniel Mossing and Tong Mu and Mira Murati and Oleg Murk and David Mély and Ashvin Nair and Reiichiro Nakano and Rajeev Nayak and Arvind Neelakantan and Richard Ngo and Hyeonwoo Noh and Long Ouyang and Cullen O'Keefe and Jakub Pachocki and Alex Paino and Joe Palermo and Ashley Pantuliano and Giambattista Parascandolo and Joel Parish and Emy Parparita and Alex Passos and Mikhail Pavlov and Andrew Peng and Adam Perelman and Filipe de Avila Belbute Peres and Michael Petrov and Henrique Ponde de Oliveira Pinto and Michael and Pokorny and Michelle Pokrass and Vitchyr H. Pong and Tolly Powell and Alethea Power and Boris Power and Elizabeth Proehl and Raul Puri and Alec Radford and Jack Rae and Aditya Ramesh and Cameron Raymond and Francis Real and Kendra Rimbach and Carl Ross and Bob Rotsted and Henri Roussez and Nick Ryder and Mario Saltarelli and Ted Sanders and Shibani Santurkar and Girish Sastry and Heather Schmidt and David Schnurr and John Schulman and Daniel Selsam and Kyla Sheppard and Toki Sherbakov and Jessica Shieh and Sarah Shoker and Pranav Shyam and Szymon Sidor and Eric Sigler and Maddie Simens and Jordan Sitkin and Katarina Slama and Ian Sohl and Benjamin Sokolowsky and Yang Song and Natalie Staudacher and Felipe Petroski Such and Natalie Summers and Ilya Sutskever and Jie Tang and Nikolas Tezak and Madeleine B. Thompson and Phil Tillet and Amin Tootoonchian and Elizabeth Tseng and Preston Tuggle and Nick Turley and Jerry Tworek and Juan Felipe Cerón Uribe and Andrea Vallone and Arun Vijayvergiya and Chelsea Voss and Carroll Wainwright and Justin Jay Wang and Alvin Wang and Ben Wang and Jonathan Ward and Jason Wei and CJ Weinmann and Akila Welihinda and Peter Welinder and Jiayi Weng and Lilian Weng and Matt Wiethoff and Dave Willner and Clemens Winter and Samuel Wolrich and Hannah Wong and Lauren Workman and Sherwin Wu and Jeff Wu and Michael Wu and Kai Xiao and Tao Xu and Sarah Yoo and Kevin Yu and Qiming Yuan and Wojciech Zaremba and Rowan Zellers and Chong Zhang and Marvin Zhang and Shengjia Zhao and Tianhao Zheng and Juntang Zhuang and William Zhuk and Barret Zoph},
      year={2024},
      eprint={2303.08774},
      archivePrefix={arXiv},
      primaryClass={cs.CL},
      url={https://arxiv.org/abs/2303.08774}, 
}

@article{roshanaei2025eeg,
  title={EEG-based functional and effective connectivity patterns during emotional episodes using graph theoretical analysis},
  author={Roshanaei, Majid and Norouzi, Hamzeh and Onton, Julie and Makeig, Scott and Mohammadi, Alireza},
  journal={Scientific reports},
  volume={15},
  number={1},
  pages={2174},
  year={2025},
  publisher={Nature Publishing Group UK London}
}

@article{emotion_brain_basis,
  title={The brain basis of emotion: A meta-analytic review},
  author={Kober, Hedy and Wager, Tor D.},
  journal={Neuropsychologia},
  volume={49},
  number={5},
  pages={901-909},
  year={2011},
  publisher={Elsevier},
  url={https://www.affective-science.org/pubs/2012/lindquist-et-al-bbs-2012.pdf}
}

@article{russell_circumplex_2003,
  title={Core Affect and the Psychological Construction of Emotion},
  author={Russell, James A.},
  journal={Psychological Review},
  volume={110},
  number={1},
  pages={145--172},
  year={2003},
  doi={10.1037/0033-295X.110.1.145}
}

@article{phan_insula_disgust,
  title={Functional Neuroanatomy of Emotion: A Meta-Analysis of Emotion Activation Studies in PET and fMRI},
  author={Phan, K. Luan and Wager, Tor D. and others},
  journal={NeuroImage},
  volume={16},
  number={2},
  pages={331--348},
  year={2002},
  doi={10.1006/nimg.2002.1087}
}

@article{ledoux_amygdala,
title = {The amygdala: contributions to fear and stress},
journal = {Seminars in Neuroscience},
volume = {6},
number = {4},
pages = {231-237},
year = {1994},
issn = {1044-5765},
doi = {https://doi.org/10.1006/smns.1994.1030},
url = {https://www.sciencedirect.com/science/article/pii/S104457658471030X},
author = {Joseph E. LeDoux},
}

@article{Saarimaki2016,
  title={Discrete Neural Signatures of Basic Emotions},
  author={Saarimäki, Heini and Gotsopoulos, Athanasios and Jääskeläinen, Iiro P and Lampinen, Jussi and Vuilleumier, Patrik and Hari, Riitta and Sams, Mikko and Nummenmaa, Lauri},
  journal={Cerebral Cortex},
  volume={26},
  number={6},
  pages={2563--2573},
  year={2016},
  publisher={Oxford University Press}
}

@article{Kassam2013,
  title={Identifying Emotions on the Basis of Neural Activation},
  author={Kassam, Karim S and Markey, Amanda R and Cherkassky, Vladimir L and Loewenstein, George and Just, Marcel A},
  journal={PLOS ONE},
  volume={8},
  number={6},
  pages={e66032},
  year={2013},
  publisher={Public Library of Science}
}

@article{Wager2015,
  title={A Bayesian Model of Category-Specific Emotional Brain Responses},
  author={Wager, Tor D and Kang, Jeong-Woo and Johnson, Timothy D and Nichols, Thomas E and Satpute, Ajay B and Barrett, Lisa Feldman},
  journal={PLoS Computational Biology},
  volume={11},
  number={4},
  pages={e1004066},
  year={2015},
  publisher={Public Library of Science}
}

@article{huang2023graph,
  title={Graph-Enhanced Emotion Neural Decoding},
  author={Huang, Zhongyu and Du, Changde and Wang, Yingheng and Fu, Kaicheng and He, Huiguang},
  journal={IEEE Transactions on Medical Imaging},
  volume={42},
  number={8},
  pages={2262--2273},
  year={2023},
  month={August},
  doi={10.1109/TMI.2023.3246220},
  publisher={IEEE},
  issn={1558-254X},
  pmid={37027550}
}

@article{lettieri2019emotionotopy,
  title={Emotionotopy in the human right temporo-parietal cortex},
  author={Lettieri, Gabriella and Handjaras, Giacomo and Ricciardi, Emiliano and Leo, Andrea and Papale, Paolo and Betta, Marco and Pietrini, Pietro},
  journal={Nature Communications},
  volume={10},
  number={1},
  pages={1--13},
  year={2019},
  publisher={Nature Publishing Group},
  doi={10.1038/s41467-019-13599-z},
  url={https://doi.org/10.1038/s41467-019-13599-z}
}

@article{huang_dataset2,
  title={Distinct dimensions of emotion in the human brain and their representation on the cortical surface},
  author={Koide-Majima, Naoko and Nakai, Tomoya and Nishimoto, Shinji},
  journal={NeuroImage},
  volume={222},
  pages={117258},
  year={2020},
  publisher={Elsevier}
}

@article{gpt_annotation1,
  title={ChatGPT outperforms crowd workers for text-annotation tasks},
  author={Gilardi, Fabrizio and Alizadeh, Meysam and Kubli, Ma{\"e}l},
  journal={Proceedings of the National Academy of Sciences},
  volume={120},
  number={30},
  pages={e2305016120},
  year={2023},
  publisher={National Academy of Sciences}
}

@ARTICLE{gpt_annotation2,
  author={Latif, Siddique and Usama, Muhammad and Malik, Muhammad Ibrahim and Schuller, Björn W.},
  journal={IEEE Computational Intelligence Magazine}, 
  title={Can Large Language Models Aid in Annotating Speech Emotional Data? Uncovering New Frontiers [Research Frontier]}, 
  year={2025},
  volume={20},
  number={1},
  pages={66-77},
  doi={10.1109/MCI.2024.3504833}
}

@inproceedings{reynolds2021prompt,
  title={Prompt programming for large language models: Beyond the few-shot paradigm},
  author={Reynolds, Laria and McDonell, Kyle},
  booktitle={Extended abstracts of the 2021 CHI conference on human factors in computing systems},
  pages={1--7},
  year={2021}
}

@book{SPM8,
  title={Statistical parametric mapping: the analysis of functional brain images},
  author={Penny, William D and Friston, Karl J and Ashburner, John T and Kiebel, Stefan J and Nichols, Thomas E},
  year={2011},
  publisher={Elsevier}
}

@article{22seconds,
  title={Tools of the trade: estimating time-varying connectivity patterns from fMRI data},
  author={Iraji, Armin and Faghiri, Ashkan and Lewis, Noah and Fu, Zening and Rachakonda, Srinivas and Calhoun, Vince D},
  journal={Social cognitive and affective neuroscience},
  volume={16},
  number={8},
  pages={849--874},
  year={2021},
  publisher={Oxford University Press UK}
}

@article{30_60seconds1,
  title={Tools of the trade: estimating time-varying connectivity patterns from fMRI data},
  author={Iraji, Armin and Faghiri, Ashkan and Lewis, Noah and Fu, Zening and Rachakonda, Srinivas and Calhoun, Vince D},
  journal={Social cognitive and affective neuroscience},
  volume={16},
  number={8},
  pages={849--874},
  year={2021},
  publisher={Oxford University Press UK}
}

@article{30_60seconds2,
  title={On spurious and real fluctuations of dynamic functional connectivity during rest},
  author={Leonardi, Nora and Van De Ville, Dimitri},
  journal={Neuroimage},
  volume={104},
  pages={430--436},
  year={2015},
  publisher={Elsevier}
}

@article{olson2007temporalpole,
  title={The enigmatic temporal pole: a review of findings on social and emotional processing},
  author={Olson, Ingrid R. and Plotzker, Abby and Ezzyat, Youssef},
  journal={Brain},
  volume={130},
  number={7},
  pages={1718--1731},
  year={2007},
  publisher={Oxford University Press}
}

@article{klasen2011emotion,
  title={Neural contributions to emotion perception in audiovisual media},
  author={Klasen, Michael and Chen, Ying and Mathiak, Klaus},
  journal={NeuroImage},
  volume={58},
  number={1},
  pages={63--70},
  year={2011},
  publisher={Elsevier}
}

@article{Grabenhorst2025,
  title={Neural signatures of temporal anticipation in human cortex represent event probability density},
  author={Matthias Grabenhorst, Georgios Michalareas David, Poeppel},
  journal={Nature Communications},
  volume={16},
  number={1},
  pages={2602},
  year={2025},
  publisher={Nature Publishing Group},
  doi={10.1038/s41467-025-57813-7}
}

@article{morawetz2016emotion,
  title={Neural representation of emotion regulation goals},
  author={Morawetz, Carmen and Bode, Stefan and Baudewig, Juergen and Jacobs, Arthur M and Heekeren, Hauke R},
  journal={Human Brain Mapping},
  volume={37},
  number={2},
  pages={600--620},
  year={2016},
  publisher={Wiley},
  doi={10.1002/hbm.23050}
}

@article{rolls2019limbic,
  title={The cingulate cortex and limbic systems for emotion, action, and memory},
  author={Rolls, Edmund T},
  journal={Brain Structure and Function},
  volume={224},
  number={9},
  pages={3001--3018},
  year={2019},
  publisher={Springer}
}

@article{kruskal1956shortest,
  title={On the shortest spanning subtree of a graph and the traveling salesman problem},
  author={Kruskal, Joseph B.},
  journal={Proceedings of the American Mathematical Society},
  volume={7},
  number={1},
  pages={48--50},
  year={1956},
  publisher={American Mathematical Society},
  doi={10.2307/2033241}
}

\newpage
\appendix

\section{Large Brain Networks in Schaefers Atlas} \label{schaefers networks}

\begin{table} [h!]
\centering
\scriptsize
\setlength{\defaultaddspace}{3pt}
\begin{tabular}{p{2.0cm}p{4.4cm}p{8.9cm}}
\toprule
\textbf{Abbreviation} & \textbf{Network Name} & \textbf{General Brain Regions Involved} \\ \addlinespace
\hlineB{2}
\rowcolor{lightgray!50} \multicolumn{3}{c}{17-network parcellation}\\ \hlineB{2}\addlinespace
ContA & Control Network A & Dorsolateral prefrontal cortex, anterior cingulate cortex \\ \addlinespace
ContB & Control Network B & Inferior parietal lobule, lateral prefrontal cortex \\ \addlinespace
ContC & Control Network C & Posterior parietal cortex, medial prefrontal cortex \\ \addlinespace
DefaultA & Default Mode Network A & Medial prefrontal cortex, posterior cingulate cortex \\ \addlinespace
DefaultB & Default Mode Network B & Inferior parietal lobule, lateral temporal cortex \\ \addlinespace
DefaultC & Default Mode Network C & Hippocampal formation, retrosplenial cortex \\ \addlinespace
DorsAttnA & Dorsal Attention Network A & Intraparietal sulcus, frontal eye fields \\ \addlinespace
DorsAttnB & Dorsal Attention Network B & Superior parietal lobule, supplementary eye fields \\ \addlinespace
LimbicA & Limbic Network A & Orbitofrontal cortex, temporal pole \\ \addlinespace
LimbicB & Limbic Network B & Parahippocampal gyrus, entorhinal cortex \\ \addlinespace
SalVentAttnA & Salience/Ventral Attention A & Anterior insula, dorsal anterior cingulate cortex \\ \addlinespace
SalVentAttnB & Salience/Ventral Attention B & Temporoparietal junction, ventrolateral prefrontal cortex \\ \addlinespace
SomMotA & Somatomotor Network A & Precentral gyrus, postcentral gyrus \\ \addlinespace
SomMotB & Somatomotor Network B & Supplementary motor area, paracentral lobule \\ \addlinespace
TempPar & Temporoparietal Network & Superior temporal gyrus, inferior parietal lobule \\ \addlinespace
VisCent & Visual Central Network & Primary visual cortex (V1), calcarine sulcus \\ \addlinespace
VisPeri & Visual Peripheral Network & Extrastriate visual areas, lateral occipital cortex \\ \addlinespace
\hlineB{2}
\rowcolor{lightgray!50} \multicolumn{3}{c}{7-network parcellation}\\ \hlineB{2}\addlinespace
Vis & Visual Network & Primary and secondary visual cortices, including the occipital lobe and parts of the fusiform gyrus \\ \addlinespace
SomMot & Somatomotor Network & Precentral gyrus (primary motor cortex), postcentral gyrus (primary somatosensory cortex), supplementary motor area \\ \addlinespace
DorsAttn & Dorsal Attention Network & Intraparietal sulcus, superior parietal lobule, frontal eye fields \\ \addlinespace
SalVentAttn & Salience/Ventral Attention Network & Anterior insula, dorsal anterior cingulate cortex, temporoparietal junction \\ \addlinespace
Limbic & Limbic Network & Orbitofrontal cortex, temporal pole, parahippocampal gyrus \\ \addlinespace
Cont & Frontoparietal Control Network & Dorsolateral prefrontal cortex, inferior parietal lobule, lateral prefrontal cortex \\ \addlinespace
Default & Default Mode Network & Medial prefrontal cortex, posterior cingulate cortex, inferior parietal lobule, lateral temporal cortex \\ \addlinespace
\bottomrule
\end{tabular}
\caption{Descriptions of the 17 and 7 network parcellation in \cite{Schaefer2018}.}
\label{tab:schaefer_17networks}
\end{table}

\section{GPT-4 Labeling Prompt Pseudo Code} \label{appendix prompt}

\begin{lstlisting}[
    basicstyle=\footnotesize\ttfamily,
    frame=single,
    breaklines=true,
    breakatwhitespace=false,
    columns=fullflexible,
    mathescape=true,
    caption={Pseudo-code for GPT-4 prompt to generate fine-grained emotional pseudo-labels.},
    captionpos=b,
    xleftmargin=.1\textwidth, 
    xrightmargin=.1\textwidth
]
Function analyze_emotions(segment):
    prompt = "Analyze the following text for sentiment. 
    Provide a score between 0 and 1" 
             + " for each of the categories, where 0 is the
             lowest and 1 is the highest: "
             + Join(', ', categories) 
             + ".\n\nText: \"" + segment + "\". Focus 
             primarily on the last sentence."

    response = OpenAI.Chat.Completions.Create(
        model = "gpt-4",
        messages = [
            {role: "system", content: "You are a sentiment
            analysis model."},
            {role: "user", content: prompt}
        ],
        max_tokens = 150,
        temperature = 0
    )
    scores = Trim(response.choices[0].message.content)

    Return scores
\end{lstlisting}

\section{Human Evaluation of Emotion Labeling} \label{emotion survey}
\begin{figure}[H]
    \begin{center}
    \includegraphics[width=461pt, height=245pt]{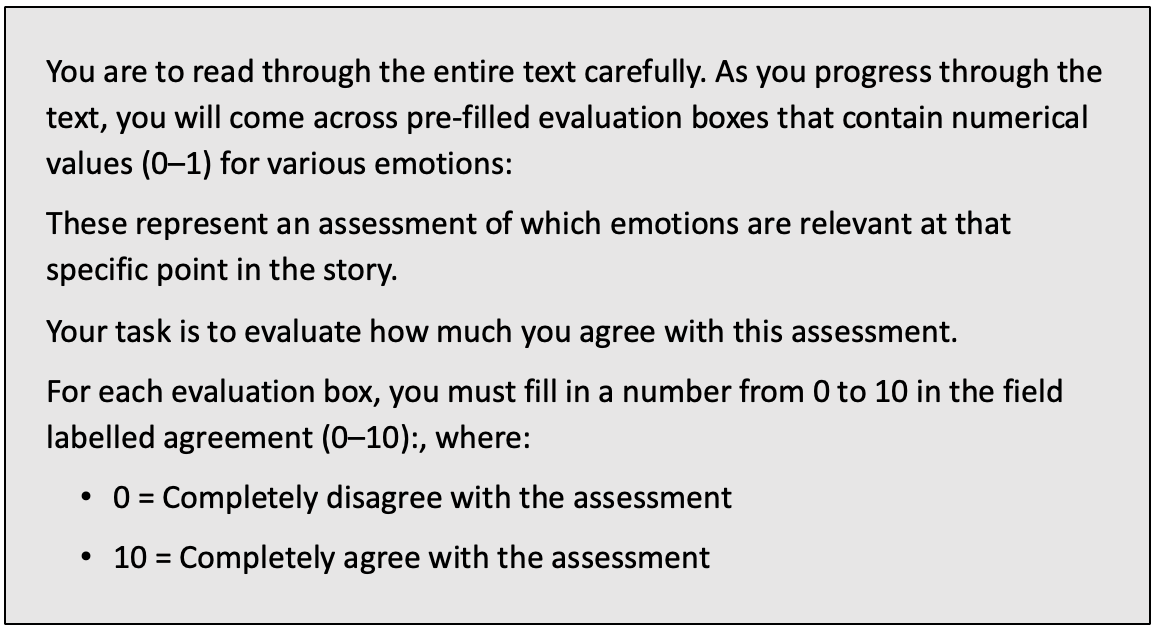}
    \caption{Instructions for the human evaluation of emotion labelling.}
    \label{fig:human_evaluation_instruction}
    \end{center}
\end{figure}

\section{Emotion Analysis over Story Segments} \label{appendix emotional analysis}

\begin{figure}[H]
    \centering
    \begin{subfigure}[b]{0.49\textwidth}
        \includegraphics[width=\textwidth]{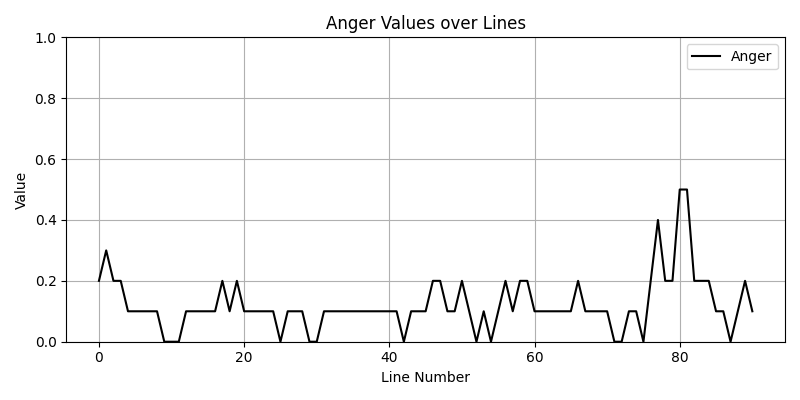}
        \label{fig:anger_plot}
    \end{subfigure}
    \hfill
    \begin{subfigure}[b]{0.49\textwidth}
        \includegraphics[width=\textwidth]{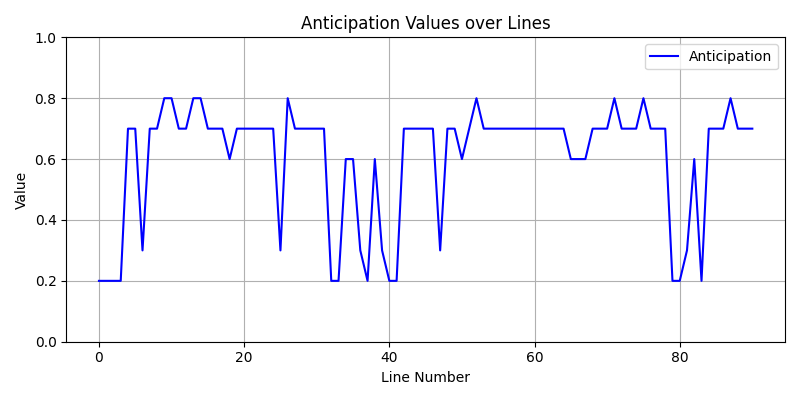}
        \label{fig:anticipation_plot}
    \end{subfigure}
    \begin{subfigure}[b]{0.49\textwidth}
        \includegraphics[width=\textwidth]{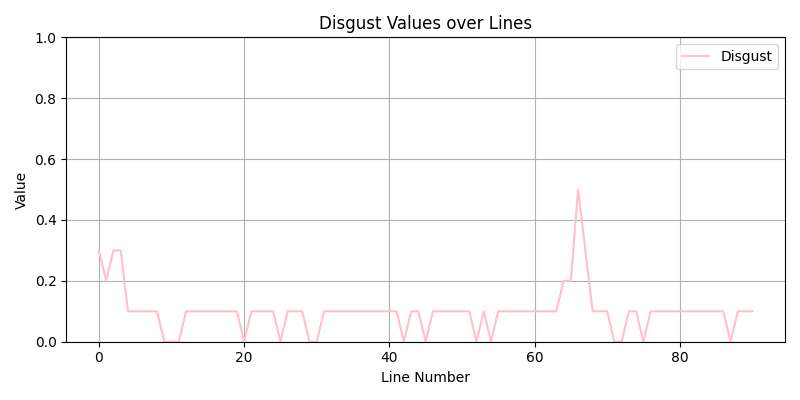}
        \label{fig:disgust_plot}
    \end{subfigure}
    \hfill
    \begin{subfigure}[b]{0.49\textwidth}
        \includegraphics[width=\textwidth]{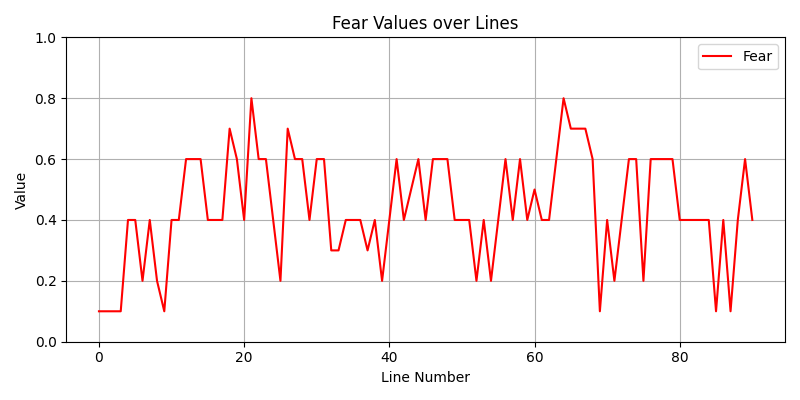}
        \label{fig:fear_plot}
    \end{subfigure}
    \begin{subfigure}[b]{0.49\textwidth}
        \includegraphics[width=\textwidth]{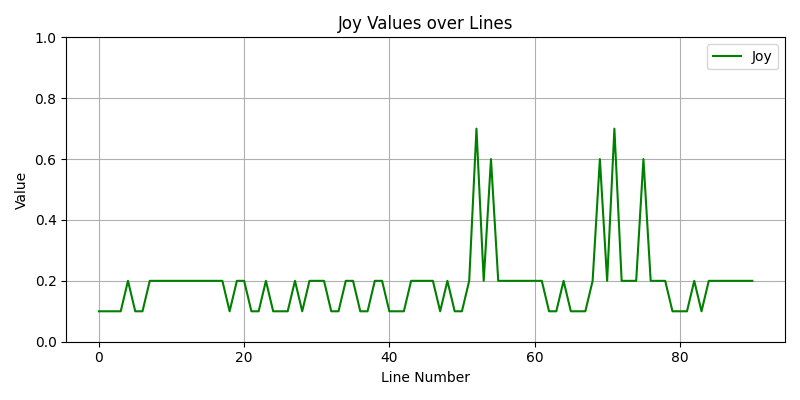}
        \label{fig:joy_plot}
    \end{subfigure}
    \hfill
    \begin{subfigure}[b]{0.49\textwidth}
        \includegraphics[width=\textwidth]{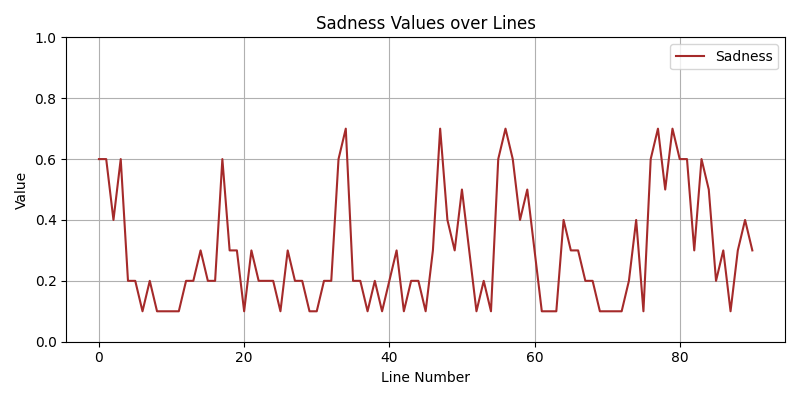}
        \label{fig:sadness_plot}
    \end{subfigure}
    \begin{subfigure}[b]{0.49\textwidth}
        \includegraphics[width=\textwidth]{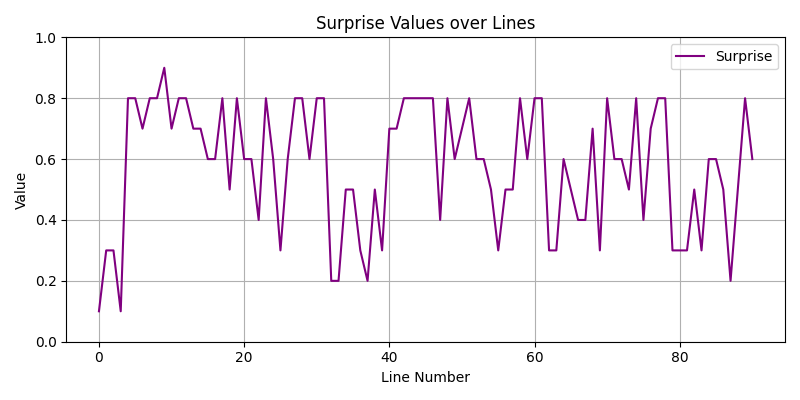}
        \label{fig:surprise_plot}
    \end{subfigure}
    \hfill
    \begin{subfigure}[b]{0.49\textwidth}
        \includegraphics[width=\textwidth]{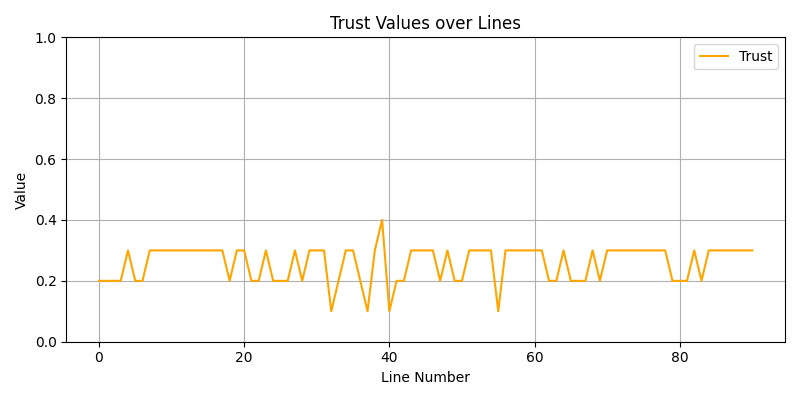}
        \label{fig:trust_plot}
    \end{subfigure}
    \label{fig:emotion_plots}
    \caption{Emotional scores for each segment in the stimuli of Alice in Wonderland chapter 1.}
\end{figure}

\section{Model Performance Comparison with 80-20 Split} \label{appendix_model_performance}

\begin{table}[H]
\vspace{-2cm}
\centering
\caption{Model Performance Comparison ($R^2$ and MSE) across ROI and DFC Datasets, using a 80-20 Train-Test split.}
\label{tab:performance_80_20}
\footnotesize
\begin{tabular}{@{}>{\raggedright\arraybackslash}m{2.4cm}@{}>{\raggedright\arraybackslash}m{0.8cm}>{\centering\arraybackslash}m{1.1cm}>{\centering\arraybackslash}m{1.1cm}>{\centering\arraybackslash}m{1.1cm}>{\centering\arraybackslash}m{1.1cm}>{\centering\arraybackslash}m{1.1cm}>{\centering\arraybackslash}m{1.2cm}>{\centering\arraybackslash}m{1.2cm}@{}}
\toprule
\textbf{Metric} & \textbf{Data} & \textbf{Linear Reg.} & \textbf{Lasso Reg.} & \textbf{Ridge Reg.} & \textbf{SVR (RBF)} & \textbf{Linear SVR} & \textbf{Random Forest} & \textbf{MM-RFR} \\
\midrule
\textbf{Test $R^2$}  & ROI & 0.0489 & 0.0265 & 0.0707 & $\bm{0.2289}$ & -0.0610 & 0.1727 & 0.1749 \\
                     & DFC & 0.3092 & 0.4101 & 0.4747 & $\bm{0.5037}$ & 0.3471 & 0.1203 & 0.1562 \\
\addlinespace
\textbf{Test MSE}    & ROI & 0.0209 & 0.0214 & 0.0206 & $\bm{0.0169}$ & 0.0236 & 0.0181 & 0.0181 \\
                     & DFC & 0.0074 & 0.0066 & 0.0060 & $\bm{0.0056}$ & 0.0071 & 0.0105 & 0.0100 \\
\bottomrule
\end{tabular}
\vspace{-1cm}
\end{table}

\section{ROI Explanations} \label{appendix_roi_explanations}
\newcommand{\roiwidth}{0.24\textwidth}

\begin{figure}[H]
\vspace{-1cm}
\centering
\begin{subfigure}[t]{\roiwidth}
\includegraphics[width=\linewidth]{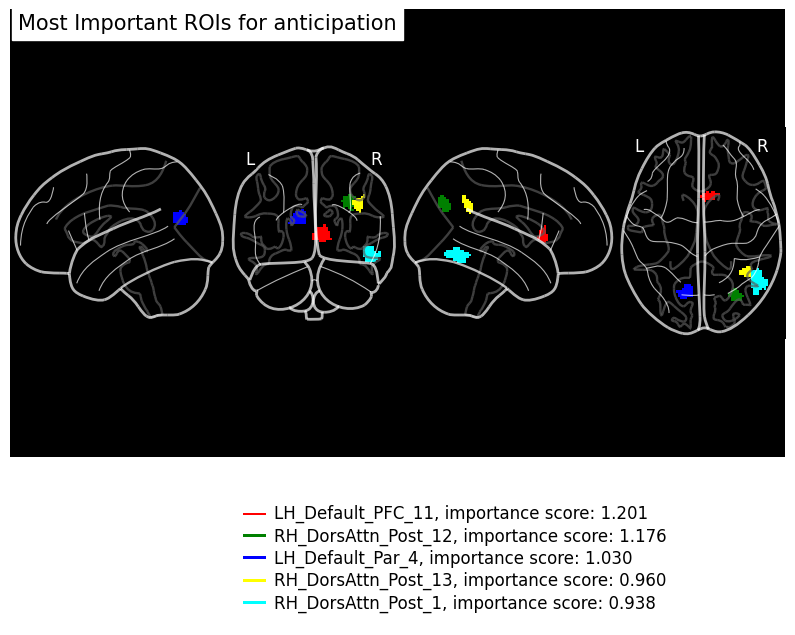}
\caption{Anticipation}
\end{subfigure}
\hfill
\begin{subfigure}[t]{\roiwidth}
\includegraphics[width=\linewidth]{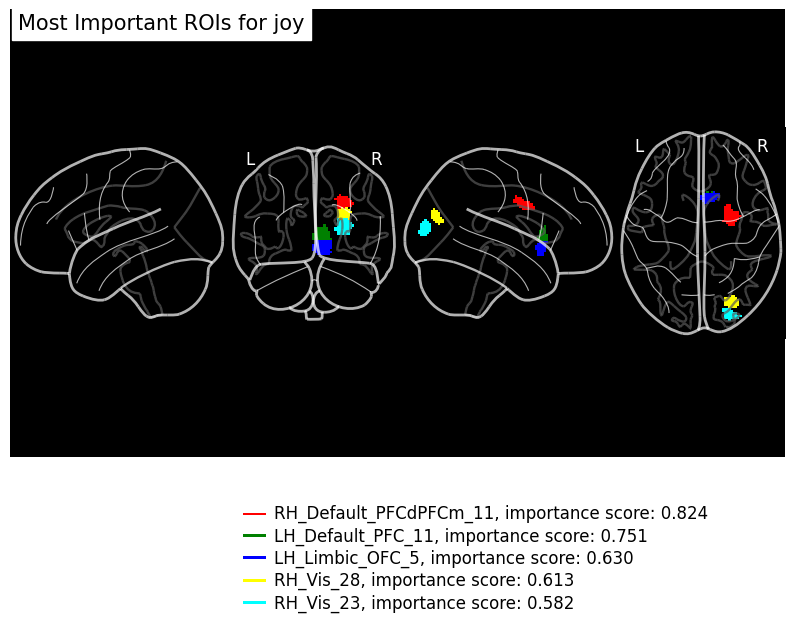}
\caption{Joy}
\end{subfigure}
\hfill
\begin{subfigure}[t]{\roiwidth}
\includegraphics[width=\linewidth]{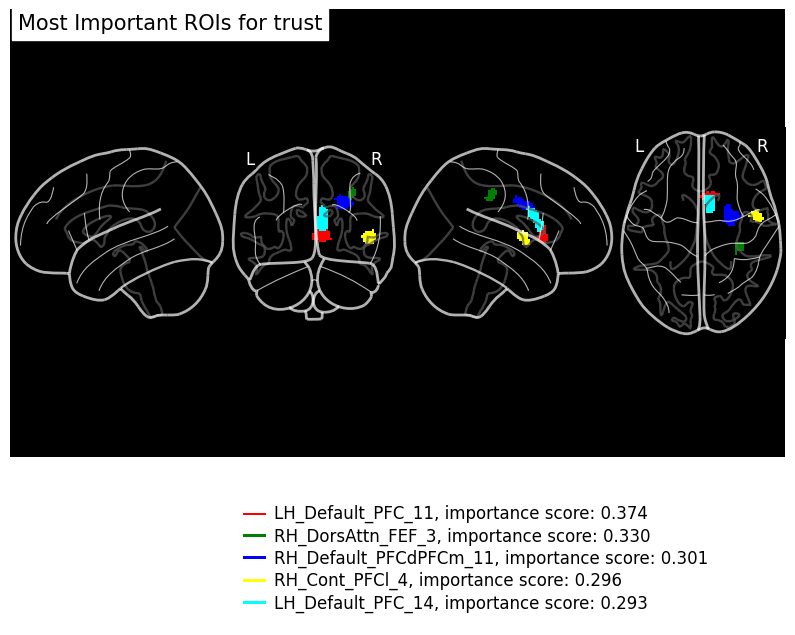}
\caption{Trust}
\end{subfigure}
\hfill
\begin{subfigure}[t]{\roiwidth}
\includegraphics[width=\linewidth]{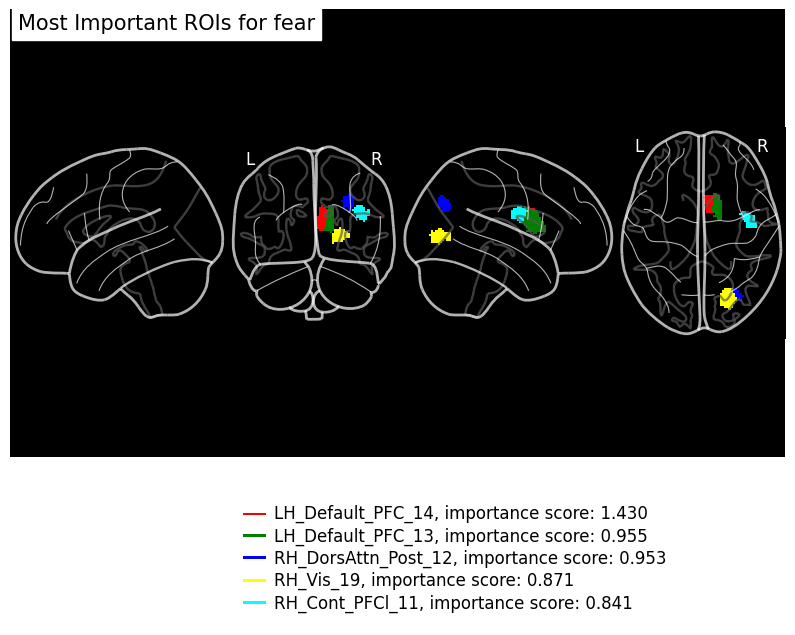}
\caption{Fear}
\end{subfigure}

\vspace{0.3cm}

\begin{subfigure}[t]{\roiwidth}
\includegraphics[width=\linewidth]{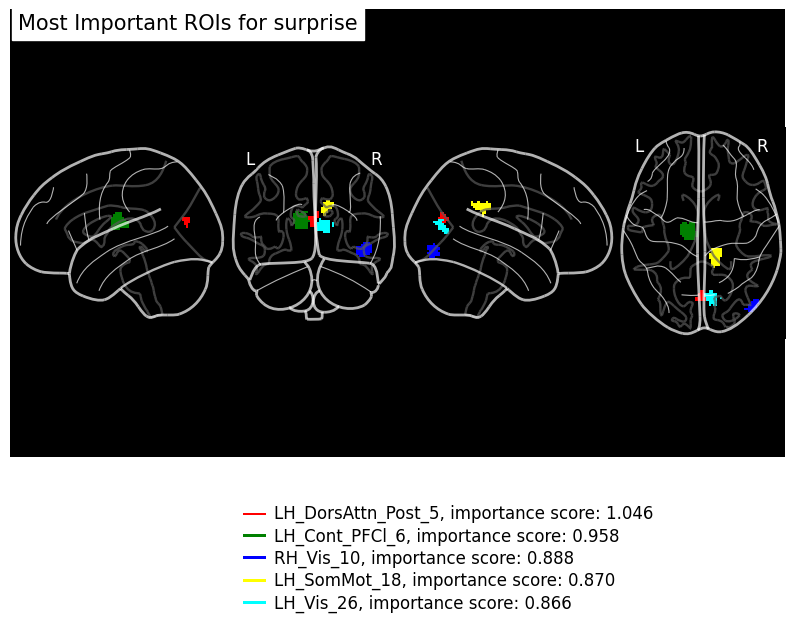}
\caption{Surprise}
\end{subfigure}
\hfill
\begin{subfigure}[t]{\roiwidth}
\includegraphics[width=\linewidth]{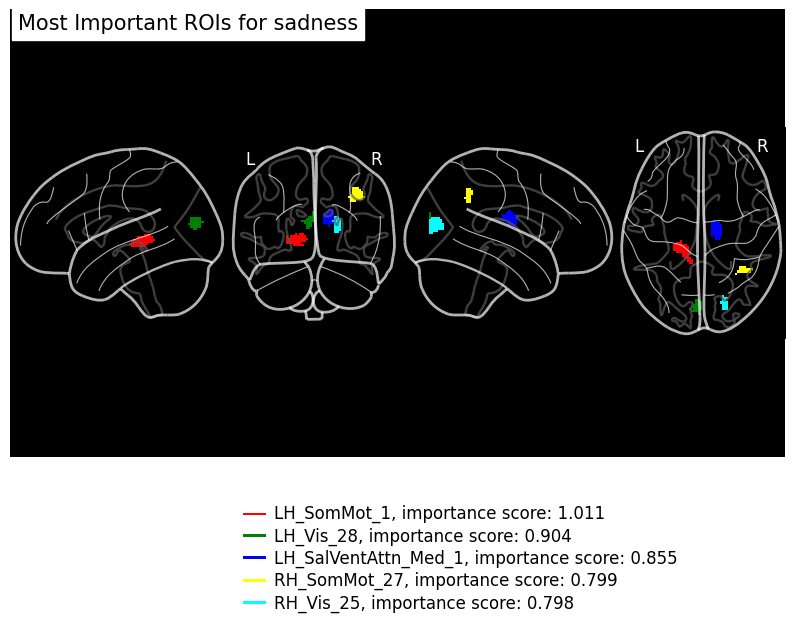}
\caption{Sadness}
\end{subfigure}
\hfill
\begin{subfigure}[t]{\roiwidth}
\includegraphics[width=\linewidth]{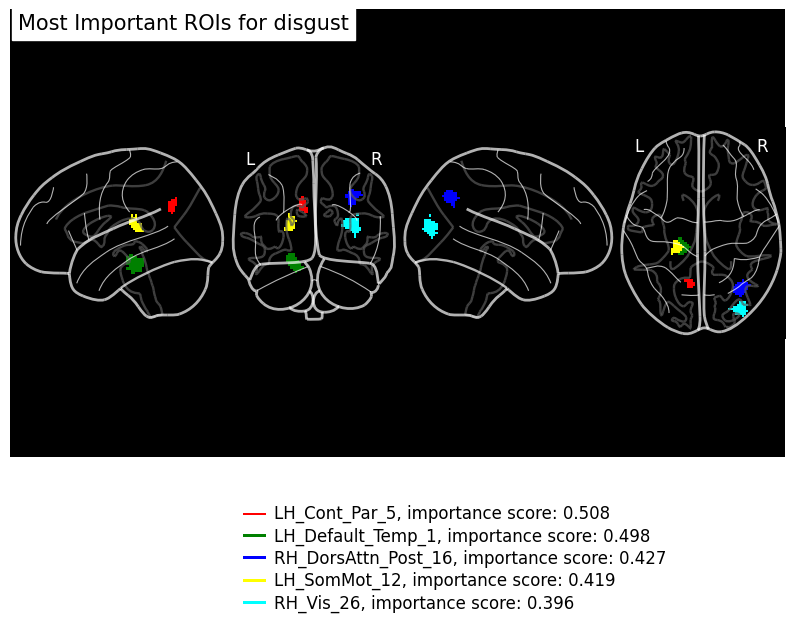}
\caption{Disgust}
\end{subfigure}
\hfill
\begin{subfigure}[t]{\roiwidth}
\includegraphics[width=\linewidth]{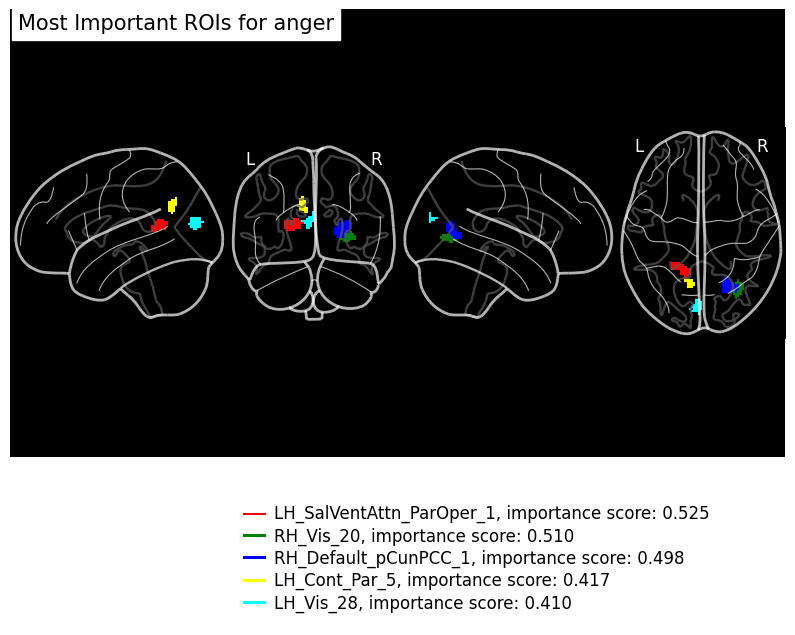}
\caption{Anger}
\end{subfigure}

\caption{Linear Regression ROI Importance Maps for the eight emotion categories.}
\label{fig:LM_ROI_All}
\vspace{2cm}
\end{figure}

\begin{figure}[H]
\centering
\begin{subfigure}[t]{\roiwidth}
\includegraphics[width=\linewidth]{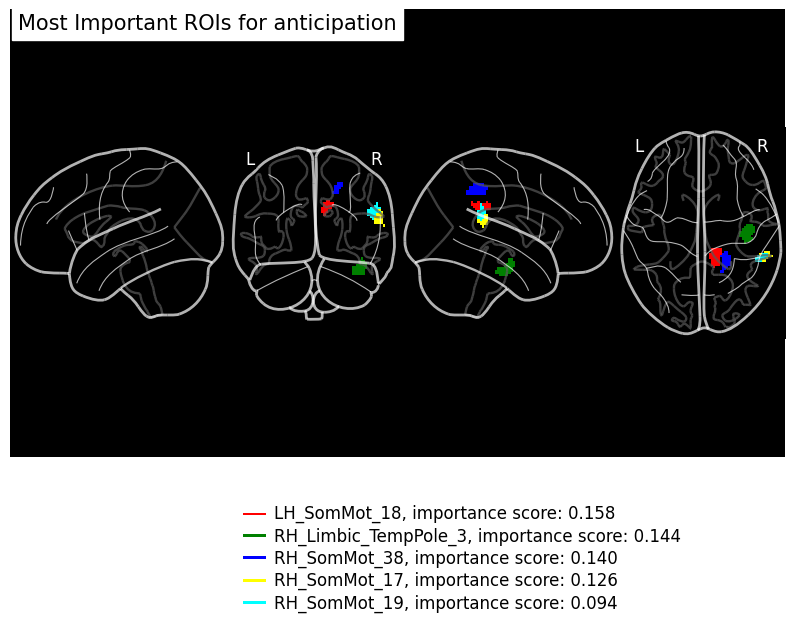}
\caption{Anticipation}
\end{subfigure}
\hfill
\begin{subfigure}[t]{\roiwidth}
\includegraphics[width=\linewidth]{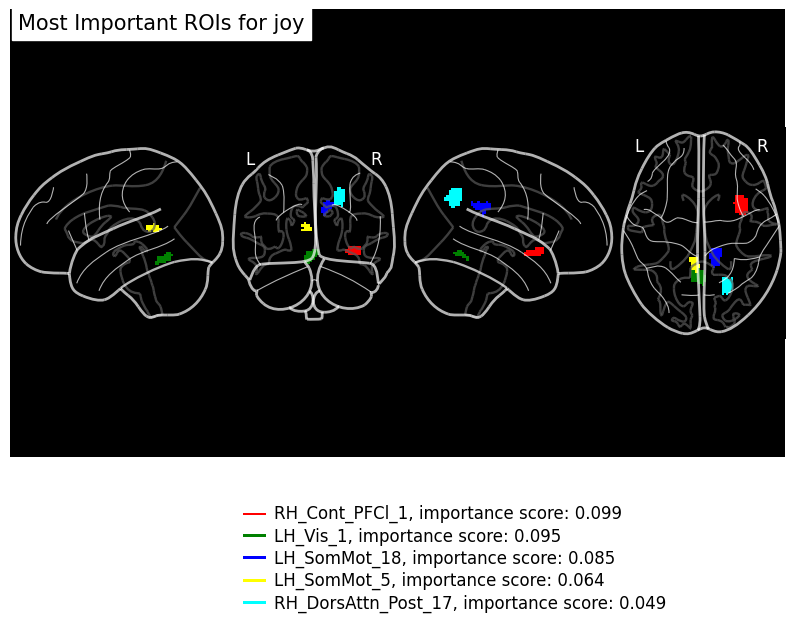}
\caption{Joy}
\end{subfigure}
\hfill
\begin{subfigure}[t]{\roiwidth}
\includegraphics[width=\linewidth]{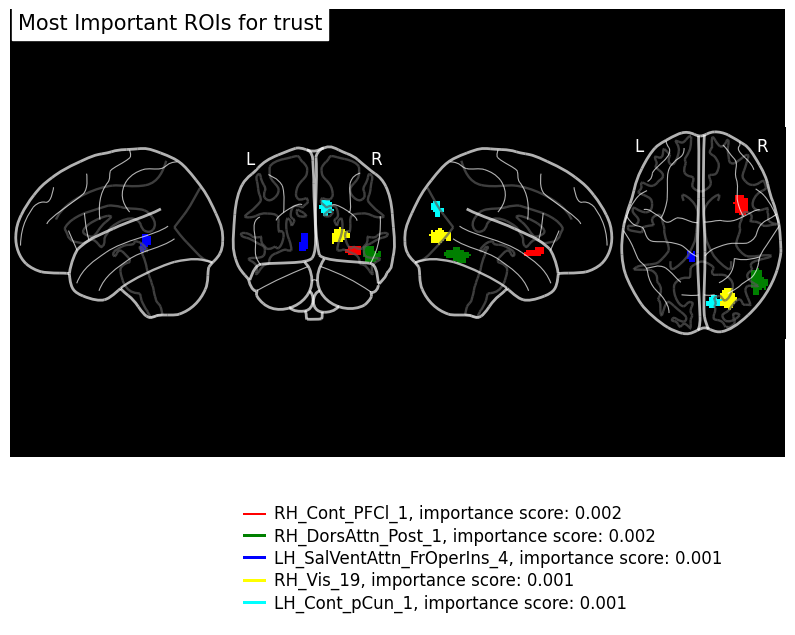}
\caption{Trust}
\end{subfigure}
\hfill
\begin{subfigure}[t]{\roiwidth}
\includegraphics[width=\linewidth]{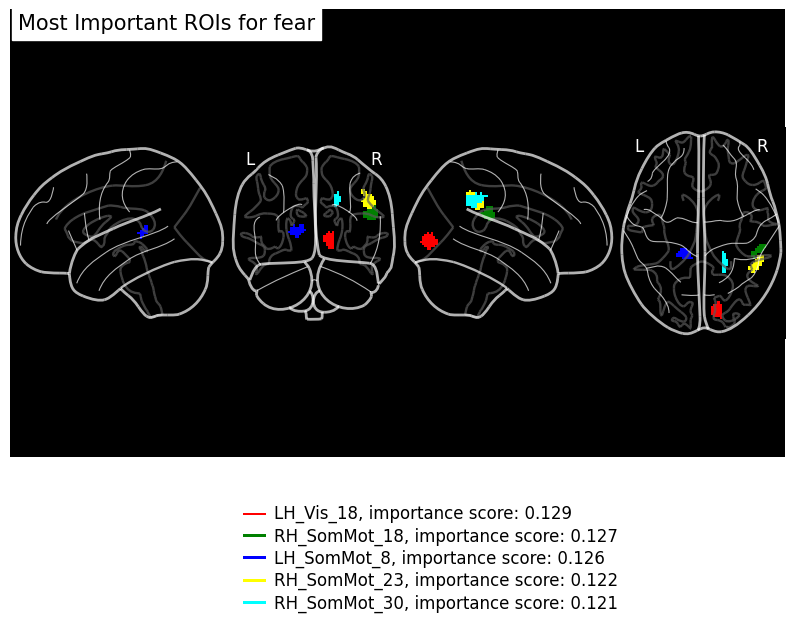}
\caption{Fear}
\end{subfigure}

\vspace{0.3cm}

\begin{subfigure}[t]{\roiwidth}
\includegraphics[width=\linewidth]{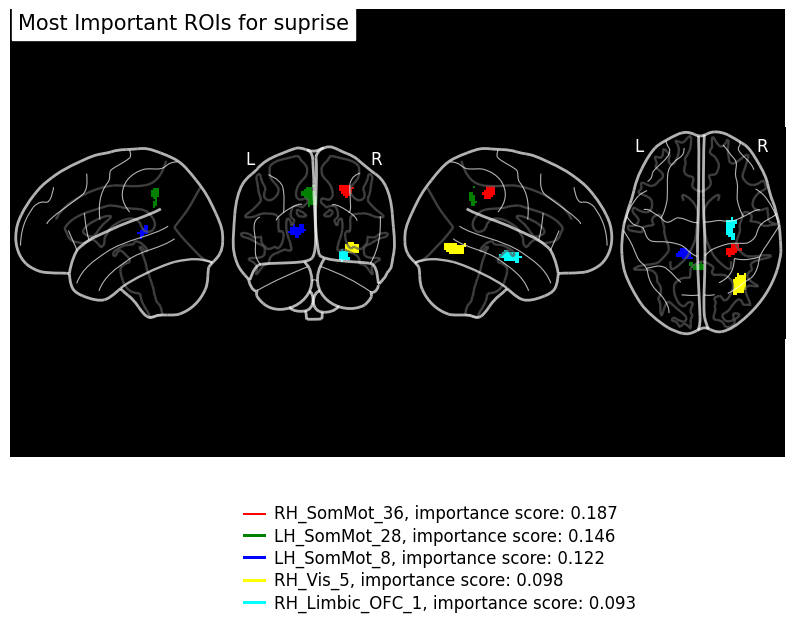}
\caption{Surprise}
\end{subfigure}
\hfill
\begin{subfigure}[t]{\roiwidth}
\includegraphics[width=\linewidth]{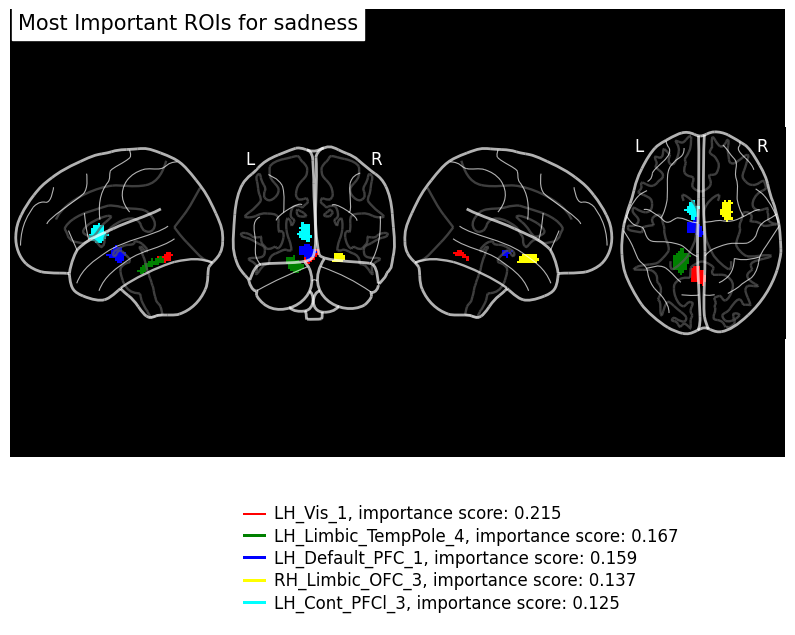}
\caption{Sadness}
\end{subfigure}
\hfill
\begin{subfigure}[t]{\roiwidth}
\includegraphics[width=\linewidth]{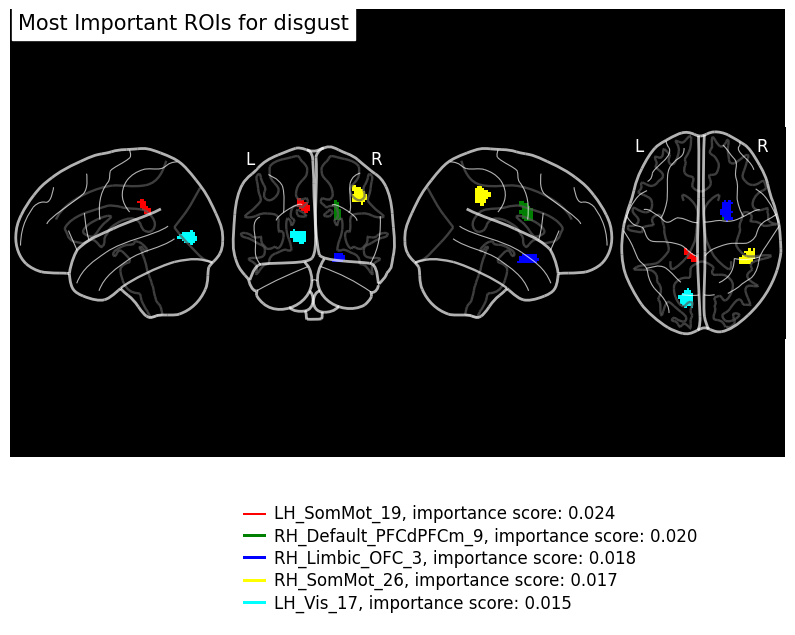}
\caption{Disgust}
\end{subfigure}
\hfill
\begin{subfigure}[t]{\roiwidth}
\includegraphics[width=\linewidth]{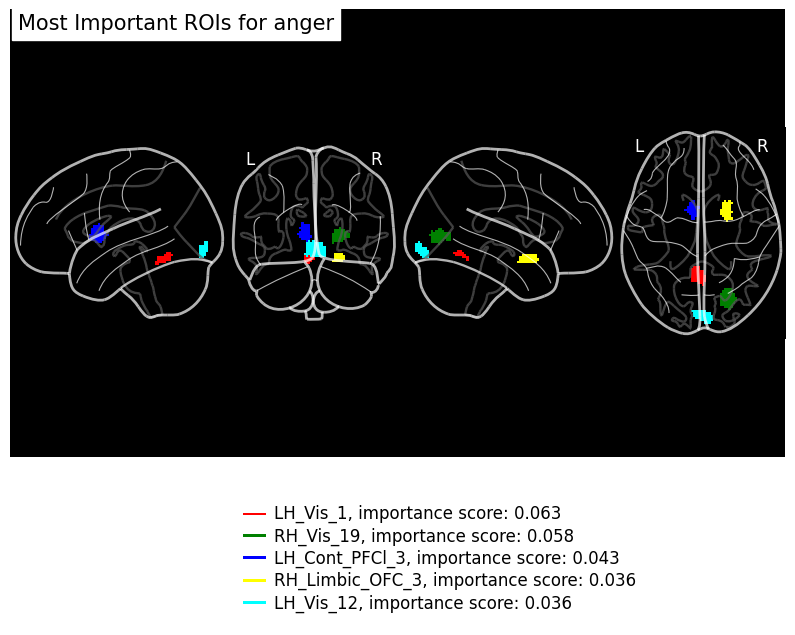}
\caption{Anger}
\end{subfigure}

\caption{Lasso ROI Importance Maps for the eight emotion categories.}
\label{fig:Lasso_ROI_All}
\vspace{-1cm}
\end{figure}


\begin{figure}[H]
\vspace{-1cm}
\centering
\begin{subfigure}[t]{\roiwidth}
\includegraphics[width=\linewidth]{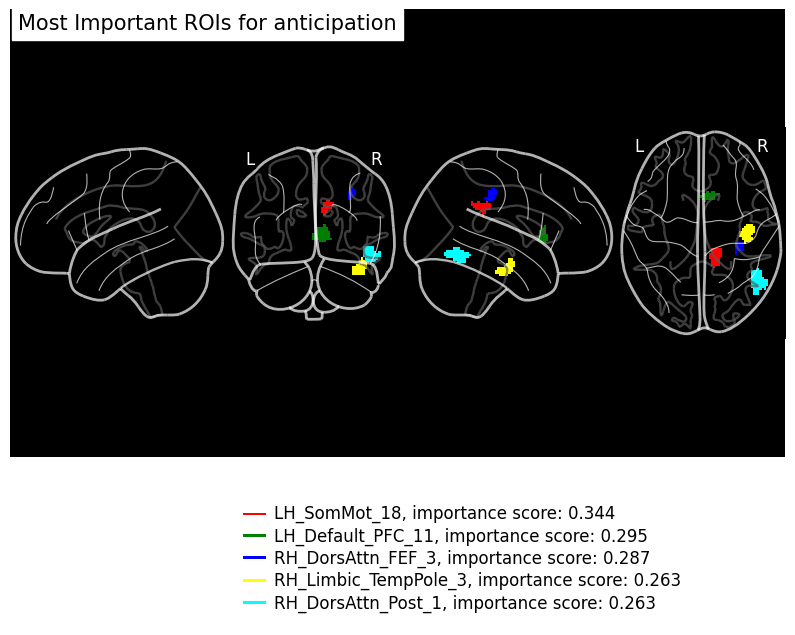}
\caption{Anticipation}
\end{subfigure}
\hfill
\begin{subfigure}[t]{\roiwidth}
\includegraphics[width=\linewidth]{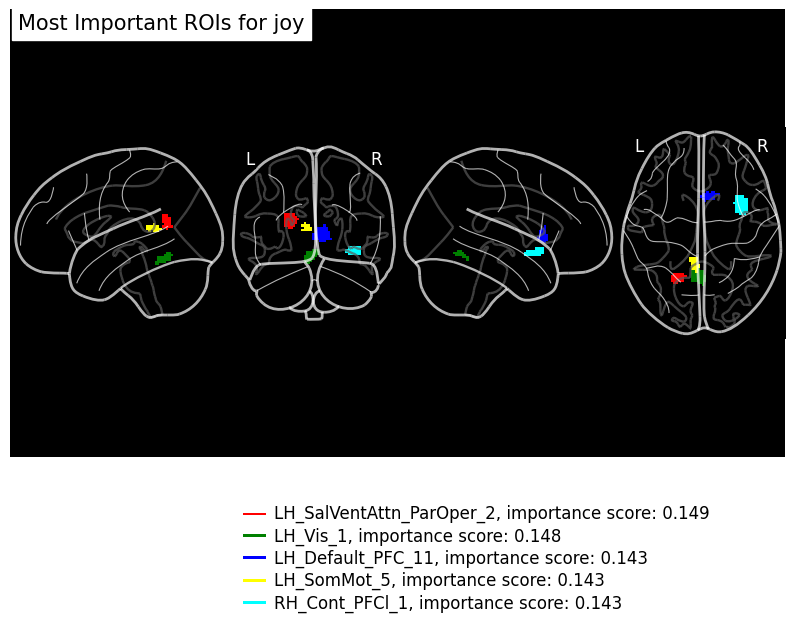}
\caption{Joy}
\end{subfigure}
\hfill
\begin{subfigure}[t]{\roiwidth}
\includegraphics[width=\linewidth]{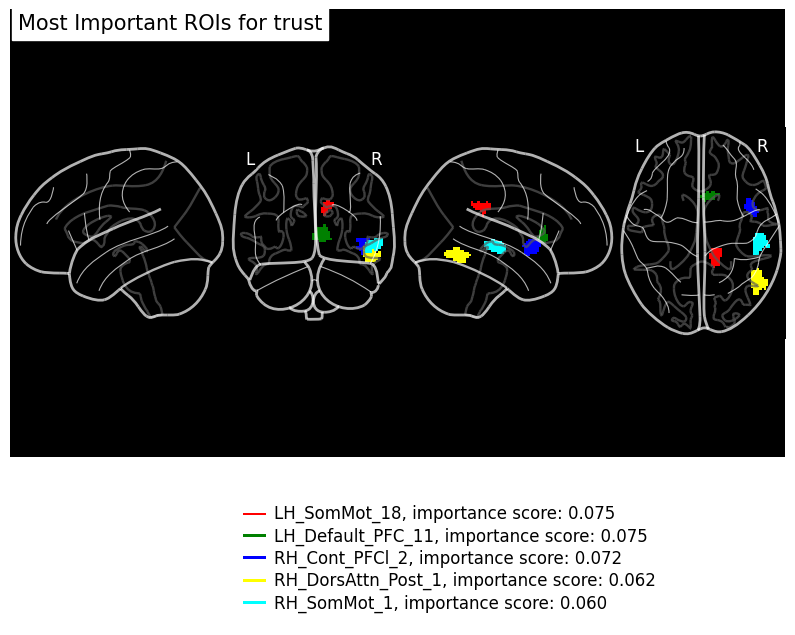}
\caption{Trust}
\end{subfigure}
\hfill
\begin{subfigure}[t]{\roiwidth}
\includegraphics[width=\linewidth]{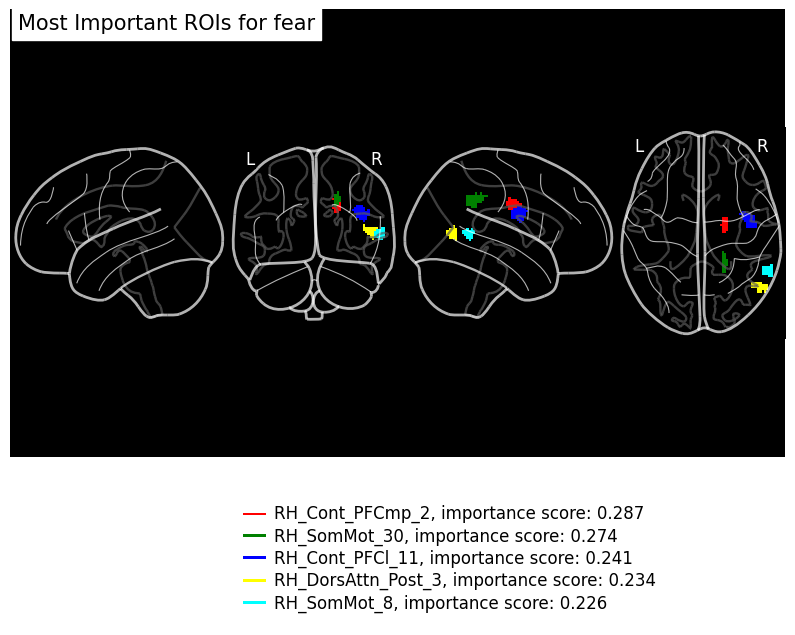}
\caption{Fear}
\end{subfigure}

\vspace{0.3cm}

\begin{subfigure}[t]{\roiwidth}
\includegraphics[width=\linewidth]{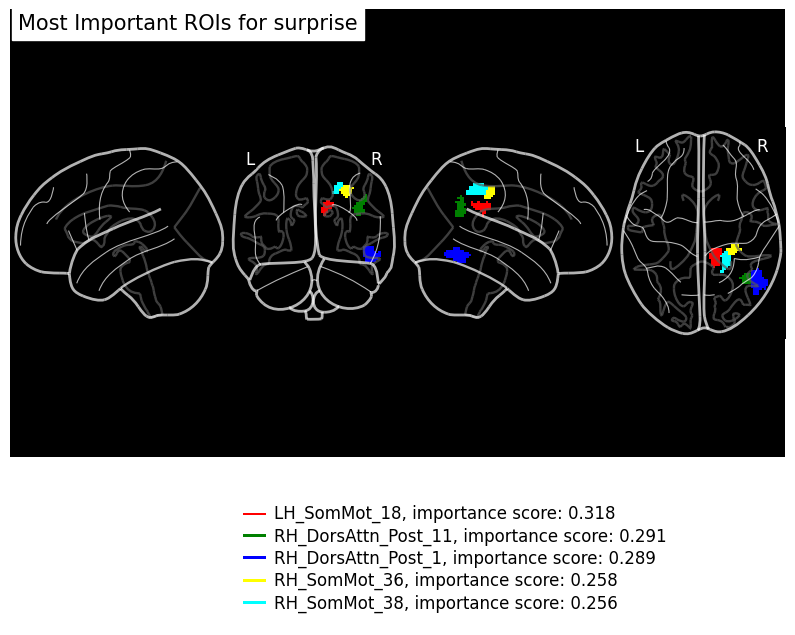}
\caption{Surprise}
\end{subfigure}
\hfill
\begin{subfigure}[t]{\roiwidth}
\includegraphics[width=\linewidth]{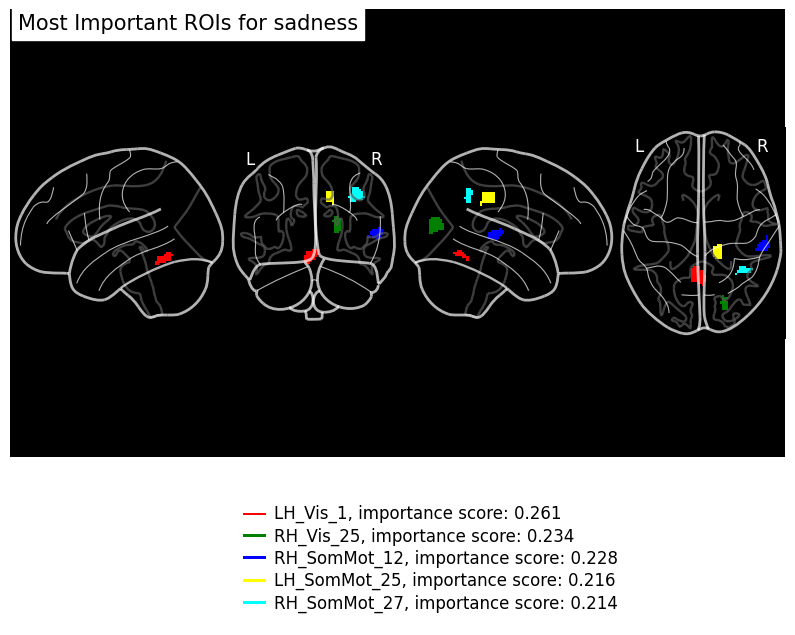}
\caption{Sadness}
\end{subfigure}
\hfill
\begin{subfigure}[t]{\roiwidth}
\includegraphics[width=\linewidth]{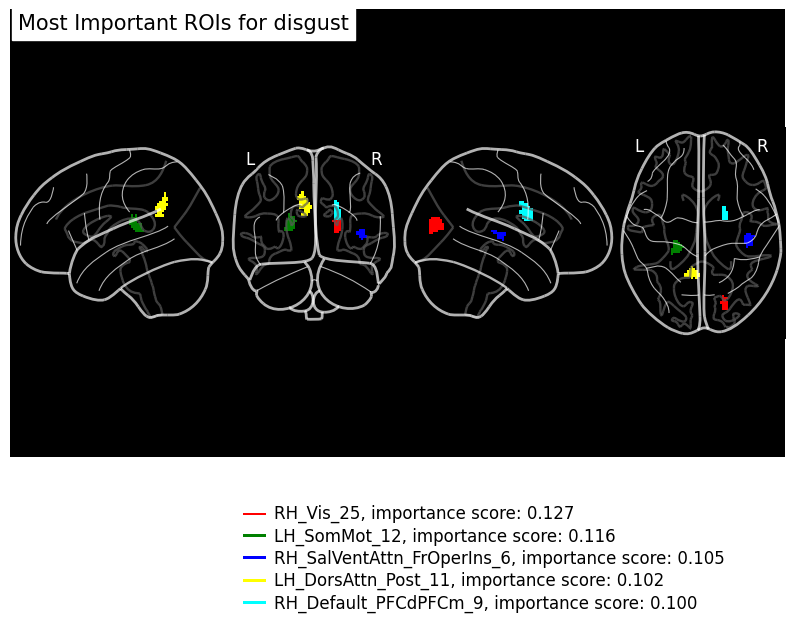}
\caption{Disgust}
\end{subfigure}
\hfill
\begin{subfigure}[t]{\roiwidth}
\includegraphics[width=\linewidth]{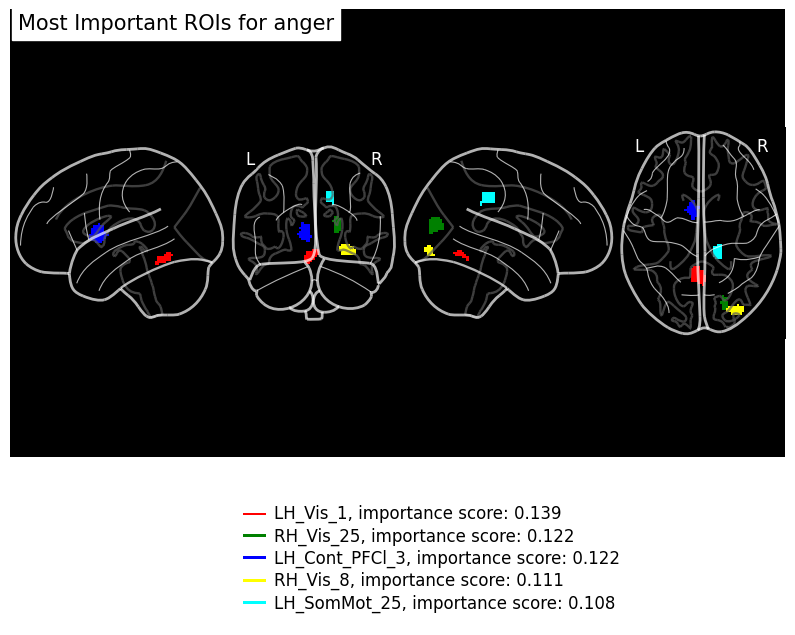}
\caption{Anger}
\end{subfigure}

\caption{Ridge ROI Importance Maps for the eight emotion categories.}
\label{fig:Ridge_ROI_All}
\vspace{2cm}
\end{figure}


\begin{figure}[H]
\centering
\begin{subfigure}[t]{\roiwidth}
\includegraphics[width=\linewidth]{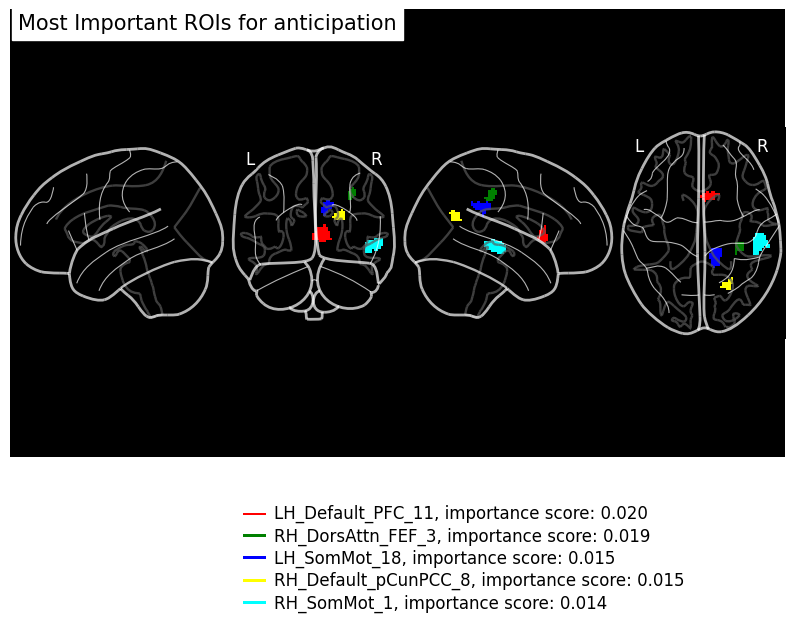}
\caption{Anticipation}
\end{subfigure}
\hfill
\begin{subfigure}[t]{\roiwidth}
\includegraphics[width=\linewidth]{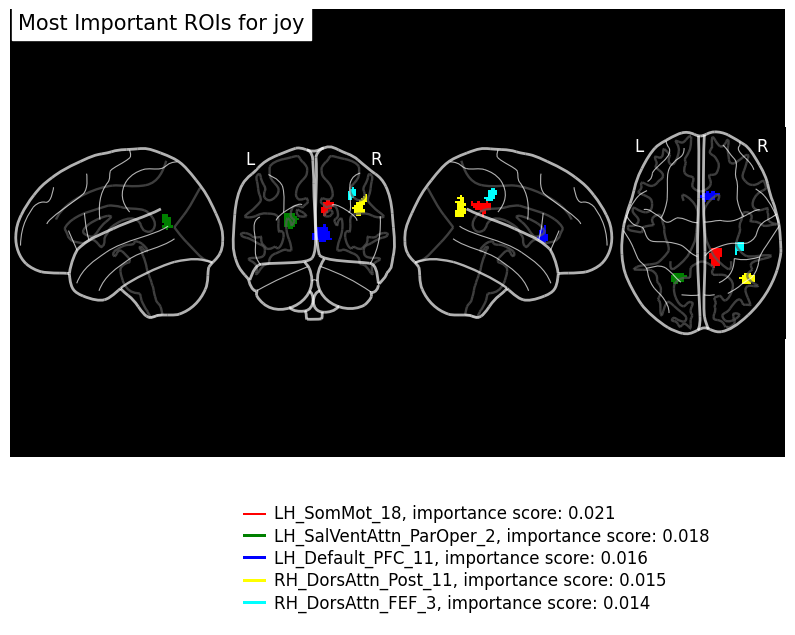}
\caption{Joy}
\end{subfigure}
\hfill
\begin{subfigure}[t]{\roiwidth}
\includegraphics[width=\linewidth]{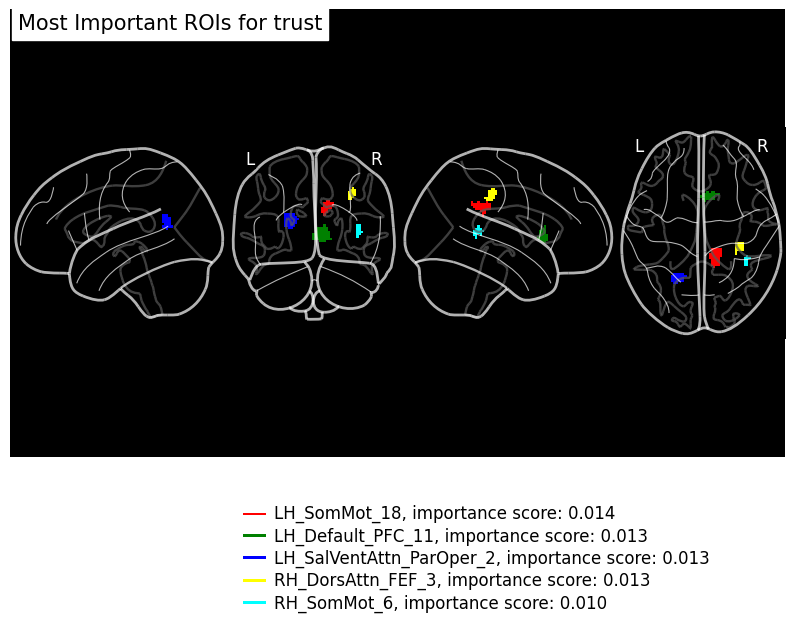}
\caption{Trust}
\end{subfigure}
\hfill
\begin{subfigure}[t]{\roiwidth}
\includegraphics[width=\linewidth]{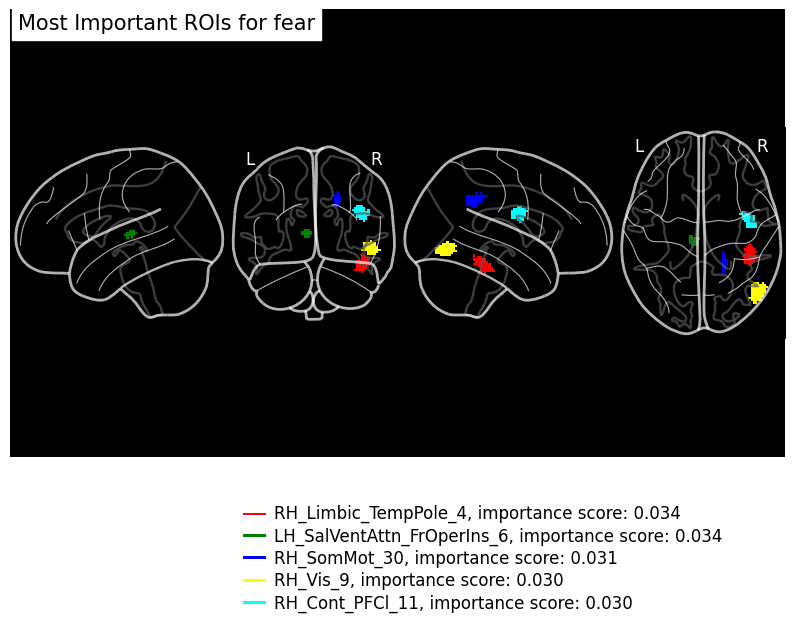}
\caption{Fear}
\end{subfigure}

\vspace{0.3cm}

\begin{subfigure}[t]{\roiwidth}
\includegraphics[width=\linewidth]{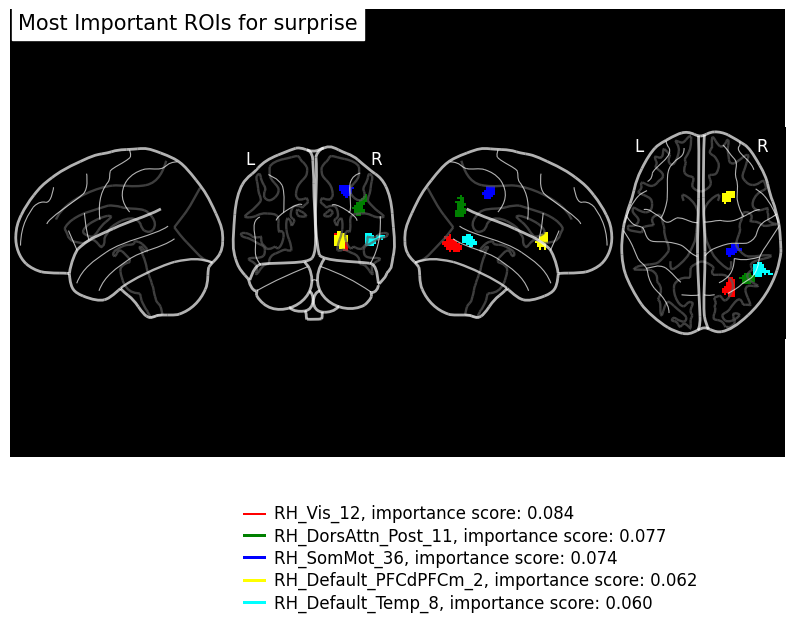}
\caption{Surprise}
\end{subfigure}
\hfill
\begin{subfigure}[t]{\roiwidth}
\includegraphics[width=\linewidth]{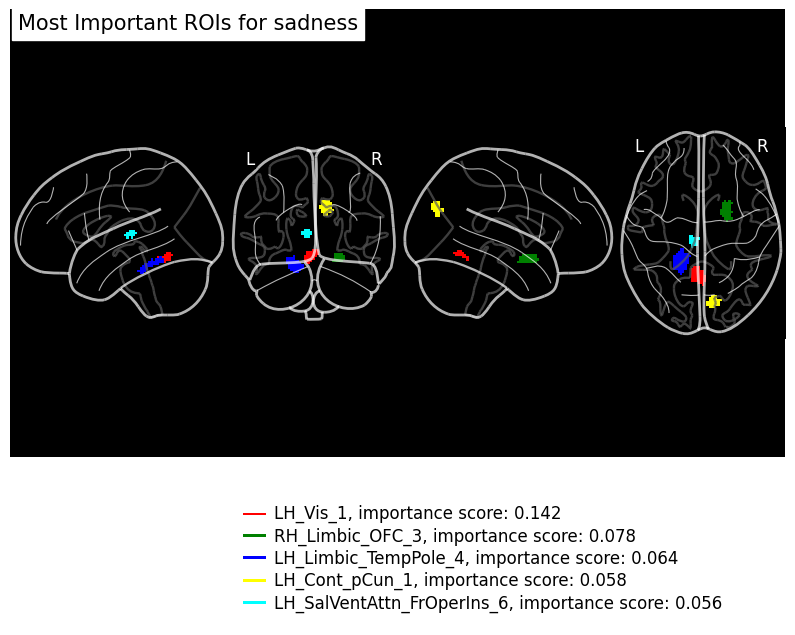}
\caption{Sadness}
\end{subfigure}
\hfill
\begin{subfigure}[t]{\roiwidth}
\includegraphics[width=\linewidth]{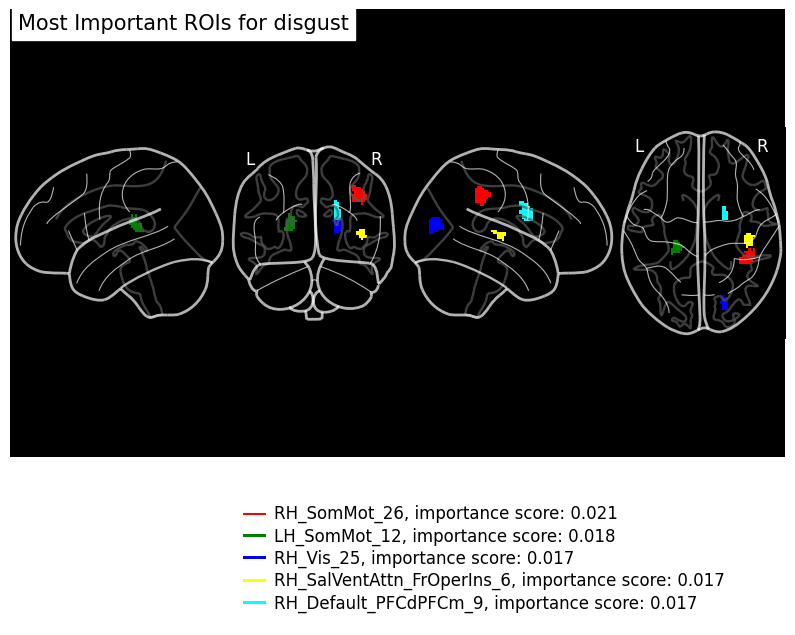}
\caption{Disgust}
\end{subfigure}
\hfill
\begin{subfigure}[t]{\roiwidth}
\includegraphics[width=\linewidth]{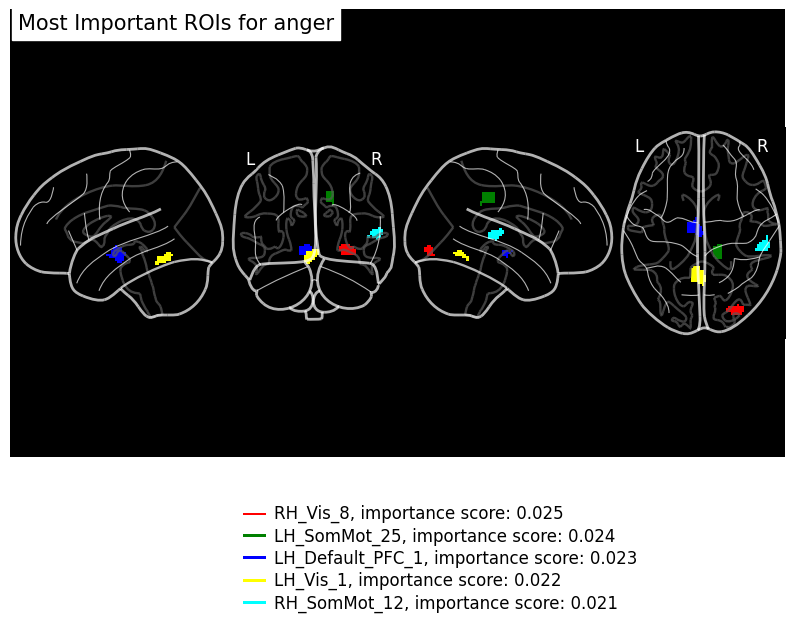}
\caption{Anger}
\end{subfigure}

\caption{Linear SVR ROI Importance Maps for the eight emotion categories.}
\label{fig:SVR_ROI_All}
\vspace{1cm}
\end{figure}


\begin{figure}[H]
\centering
\begin{subfigure}[t]{\roiwidth}
\includegraphics[width=\linewidth]{Images/ROI_explanations/rfr_roi_anticipation.png}
\caption{Anticipation}
\end{subfigure}
\hfill
\begin{subfigure}[t]{\roiwidth}
\includegraphics[width=\linewidth]{Images/ROI_explanations/rfr_roi_joy.png}
\caption{Joy}
\end{subfigure}
\hfill
\begin{subfigure}[t]{\roiwidth}
\includegraphics[width=\linewidth]{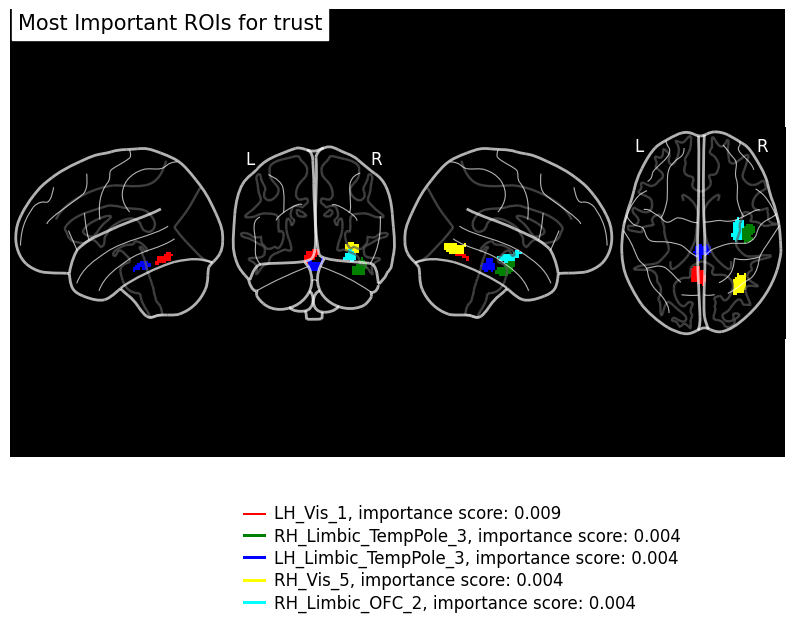}
\caption{Trust}
\end{subfigure}
\hfill
\begin{subfigure}[t]{\roiwidth}
\includegraphics[width=\linewidth]{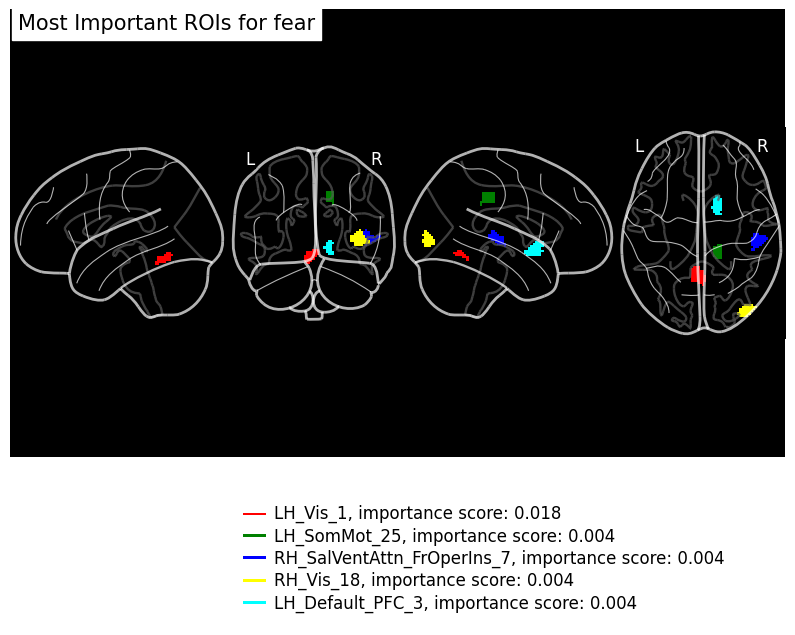}
\caption{Fear}
\end{subfigure}

\vspace{0.3cm}

\begin{subfigure}[t]{\roiwidth}
\includegraphics[width=\linewidth]{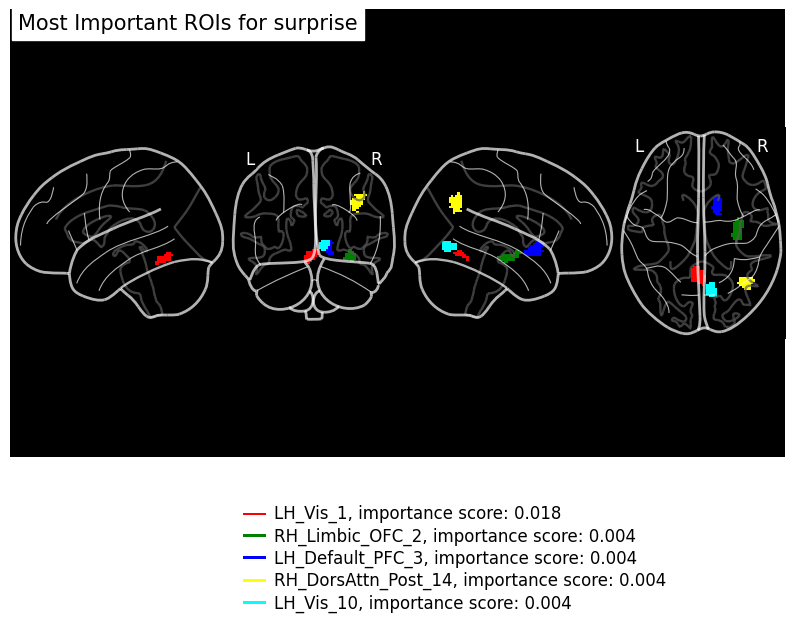}
\caption{Surprise}
\end{subfigure}
\hfill
\begin{subfigure}[t]{\roiwidth}
\includegraphics[width=\linewidth]{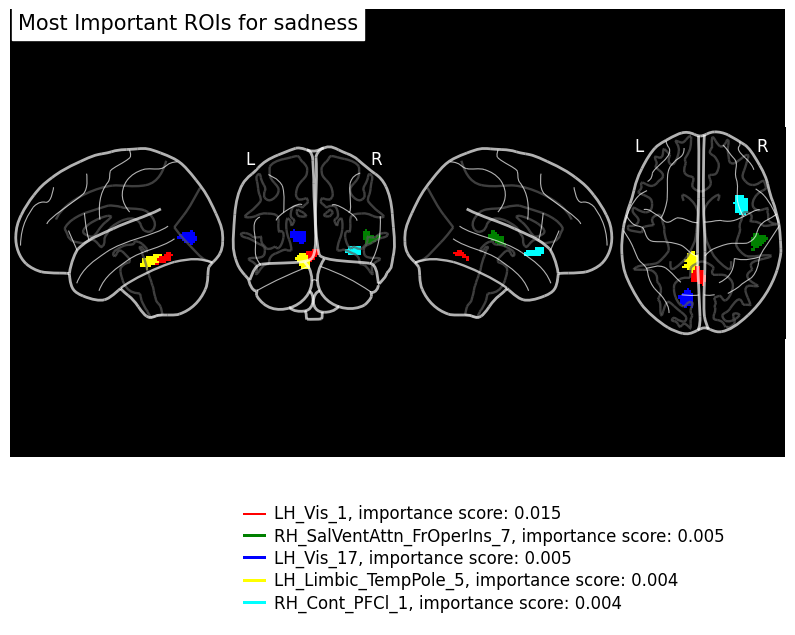}
\caption{Sadness}
\end{subfigure}
\hfill
\begin{subfigure}[t]{\roiwidth}
\includegraphics[width=\linewidth]{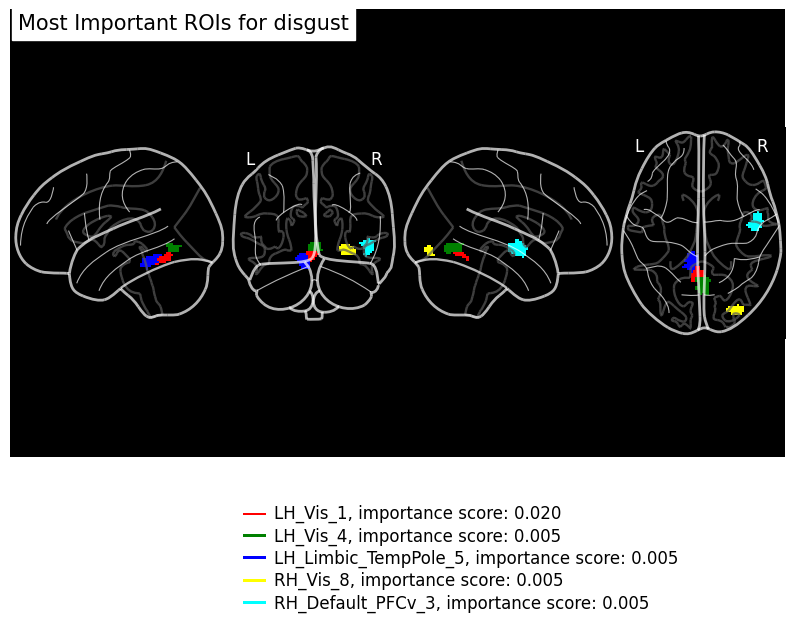}
\caption{Disgust}
\end{subfigure}
\hfill
\begin{subfigure}[t]{\roiwidth}
\includegraphics[width=\linewidth]{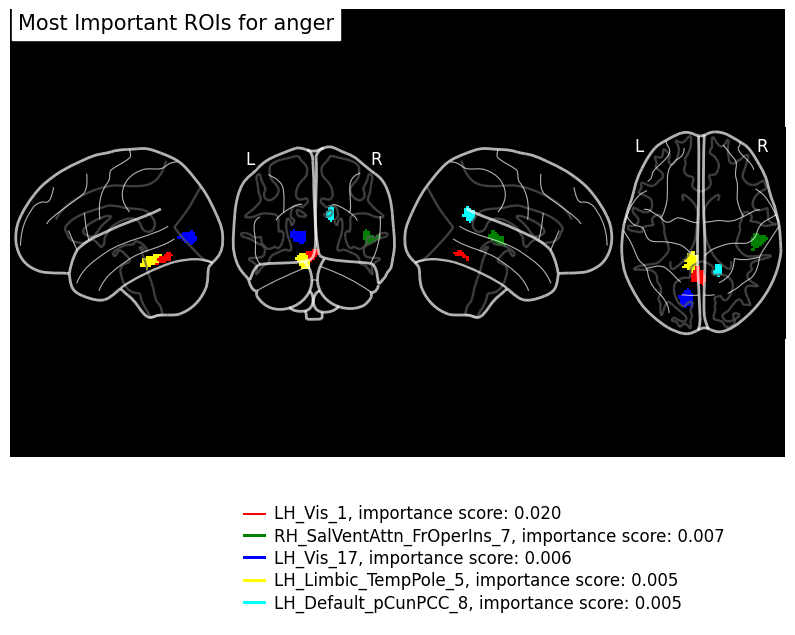}
\caption{Anger}
\end{subfigure}

\caption{RFR ROI Importance Maps for the eight emotion categories.}
\label{fig:RFR_ROI_All}
\vspace{2cm}
\end{figure}

\section{DFC Explanations} \label{Appendix DFC explanations}


\newcommand{\DFCEmotionFigure}[3]{ \begin{figure}[H] \centering \begin{subfigure}[t]{0.49\textwidth} \centering \includegraphics[width=\linewidth] {Images/DFC_explanations/#1/7N/#2_final_output.png} \caption{7-network parcellation} \end{subfigure} \hfill \begin{subfigure}[t]{0.49\textwidth} \centering \includegraphics[width=\linewidth] {Images/DFC_explanations/#1/17N/#2_final_output.png} \caption{17-network parcellation} \end{subfigure} \caption{ Maximum spanning trees of #3 model feature importance matrices for the emotion \textit{#2}, using the 7- and 17-network parcellations. } \end{figure} }

\subsection{Linear Model} 
\begin{figure}[H] 
\centering 
\begin{subfigure}[t]{0.48\textwidth} 
\centering 
\includegraphics[width=\linewidth, height=0.78\textheight] {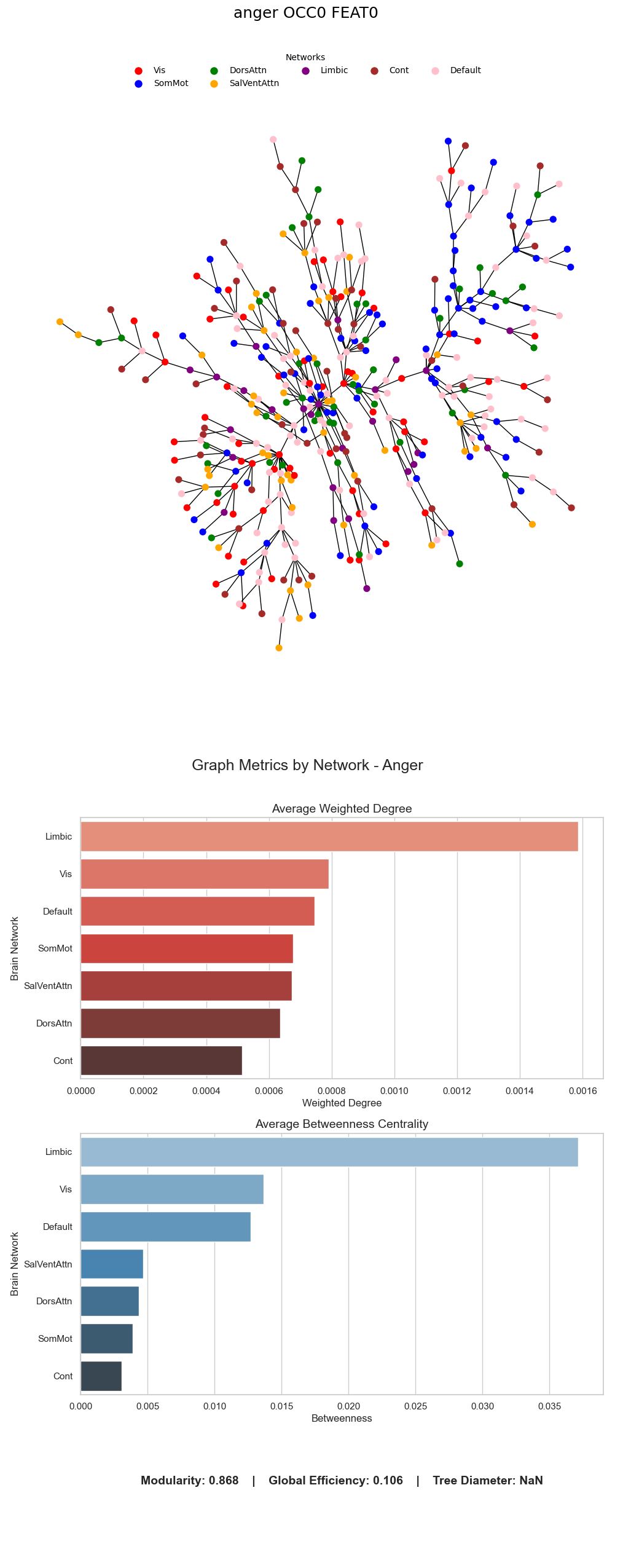} 
\caption{7-network parcellation} 
\end{subfigure} 
\hfill 
\begin{subfigure}[t]{0.48\textwidth} 
\centering 
\includegraphics[width=\linewidth, height=0.78\textheight] {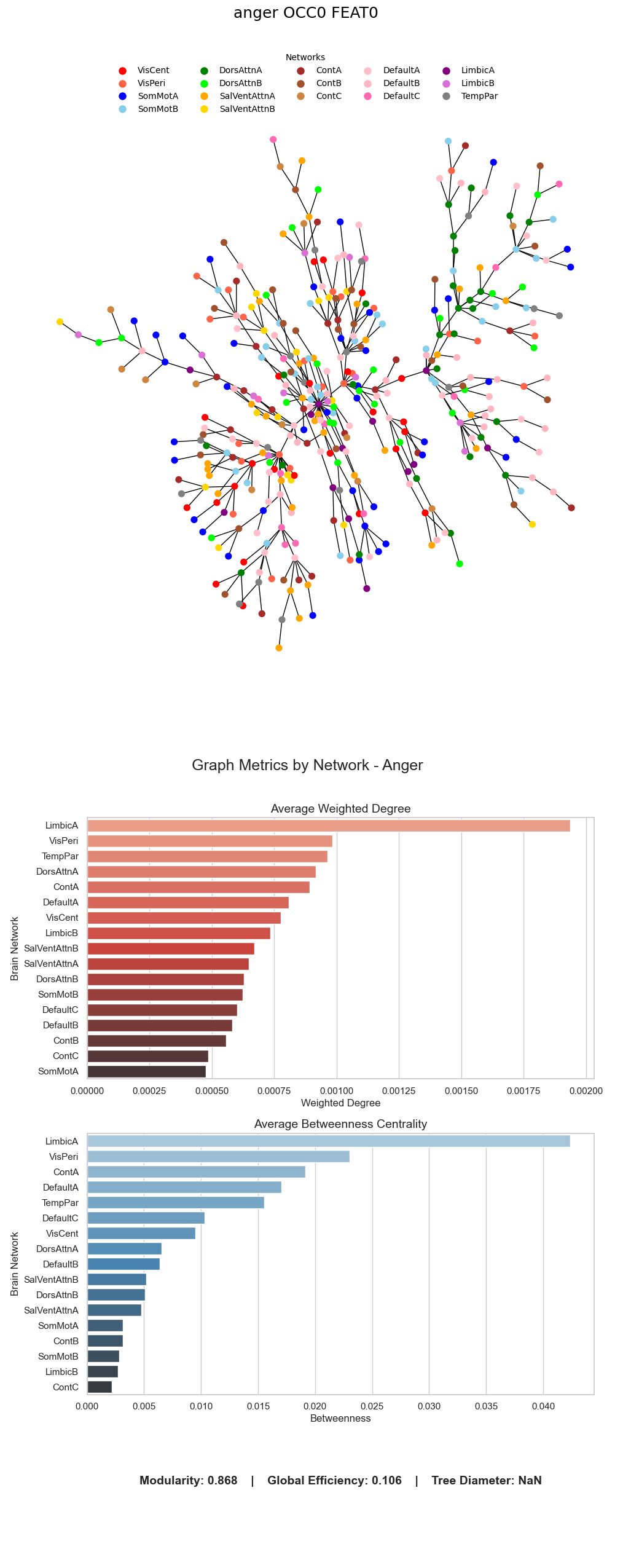} 
\caption{17-network parcellation} 
\end{subfigure} 
\caption{ Maximum spanning trees of Linear model feature importance matrices for the emotion \textit{anger}, using the 7- and 17-network parcellations. } 
\end{figure}

\DFCEmotionFigure{Linear}{anticipation}{Linear} 
\DFCEmotionFigure{Linear}{disgust}{Linear} 
\DFCEmotionFigure{Linear}{fear}{Linear} 
\DFCEmotionFigure{Linear}{joy}{Linear} 
\DFCEmotionFigure{Linear}{sadness}{Linear} 
\DFCEmotionFigure{Linear}{suprise}{Linear} 
\DFCEmotionFigure{Linear}{trust}{Linear} 


\subsection{Ridge Model} 

\begin{figure}[H] 
\centering 
\begin{subfigure}[t]{0.48\textwidth} 
\centering 
\includegraphics[width=\linewidth, height=0.85\textheight] {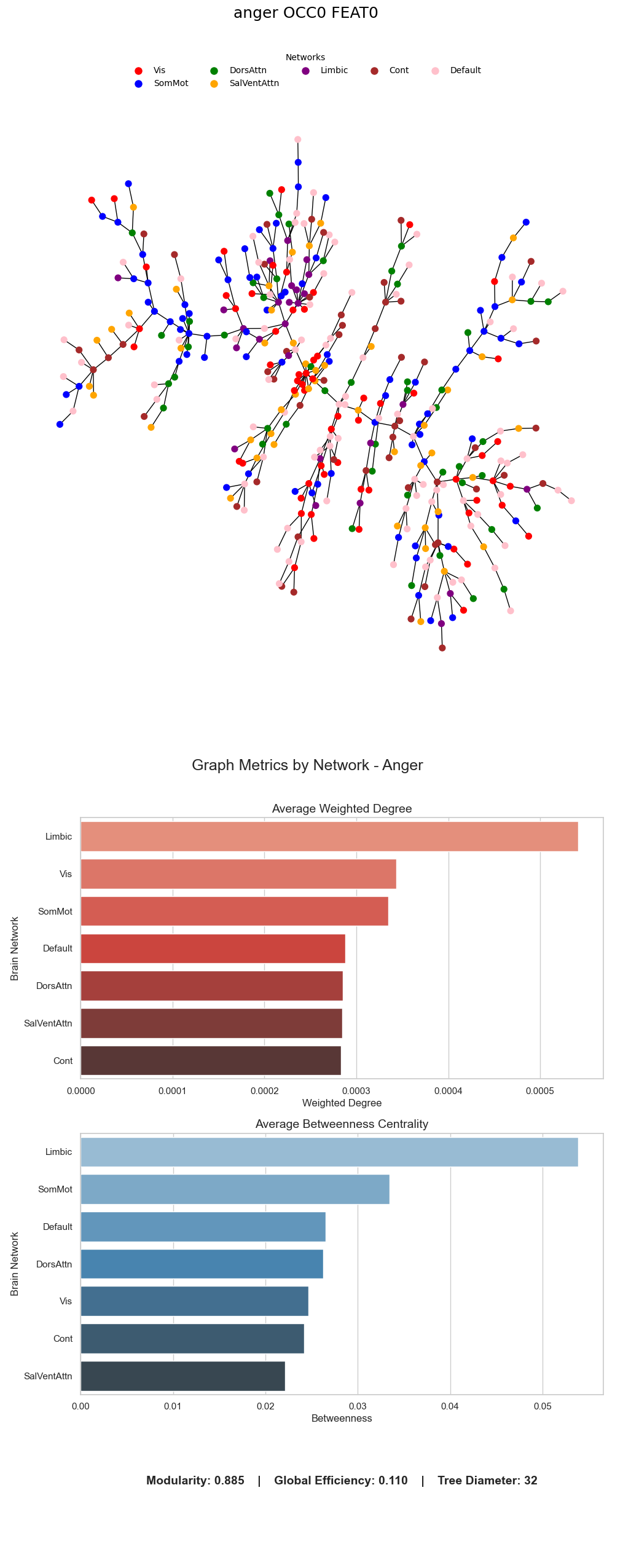} 
\caption{7-network parcellation} 
\end{subfigure} 
\hfill 
\begin{subfigure}[t]{0.48\textwidth} 
\centering 
\includegraphics[width=\linewidth, height=0.85\textheight] {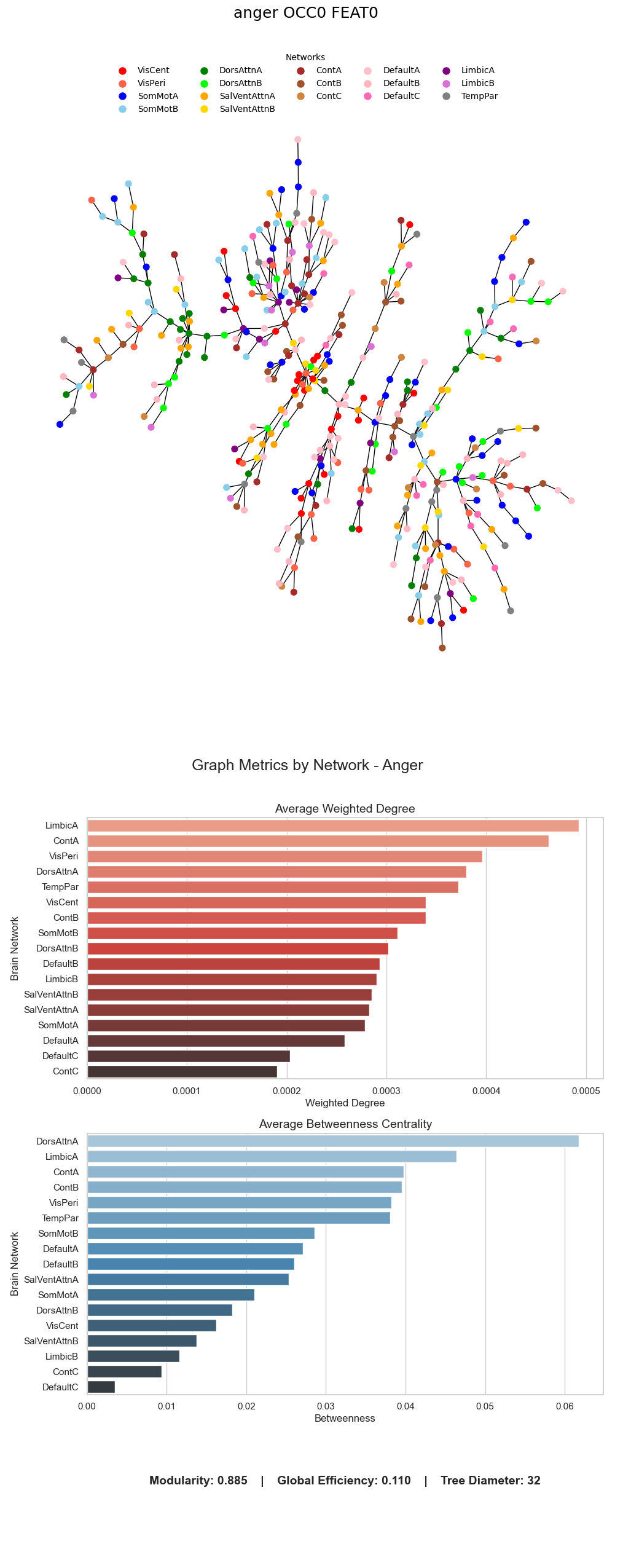} 
\caption{17-network parcellation} 
\end{subfigure} 
\caption{ Maximum spanning trees of Ridge model feature importance matrices for the emotion \textit{anger}, using the 7- and 17-network parcellations. } 
\end{figure}

\DFCEmotionFigure{Ridge}{anticipation}{Ridge} 
\DFCEmotionFigure{Ridge}{disgust}{Ridge} 
\DFCEmotionFigure{Ridge}{fear}{Ridge} 
\DFCEmotionFigure{Ridge}{joy}{Ridge} 
\DFCEmotionFigure{Ridge}{sadness}{Ridge} 
\DFCEmotionFigure{Ridge}{suprise}{Ridge} 
\DFCEmotionFigure{Ridge}{trust}{Ridge} 


\subsection{SVR Model} 

\begin{figure}[H] 
\centering 
\begin{subfigure}[t]{0.48\textwidth} 
\centering 
\includegraphics[width=\linewidth, height=0.85\textheight] {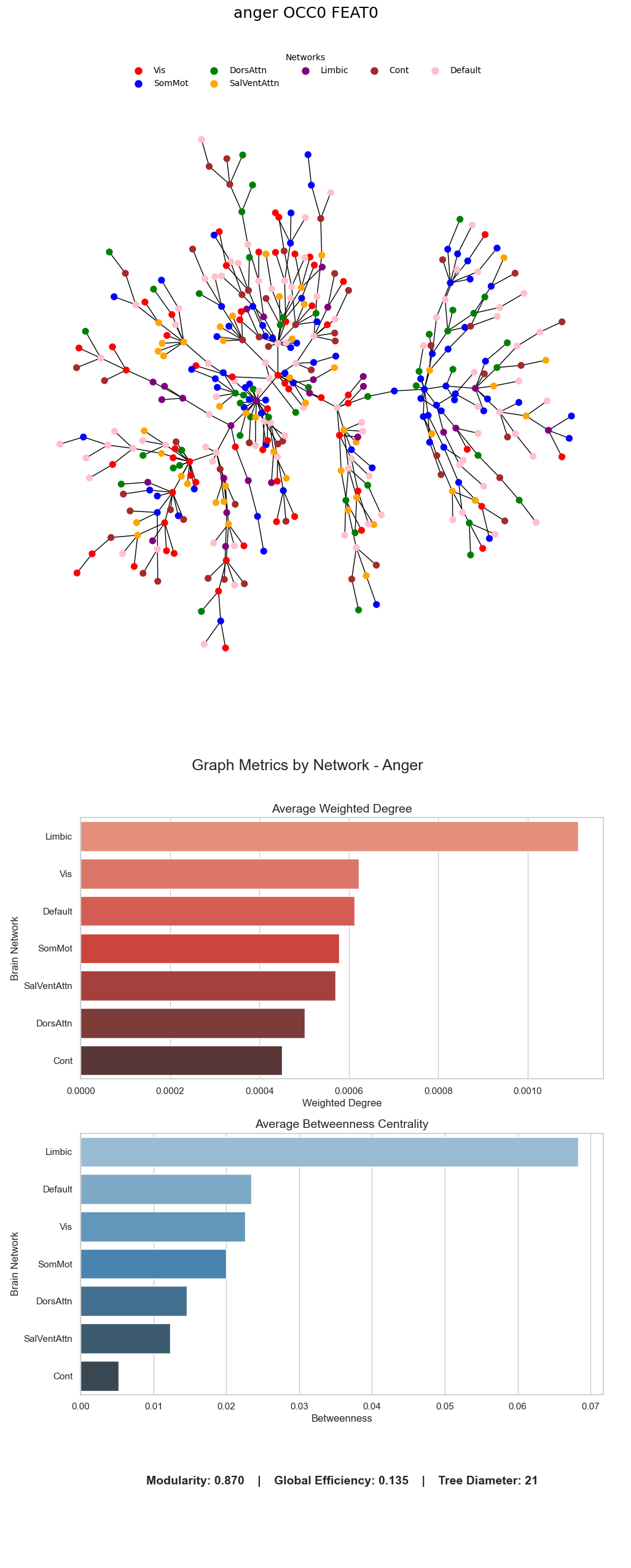} 
\caption{7-network parcellation} 
\end{subfigure} 
\hfill 
\begin{subfigure}[t]{0.48\textwidth} 
\centering 
\includegraphics[width=\linewidth, height=0.85\textheight] {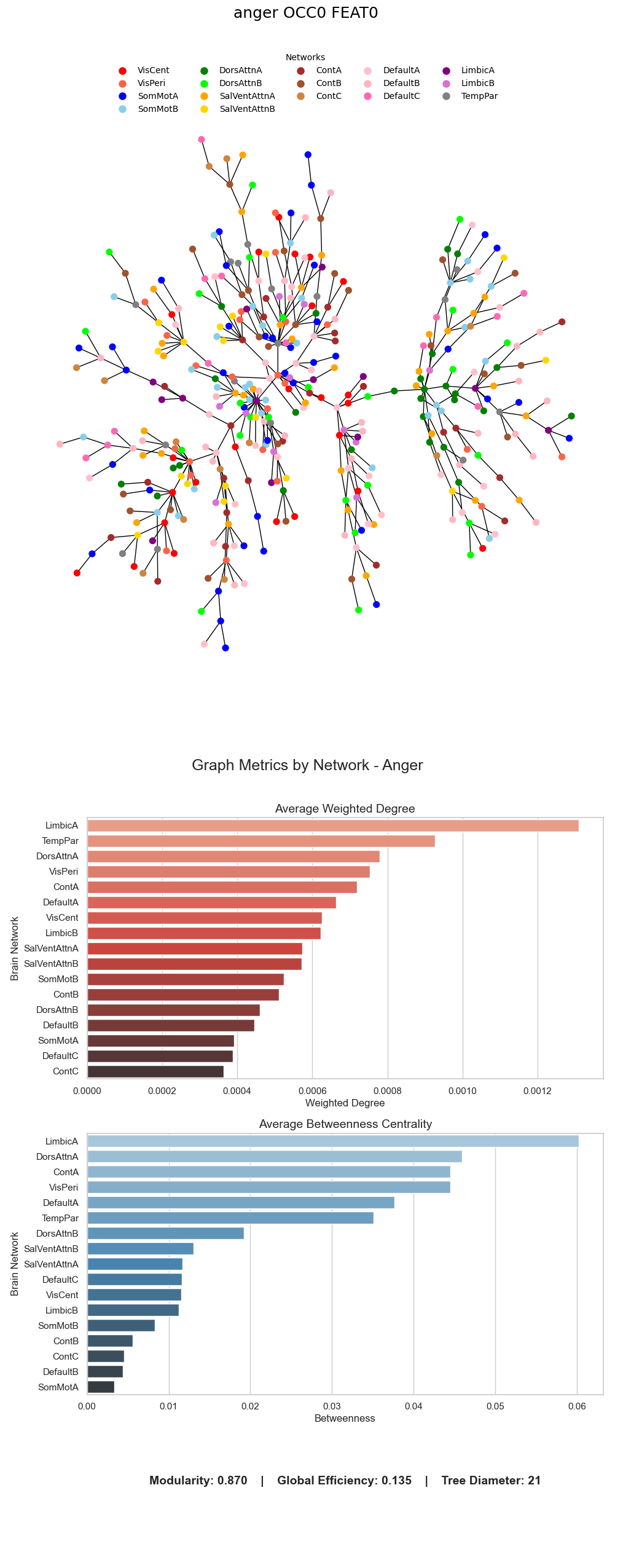} 
\caption{17-network parcellation} 
\end{subfigure} 
\caption{ Maximum spanning trees of SVR model feature importance matrices for the emotion \textit{anger}, using the 7- and 17-network parcellations. } 
\end{figure}

\DFCEmotionFigure{SVM}{anticipation}{SVR} 
\DFCEmotionFigure{SVM}{disgust}{SVR} 
\DFCEmotionFigure{SVM}{fear}{SVR} 
\DFCEmotionFigure{SVM}{joy}{SVR} 
\DFCEmotionFigure{SVM}{sadness}{SVR} 
\DFCEmotionFigure{SVM}{suprise}{SVR} 
\DFCEmotionFigure{SVM}{trust}{SVR} 


\subsection{RFR Model} 

\begin{figure}[H] 
\centering 
\begin{subfigure}[t]{0.48\textwidth} 
\centering 
\includegraphics[width=\linewidth, height=0.85\textheight] {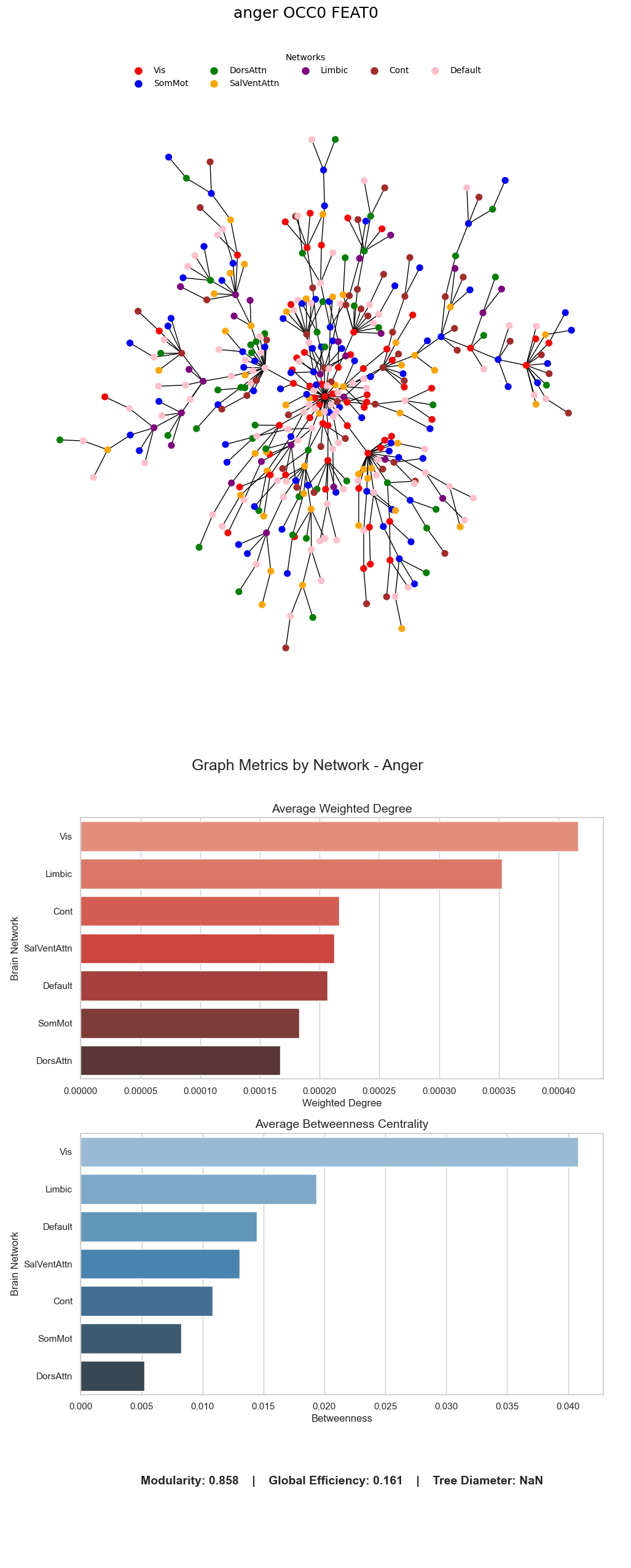} 
\caption{7-network parcellation} 
\end{subfigure} 
\hfill 
\begin{subfigure}[t]{0.48\textwidth} 
\centering 
\includegraphics[width=\linewidth, height=0.85\textheight] {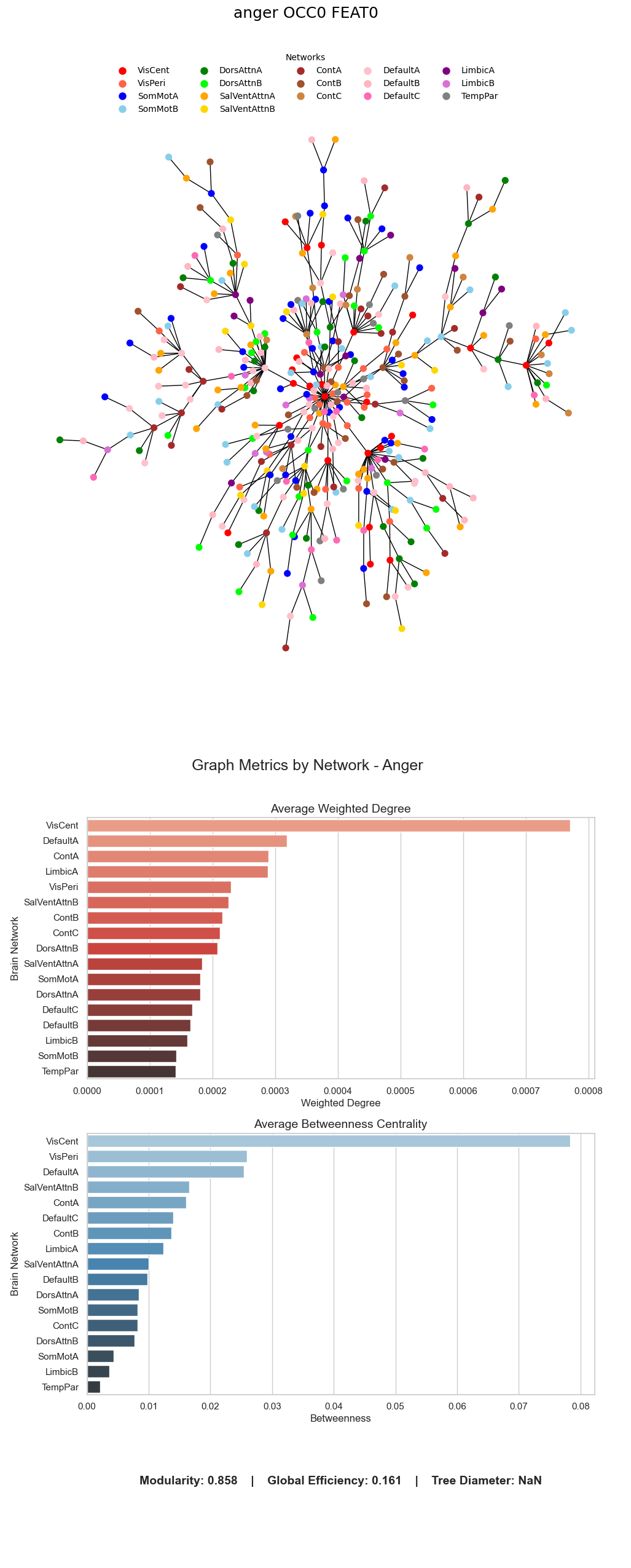} 
\caption{17-network parcellation} 
\end{subfigure} 
\caption{ Maximum spanning trees of RFR model feature importance matrices for the emotion \textit{anger}, using the 7- and 17-network parcellations. } 
\end{figure}

\DFCEmotionFigure{RFR}{anticipation}{RFR} 
\DFCEmotionFigure{RFR}{disgust}{RFR} 
\DFCEmotionFigure{RFR}{fear}{RFR} 
\DFCEmotionFigure{RFR}{joy}{RFR} 
\DFCEmotionFigure{RFR}{sadness}{RFR} 
\DFCEmotionFigure{RFR}{suprise}{RFR} 
\DFCEmotionFigure{RFR}{trust}{RFR}


\subsection{Lasso Model} 

\begin{figure}[H] 
\centering 
\begin{subfigure}[t]{0.48\textwidth} 
\centering 
\includegraphics[width=\linewidth, height=0.85\textheight] {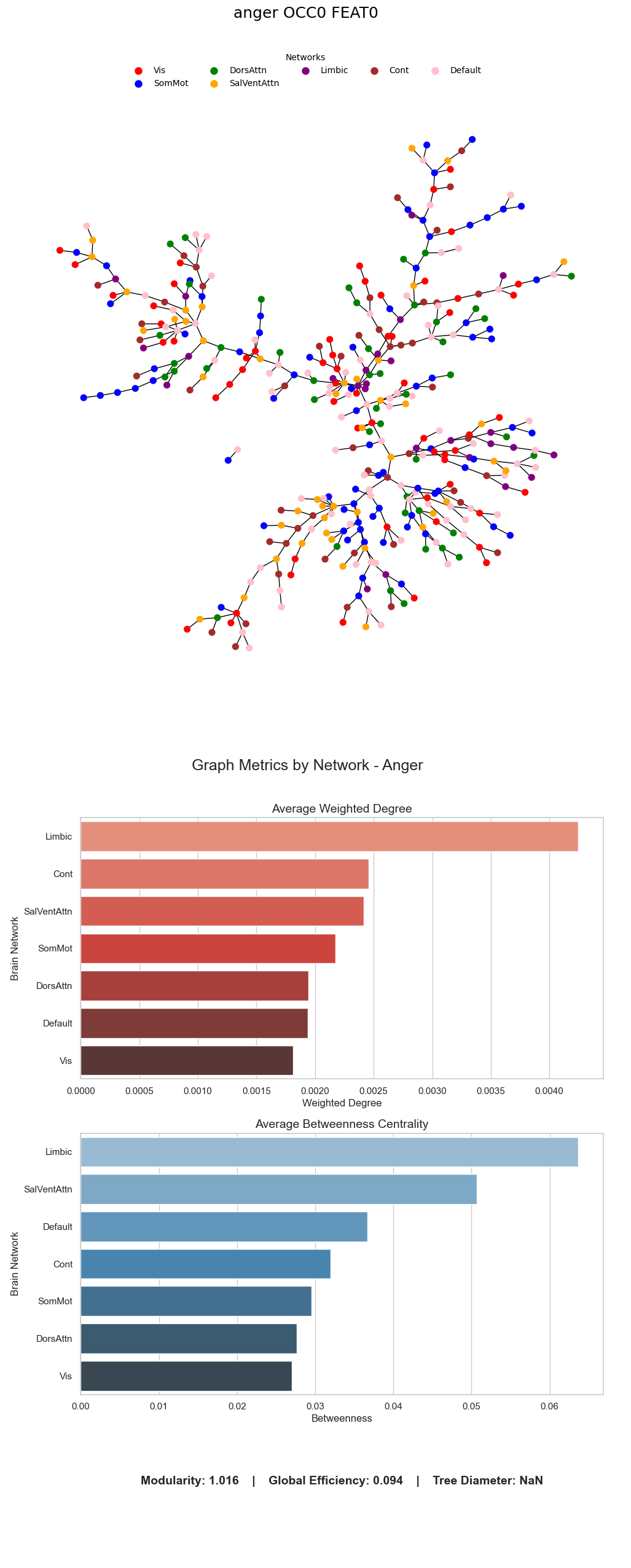} 
\caption{7-network parcellation} 
\end{subfigure} 
\hfill 
\begin{subfigure}[t]{0.48\textwidth} 
\centering 
\includegraphics[width=\linewidth, height=0.85\textheight] {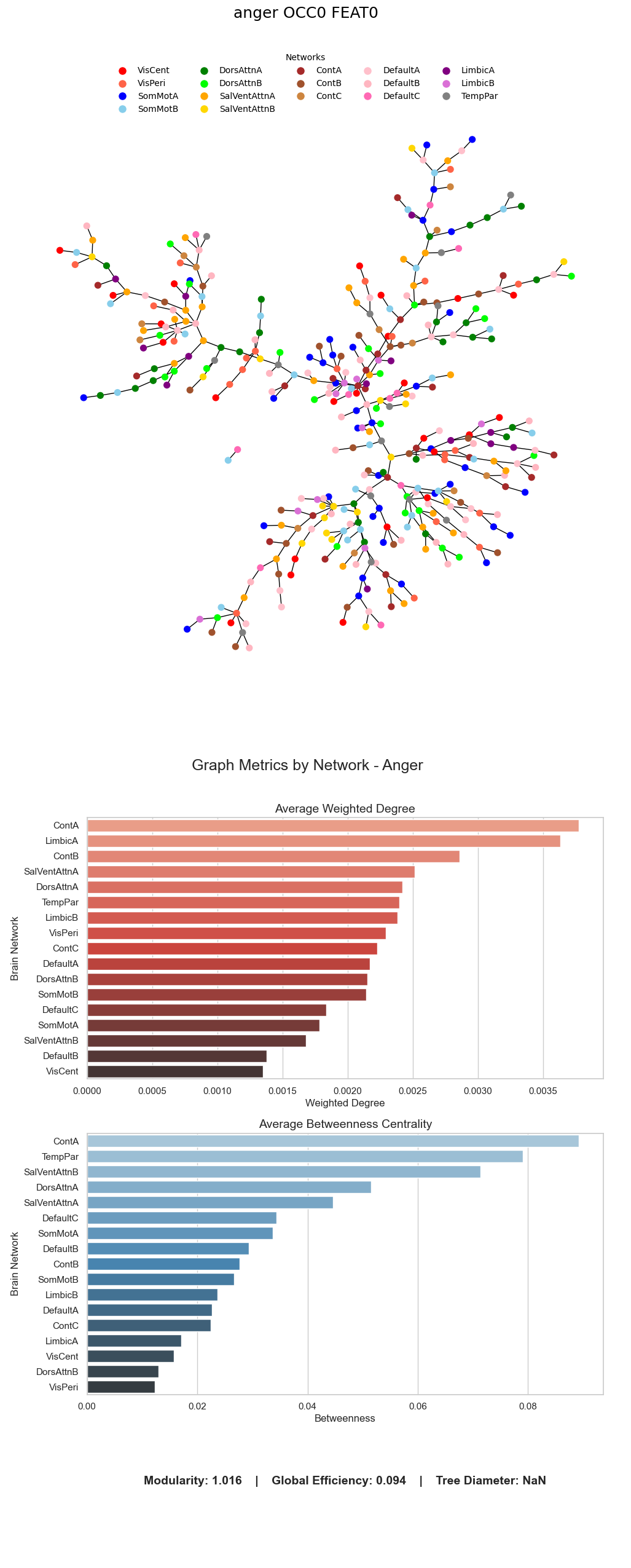} 
\caption{17-network parcellation} 
\end{subfigure} 
\caption{ Maximum spanning trees of Lasso model feature importance matrices for the emotion \textit{anger}, using the 7- and 17-network parcellations. } 
\end{figure}

\DFCEmotionFigure{Lasso}{anticipation}{Lasso} 
\DFCEmotionFigure{Lasso}{disgust}{Lasso} 
\DFCEmotionFigure{Lasso}{fear}{Lasso} 
\DFCEmotionFigure{Lasso}{joy}{Lasso} 
\DFCEmotionFigure{Lasso}{sadness}{Lasso} 
\DFCEmotionFigure{Lasso}{suprise}{Lasso} 
\DFCEmotionFigure{Lasso}{trust}{Lasso}

\end{document}